\documentclass{applemlr}

\usepackage{amsmath}
\usepackage{enumerate}
\usepackage{algorithm}
\usepackage{algpseudocode}
\usepackage{amsfonts}
\usepackage{amsthm}
\usepackage{cleveref}
\usepackage{diagbox}
\usepackage{colortbl}
\usepackage{amssymb}
\usepackage{xspace}
\usepackage{wrapfig}
\usepackage{adjustbox}
\usepackage{tabularx}
\usepackage{booktabs}
\usepackage{mathtools}
\usepackage{tikz}
\usepackage{enumitem}
\usepackage{silence}
\usepackage{dsfont}
\usepackage[table]{xcolor}
\usepackage[dvipsnames]{xcolor}
\usepackage{multirow}
\usepackage{makecell}
\usepackage{xfakebold}

\usepackage{amsmath,amsfonts,bm}

\def\eqref#1{equation~\ref{#1}}

\def\1{\bm{1}}

\def\mK{{\bm{K}}}

\def\mV{{\bm{V}}}

\DeclareMathAlphabet{\mathsfit}{\encodingdefault}{\sfdefault}{m}{sl}
\SetMathAlphabet{\mathsfit}{bold}{\encodingdefault}{\sfdefault}{bx}{n}

\definecolor{textgray}{HTML}{6E6E73}
\usetikzlibrary{positioning, calc}
\usetikzlibrary{decorations.pathmorphing}

\makeatletter
\patchcmd{\wrong@fontshape}{\@gobbletwo}{}{}{}
\makeatother
\numberwithin{equation}{section}
\makeatletter
\AtBeginDocument{
  \urlstyle{sf}
  
}
\makeatother

\definecolor{light}{RGB}{125, 125, 125}
\crefname{tcb@cnt@pbox}{code}{code}
\Crefname{tcb@cnt@pbox}{Code}{Code}
\crefname{assumption}{assumption}{assumption}
\Crefname{assumption}{Assumption}{Assumptions}

\newtcolorbox[auto counter]{pbox}[2][]{
  colback=white,
  title=Code~\thetcbcounter: #2,
  #1,fonttitle=\sffamily,
  fontupper=\sffamily,
  arc=2pt,
  colframe=bgcolor,
  coltitle=fgcolor,
  colbacktitle=bgcolor,
  toptitle=0.25cm,
  bottomtitle=0.125cm
}

\makeatletter
\newcommand\applefootnote[1]{%
  \begingroup
  \renewcommand\thefootnote{}%
  \renewcommand\@makefntext[1]{\noindent##1}%
  \footnote{#1}%
  \addtocounter{footnote}{-1}%
  \endgroup
}
\makeatother

\definecolor{cverbbg}{gray}{0.90}

\usepackage{url}
\usepackage{graphicx}
\usepackage{subcaption}
\usepackage[most]{tcolorbox}
\usepackage{pifont}

\usetikzlibrary{shapes.geometric, arrows.meta, positioning, fit, backgrounds, calc}

\title{\textsc{Cartridges++}: KV Cache Compression\\ without Off-Context Derailment}

\author{Sonia Laguna}
\author{Joao Monteiro}
\author{Marco Cuturi}
\author{Pierre Ablin}
\author{Eleonora Gualdoni}

\affiliation{Apple}

\abstract{
Serving long documents to a Large Language Model (LLM) repeatedly is expensive: computations grow with context length, and the memory footprint of the key-value (KV) cache balloons.
\textit{Compressed} KV (CKV) representations aim to mimic the cache of a document and are typically computed once and for all, ahead of inference time.
Methods to obtain CKVs range from drop mechanisms that reduce their number of columns, to learned approaches. Among the latter, \textit{Cartridges} have emerged as a leading compression method, learning compact KV representations through distillation on relevant Q/A pairs.
While existing evaluations focus primarily on whether Cartridges and other CKVs yield approximately similar responses to document-related, \textit{on-context} queries, we investigate the crucial deployment question of whether they can handle \textit{off-context} queries, something the native KV representation is particularly good at, thanks to the mechanics of attention.
We observe a fundamental trade-off: while Cartridges perform better for on-context queries, heuristic-variants preserve better the original LLM's ability to operate off-context. We measure this through their capability to avoid context contamination in their response, retain general knowledge, and follow instructions.
We propose \textit{\textsc{Cartridges++}}, simple modifications to Cartridges that retain off-context abilities at small or negligible cost. The \textit{router} variant decides at inference time whether the query should use the learned long-context memory, while the \textit{data-mixing} variant allocates a small fraction of training Q/As to queries outside the reference long document.
Our study shows that assessing CKVs on document utility alone can mask substantial degradation in broader model capabilities, yet those issues can be fixed with benign changes to CKV inference or training.\looseness-1
}

\metadata[Correspondence]{\sffamily SL: \url{slaguna@ethz.ch}; JM: \url{jmonteiro2@apple.com}; MC: \url{m_cuturi@apple.com}; PA: \url{p_ablin@apple.com}; EG: \url{e_gualdoni@apple.com}.}
\metadata[Note]{\sffamily SL: work done as intern at Apple.}
\date{\sffamily\today}

\begin{document}

\maketitle

\section{Introduction}
\label{intro}

\paragraph{Long-context reuse is expensive.}
Large language models (LLMs) are increasingly used to reason over persistent,
long-form information, including reports, knowledge bases, or
medical records~\citep{bai2024longbench}. In these settings, the same context is
reused across many queries, creating a need for memory mechanisms that persist this context across queries~\citep{zhang2023h2o,xiao2024streaming,
eyuboglu2025cartridges,hardalov2026cartridges}.
Transformers naturally provide such a representation through the key--value
(KV) cache, which stores a state for every prefix token at every layer. 
However, retaining the cache introduces a second cost: its memory footprint grows linearly with context length, and the increasing length of the cache also raises decoding cost, making long contexts expensive both to store and to serve.
This motivates \textit{compressed} KV (CKV) representations: compact versions of the document's
KV cache that can be computed once and reused across subsequent queries. CKVs can be constructed through retained-state methods that select or merge existing KV states~\citep{zhang2023h2o,xiao2024streaming,li2024snapkv,
yang2024pyramidinfer,kim2025kvzip}, or learned methods that optimize compact
representations~\citep{zweiger2026fast,eyuboglu2025cartridges,hardalov2026cartridges}.
We study reusable, document-specific CKVs for a fixed pretrained LLM: each
memory is constructed for one document and reused without knowing future
queries. Among learned variants, \textit{Cartridges} have emerged as a prominent method, optimizing document-specific KV states through distillation on question--answer pairs~\citep{eyuboglu2025cartridges}.\looseness-1

\begin{figure}[t]
    \centering
    \begin{subfigure}[c]{0.62\textwidth}
        \centering
        \includegraphics[width=\linewidth]{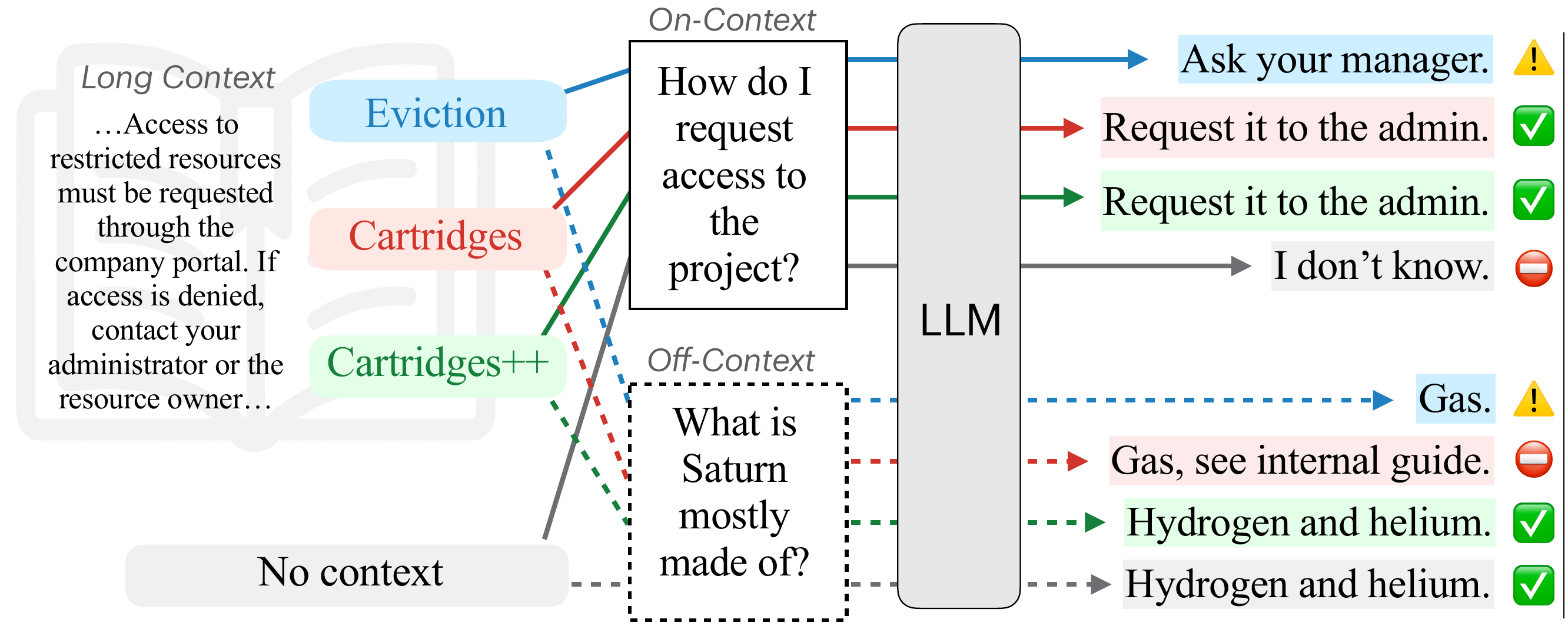}
    \end{subfigure}\hfill
    \begin{subfigure}[c]{0.375\textwidth}
        \centering
        \includegraphics[width=\linewidth]{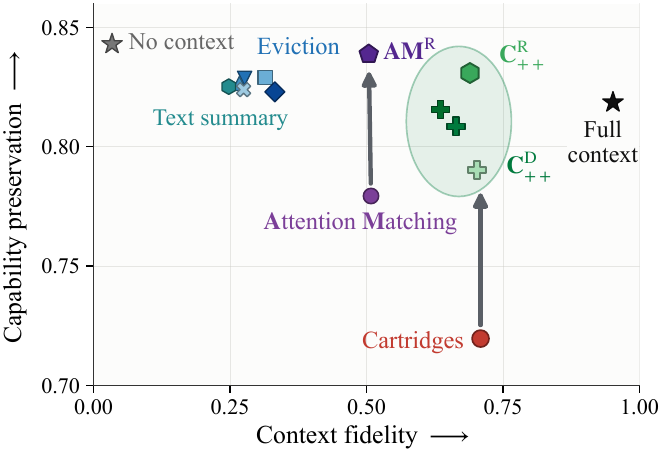}
    \end{subfigure}
\caption{
\textbf{Cartridges preserve context fidelity but disrupt capability; \textsc{Cartridges++} cancels that degradation.} \textit{Left}:
illustrative responses to an on-context query (solid arrows) and an unrelated Saturn
question (dashed arrows) handling context with eviction, Cartridges and our method. \textit{Right}: aggregate context fidelity and capability preservation across long-context datasets (see Appendix~\ref{app:per-benchmark} for details). $C_{++}^{D}$ and $C_{++}^{R}$ denote data mixing and routing variants of \textsc{Cartridges++},
$\mathrm{AM}^{\mathrm{R}}$ routed Attention Matching. The arrows highlight the gains attributed to our variants: they retain the same context fidelity while upgrading significantly their general capabilities.
}
    \label{fig:overview}
    \vspace*{-0.6cm}
\end{figure}

\vspace*{-0.2cm}
\paragraph{Cartridges amplify context interference.}
Cartridges offer a favorable compression--utility trade-off, preserving document-level performance where retained-state methods degrade.
Yet, this notion of utility leaves open a complementary question:
\textit{at what cost?}
Recent work shows that answer accuracy alone is insufficient to evaluate eviction methods for
KV compression~\citep{wang2025semantic,liu2025pitfalls,ai2026evidence}, which can discard semantically important information or disrupt instructions and reasoning in the context.
These evaluations, however, remain \emph{on context}: they concern what compression preserves or loses from the source document itself. They do not test the reality of deployment where future queries need not concern the stored document.\looseness-1

As the interaction with a user slightly veers off-context, we argue that CKVs can be more brittle and run the risk of being influenced by the stored memory when the query does not call for it.
For this reason, CKV mechanisms should also be evaluated \textit{off-context}. The rationale is that the mechanics of attention can efficiently turn off interactions between the native KV caches and queries that are unrelated to the context. We posit (and verify experimentally) that such a property is lost for CKVs, particularly learned ones such as Cartridges.
A full long context already induces interference on such queries~\citep{shi2023large,yoran2024irrelevant}, and heuristic eviction methods retaining native KV states show a similar, limited degradation. A recent hybrid method, Attention Matching~\citep{zweiger2026fast}, retains selected native keys while fitting attention biases and new values, and occupies an intermediate regime of off-context degradation.
Cartridges, however, sit at the more disruptive end of this spectrum. Despite providing substantially stronger context fidelity, they introduce larger interference, perturbing the model even more than the full, uncompressed context they are meant to approximate. We call the missing
requirement \emph{capability preservation}: when a memory is irrelevant, its
presence should not substantially degrade the underlying LLM's behavior relative to operating without the context.
Thus, the method that best preserves the utility of the stored document is also the one for which context-external interference becomes a substantial deployment concern.
To our knowledge, we are the first to expose this hidden cost of Cartridges, characterizing its failure modes and evaluating mitigations, as illustrated in Figure~\ref{fig:overview}.\looseness-1

\vspace*{-0.25cm}
\paragraph{Mitigating Cartridge interference.}
This observation motivates targeting Cartridges directly. We seek to preserve
their context fidelity while recovering the benign off-context behavior of the
underlying model.
We introduce \textsc{\textsc{Cartridges++}}, with two
complementary strategies. First, an inference-time relevance router decides
whether an already-trained Cartridge should be used. Second, a training-time
strategy mixes document-relevant examples with a small fraction of general-purpose
ones. Together, these strategies preserve the LLM capabilities when the stored memory is
irrelevant and its utility on document-dependent queries. Our
contributions are as follows:\looseness-1

\begin{itemize}[nosep, leftmargin=*]
  \item We introduce a two-sided evaluation of reusable, document-specific CKVs  that measures not only context fidelity on relevant queries, but also capability preservation on unrelated queries across general knowledge, instruction following, and context contamination during generation.

    \item We quantify off-context interference in Cartridges: despite their strong on-context fidelity, Cartridges amplify interference off-context.

    \item We introduce \textsc{\textsc{Cartridges++}} to mitigate off-context derailment when Cartridges are loaded. We propose two complementary variants: $C_{++}^{D}$, which uses data mixing during training, and $C_{++}^{R}$, which leverages a relevance router at inference. Across four long-context benchmarks, both improve capability preservation while retaining the document-utility gains.\looseness-1

    \looseness-1
\end{itemize}

\section{Related Work}
\label{rel_work}
\vspace*{-0.1cm}
\paragraph{Training-free KV compression.}
KV compression is particularly useful when a long document is processed once and reused across many future queries. We focus on this setting, with query-independent KV compression and a fixed pretrained LLM.
Training-free eviction retains a subset of the document's original KV states. Representative methods differ in how they select these states:
H2O keeps recent
tokens and accumulated attention heavy hitters~\citep{zhang2023h2o},
StreamingLLM preserves attention sinks and a recent window
~\citep{xiao2024streaming}, SnapKV selects head-specific positions using an
observation window near the end of the prompt~\citep{li2024snapkv}, and
PyramidKV and PyramidInfer allocate non-uniform layer budgets
~\citep{cai2024pyramidkv,yang2024pyramidinfer}. Like SnapKV, the latter methods
use the downstream query for selection in their standard formulations and are
therefore query-dependent. KVzip instead uses context reconstruction to build a
query-agnostic cache~\citep{kim2025kvzip}. %
Unlike learned methods, eviction compresses by discarding native KV states rather than optimizing new representations, and was primarily developed for decoding efficiency or bounded-context settings.
Text-space methods such as LLMLingua~\citep{jiang2023llmlingua} shorten the prompt rather than directly optimizing KV states; we use budget-matched model-generated summaries as a representative baseline.\looseness-1 

\vspace*{-0.1cm}
\paragraph{Learned context representations.}
Learned compressors either amortize a shared encoding mechanism across documents~\citep{mu2023gist,chevalier2023autocompressors,ge2024icae,li2025fivehundredcompressor} or optimize a new representation for each source.
Cartridges take the latter approach, directly optimizing document-specific key and value tensors in the frozen LLM through self-study, without training an encoder or modifying model weights~\citep{eyuboglu2025cartridges}.
Attention Matching follows a closely related document-specific formulation, but constructs the compressed cache by retaining selected original keys and fitting attention biases and new values through least-squares objectives that reconstruct full-cache attention~\citep{zweiger2026fast}.
Finally, Cartridges at Scale extends the approach to document collections~\citep{hardalov2026cartridges}.

\vspace*{-0.1cm}
\paragraph{Evaluation and off-context robustness.}
Recent work shows that answer accuracy can conceal failures of KV compression,
including loss of reasoning-relevant content, instruction disruption, and degraded
evidential support~\citep{wang2025semantic,liu2025pitfalls,ai2026evidence}.
These analyses remain largely \emph{context-internal}, asking what compression
loses from the source context.~\citet{eyuboglu2025cartridges} report only a small MMLU evaluation
in a parameterization ablation comparing prefix tuning with
LoRA, rather than a dedicated study of
off-context failures or their mitigation. We instead study \emph{context-external}
behavior across general knowledge, instruction following, and contamination,
using full-context baselines to separate ordinary distraction from interference
induced by optimized KV state.
Related work motivates preservation and selective activation: RAG systems
filter or route irrelevant context~\citep{shi2023large,lewis2020retrieval,yoran2024irrelevant,xu2023recomp,
asai2024selfrag,luo2025zerorag}, while memory-based methods study locality and
selective activation~\citep{mitchell2022serac,wang2024wise,hartvigsen2023grace,
chen2024recipe,diao2026doctoatom,zheng2026latentmemorymanagement}.
We adapt these ideas to document-specific KV memories through post-hoc routing
and no-context-teacher data mixing. Appendix~\ref{app:additional-related-work} discusses complementary cache-reduction techniques and broader work on learned context representations and memory reliability.\looseness-1

\newpage
\section{Problem Formalization}
\label{background}
\subsection{KV cache and compression}
Consider a decoder-only transformer processing a document
$x=(x_1,\ldots,x_L)$. At each layer $\ell$, self-attention produces key and
value states $\mK^\ell$ and $\mV^\ell$, which are cached for
autoregressive generation. Here, $\mK^\ell,\mV^\ell\in\mathbb R^{H_\ell\times L\times d_\ell}$, with $H_\ell$ KV heads of dimension $d_\ell$ (batch dimension omitted). We denote the full document cache by
    $\mathcal{C}(x)=\{(\mK^\ell,\mV^\ell)\}_{\ell=1}^{N}$.
Its memory footprint grows linearly with document length $L$ and model depth
$N$, motivating compression when a document is reused across many
queries.
A KV compression method constructs a compact representation $\mathcal C_x$ of length $m \ll L$, which replaces the full document cache at inference time.
We define the compression factor as
$\rho=m/L$.
With CKV, the cost of generating one token goes from $\mathcal{O}(L)$ to $\mathcal{O}(m)$: $\rho$ quantifies the acceleration gains of CKV for generation with long context.
Given a query $q$, the compressed memory
induces
    $p_\theta(y\mid {\mathcal C}_x,q)$.
Retained-state methods  construct ${\mathcal C}_x$ by selecting or combining states from the original document
cache, whereas
Cartridges optimize new compact KV states by
distilling supervision from document-related question--answer pairs.
When discussing Cartridges below, $\mathcal C_x$ denotes this learned
document-specific memory.

\subsection{Distillation-based compression: Cartridges}

A Cartridge for document $x$ is a
document-specific trainable KV cache
    $\mathcal{C}_x = \{(\mK_C^\ell,\mV_C^\ell)\}_{\ell=1}^{N}$,
with $m$ virtual KV positions per layer. The LLM parameters $\theta$ remain
frozen; only the key and value tensors in $\mathcal C_x$ are optimized.
Cartridges are trained through \emph{self-study}, where the document is divided into
subcontexts $\tilde{x}\subset x$, from which the frozen model generates
synthetic conversational traces
$s^{(j)}=(s^{(j)}_1,\ldots,s^{(j)}_{T_j})$ of varying length $T_j$.
This produces a training set
$\mathcal D_x=\{(s^{(j)},\tilde{x}_j)\}_{j=1}^{n}$.
For each trace, the model with $\tilde{x}$ in context acts as the teacher,
while the same frozen model conditioned on $\mathcal{C}_x$ acts as the student.
The Cartridge is optimized by context distillation:
\vspace*{-0.15cm}
\begin{equation}
\vspace*{-0.1cm}
\mathcal{C}_x^*
=
\arg\min_{\mathcal{C}_x}
\sum_{(s,\tilde{x})\in\mathcal D_x}
\sum_{t=1}^{|s|}
D_{\mathrm{KL}}\!\left(
p_\theta(\cdot\mid \tilde{x},s_{<t})
\;\middle\|\;
p_\theta(\cdot\mid \mathcal{C}_x,s_{<t})
\right).
\label{eq:cartridge-objective}
\end{equation}
At inference time, $\mathcal{C}_x^*$ is loaded as the prefix cache and is reused across queries without access to $x$.

\subsection{Two-sided requirements on a reusable memory}
\label{background:two-sided}

Let
$\mathcal Q_x^+$ denote queries that concern information from document $x$, and
$\mathcal Q_x^-$ queries that do not. For $q\sim\mathcal Q_x^+$, the compressed
memory should reproduce the behavior of the model conditioned on the full
document. For $q\sim\mathcal Q_x^-$, it should ideally preserve the model's
behavior without the document. %
 Using a divergence $\mathfrak D$ between model output distributions, we
formalize these requirements as\looseness-1
\begin{align}
R^+(\mathcal C_x)
&=
\mathbb{E}_{q\sim\mathcal Q_x^+}
\mathfrak D\!\left(
p_\theta(\cdot\mid x,q)
\,\middle\|\,
p_\theta(\cdot\mid \mathcal C_x,q)
\right),
\label{eq:rplus}\\
R^-(\mathcal C_x)
&=
\mathbb{E}_{q\sim\mathcal Q_x^-}
\mathfrak D\!\left(
p_\theta(\cdot\mid q)
\,\middle\|\,
p_\theta(\cdot\mid \mathcal C_x,q)
\right).
\label{eq:rminus}
\end{align}
Here, $R^+$ measures \emph{context fidelity}: how closely the learned memory
reproduces full-context behavior on queries that require $x$. In contrast,
$R^-$ measures \emph{capability preservation}: how much the memory perturbs the
model's original behavior on queries that do not require $x$. Note that this is not a new requirement we impose:~\citet{eyuboglu2025cartridges} explicitly identify generality across \textit{diverse} user prompts as a desideratum for functional equivalence with in-context learning. 
In practice, however, even ordinary in-context conditioning on an irrelevant full document can itself perturb the model. We therefore use full-context interference as a reference:
$R^-_{\mathrm{full}}
=
\mathbb{E}_{q\sim\mathcal Q_x^-}
\mathfrak D\!\left(
p_\theta(\cdot\mid q)
\,\middle\|\,
p_\theta(\cdot\mid x,q)
\right).$
In particular, $R^-(\mathcal C_x)=0$ corresponds to perfect preservation of no-context
behavior, while $R^-_{\mathrm{full}}$ provides a natural reference
for the interference present without compression. A well-behaved reusable memory should therefore keep both $R^+$ and $R^-$ small.
At minimum, $R^-$ should not substantially exceed $R^-_{\mathrm{full}}$.
\paragraph{The Cartridges training objective is one-sided.}
\Cref{eq:cartridge-objective} directly supervises only document-grounded behavior. Self-study constructs $\mathcal D_x$ from traces derived from $x$, aligning training with $R^+$. Crucially, it does not include any term involving $\mathcal Q_x^-$ or the no-context reference $p_\theta(\cdot\mid q)$. As a result, $R^-$ is not explicitly constrained. We show that this leads to interference.\looseness-1

\vspace*{-0.15cm}
\section{Mitigating Memory Interference}
\label{mitig_strat}

The one-sided objective in~\Cref{background:two-sided} suggests two points of intervention: modifying how the Cartridge is trained or controlling when it is used. We consider mitigation strategies at both stages. Our training-time strategy explicitly encourages capability preservation during self-study, whereas our inference-time strategy removes the memory whenever it is predicted to be irrelevant. We also consider a lightweight prompting baseline that instructs the model to use the cached document only when relevant, reported separately in Appendix~\ref{app:prompting} as it does not eliminate interference.\looseness-1

\vspace*{-0.2cm}
\paragraph{\textsc{Cartridges}$_{\boldsymbol{++}}^{\mathbf{D}}$ ($\boldsymbol{C_{++}^{D}}$): training-time mitigation with data mixing.} To provide an explicit training signal for the missing $R^-$ supervision, in $C_{++}^{D}$ we augment self-study with examples from \texttt{Dolci-Instruct-SFT}~\citep{olmo2025olmo3}, which we use as a broad proxy for general-purpose, off-context queries $\mathcal Q_x^-$.
Let $\mathcal D^-$ denote the off-context pool and
$\gamma\in[0,1]$ the \emph{mixing fraction}. Interpreting each
corpus as an empirical distribution over supervised assistant-token
predictions, we write the target training mixture as
$\mathcal D_x^{(\gamma)}
= (1-\gamma)\mathcal D_x + \gamma\mathcal D^-$.
Thus, $\gamma$ specifies the fraction of supervised assistant tokens contributed by $\mathcal D^-$. Document-grounded examples retain the standard self-study teacher, whereas off-context examples use the frozen base model without the Cartridge as teacher, encouraging $p_\theta(\cdot\mid\mathcal C_x,q)$ to match the no-context behavior $p_\theta(\cdot\mid q)$ underlying $R^-$. The endpoints $\gamma=0$ and $\gamma=1$ recover standard self-study and off-context-only supervision, respectively.
Therefore, $C_{++}^{D}$ regularizes the learned memory without introducing a new loss or modifying the model weights $\theta$; $\gamma$ controls the trade-off between context fidelity and capability preservation.\looseness-1

\vspace*{-0.2cm}
\paragraph{\textsc{Cartridges}$_{\boldsymbol{++}}^{\mathbf{R}}$ ($\boldsymbol{C_{++}^{R}}$): inference-time mitigation with relevance routing.}

Mixed self-study asks a single Cartridge to preserve document-specific information while remaining neutral elsewhere. Relevance routing instead leaves the Cartridge unchanged and controls when it is active. Given an already-trained Cartridge, $C_{++}^{R}$ introduces a router $r_x(q)\in\{0,1\}$ that decides whether an incoming query should be served with or without it:\looseness-1
\vspace*{-0.1cm}
\begin{equation}
p_{\mathrm{route}}(y\mid q,\mathcal C_x)=
\begin{cases}
p_\theta(y\mid \mathcal C_x,q), & r_x(q)=1,\\
p_\theta(y\mid q), & r_x(q)=0.
\end{cases}
\label{eq:routing}
\vspace*{-0.1cm}
\end{equation}
When a query is rejected, the Cartridge is removed entirely, recovering the no-context behavior underlying $R^-$ without retraining either the memory or the language model. We use an embedding of the incoming query as the routing signal, extracted either from the evaluated LLM or from a smaller auxiliary encoder, and study different representations, pooling schemes, and classifiers.
We evaluate these design choices in Section~\ref{sec:implementation_mitigation} and Appendix~\ref{app:router}, and use a $k$-nearest-neighbor ($k$NN) router as our main variant for simplicity.
As in the training-time mitigation, \texttt{Dolci-Instruct-SFT} provides off-context negatives representing $\mathcal Q_x^-$, while positives come from a held-out subset of self-study. We calibrate the routing threshold on held-out negatives using one-sided conformal calibration~\citep{angelopoulos2024conformalrisk} at $\alpha=0.05$, then fix it for evaluation.
$C_{++}^{R}$ addresses the two requirements in~\Cref{background:two-sided} separately; it uses the Cartridge for relevant queries and otherwise recovers the no-context model, at the cost of a lightweight relevance decision before generation.

\section{Benchmarking Beyond Context Fidelity}
\label{sec:eval}

Building on the two-sided formulation in~\Cref{background:two-sided}, we define a benchmarking framework that operationalizes context fidelity and capability preservation independently of any particular dataset;~\Cref{exp_setup} describes its concrete instantiation. Existing evaluations largely capture the first requirement in~\Cref{background:two-sided} through \emph{context fidelity}. We propose a complementary benchmark for the second requirement, \emph{capability preservation}, by decomposing it into three off-context axes: \emph{general knowledge}, \emph{instruction following}, and \emph{context interference}. 
Together, these four axes operationalize $R^+$ and $R^-$ empirically.
For each document $x$, the compressed memory is constructed once and held fixed across all evaluation axes and subsequent queries, unless explicitly removed by routing. We compare each method against two reference conditions: \emph{Full context} serves the original document as text and provides the reference for context fidelity on $\mathcal Q_x^+$. \emph{No context} removes the document entirely and provides the reference for capability preservation on $\mathcal Q_x^-$. We ask whether a compression method can approach full-context performance on document-relevant queries and remain close to no-context behavior when the memory is irrelevant.\looseness-1

\vspace*{-0.2cm}
\paragraph{Context fidelity.}
We define it as the performance retained on queries from $\mathcal Q_x^+$, i.e., that require the source document. We use the task-native metric of each long-context benchmark. This provides our empirical measure of $R^+$ and captures any trade-off with capability preservation. For brevity, we refer to this axis as \emph{on-context} performance in the results.

\vspace*{-0.2cm}
\paragraph{General knowledge.}
We frame general-knowledge preservation as the ability to answer factual and reasoning questions unrelated to the stored document. 
We measure it on questions from $\mathcal Q_x^-$ relative to the no-context reference, to test whether the compressed memory degrades capabilities already present in the model.\looseness-1
\vspace*{-0.2cm}
\paragraph{Instruction following.}
We define instruction-following preservation as the ability to satisfy user constraints despite an irrelevant memory. We evaluate prompts with objectively verifiable requirements, capturing failures not reflected in answer accuracy.
\looseness-1
\vspace*{-0.2cm}
\paragraph{Context interference.}
Finally, we define \emph{context interference} as contamination of the rationale for a query unrelated to the long context. A response may be correct while still spuriously invoking the cached document, as illustrated in~\Cref{fig:overview} (see~\Cref{fig:contaminated-examples} for actual contaminated Cartridge generations). We therefore use an LLM judge, independent of answer correctness, to flag rationales that fabricate or attribute a quotation to the document, or claim that it supplies the answer. The final metric is the fraction of responses flagged by either behavior; merely mentioning or dismissing the document does not count as interference. Appendix~\ref{app:judge-rubric} details the judge and rubric.

\vspace{3pt}
\begingroup
\fontsize{9}{10}\selectfont
\setlength{\tabcolsep}{3pt}
\setlength{\arrayrulewidth}{0.3pt}
\setlength{\aboverulesep}{1pt}
\setlength{\belowrulesep}{1pt}
\renewcommand{\arraystretch}{1.08}
\captionsetup{font=small,skip=3pt,position=top}
\captionof{table}{Evaluation axes and benchmarks for context fidelity and capability preservation.}
\label{tab:evaluation-overview}
\noindent\begin{tabularx}{\linewidth}{@{}p{.29\linewidth}|*{3}{>{\raggedright\arraybackslash}X}@{}}
\toprule
\cellcolor{black!8}\centering\textbf{Context fidelity} ($\mathcal Q_x^+$) & \multicolumn{3}{c@{}}{\cellcolor{black!8}\textbf{Capability preservation} ($\mathcal Q_x^-$)}\\
\midrule
\textit{Long contexts} & {\fontsize{8.5}{10}\selectfont\textbf{General knowledge}} & {\fontsize{8.5}{10}\selectfont\textbf{Instruction following}} & {\fontsize{8.5}{10}\selectfont\textbf{Context interference}}\\
QASPER-16, MTOB,\newline LongHealth-10, QuALITY-20 & TinyMMLU,\newline MMLU-Pro & IFEval & LLM-as-judge\newline (MMLUs)\\
\bottomrule
\end{tabularx}\par
\endgroup
\vspace{-2pt}

\section{Experimental Setup}
\label{exp_setup}

\paragraph{Long-context datasets.}
We study four long-context datasets ranging from approximately 4K to 140K tokens: QASPER-16 (Q-16)~\citep{dasigi2021dataset} contains information-seeking questions grounded in NLP research papers; following~\citet{eyuboglu2025cartridges}, we merge 16 papers into one context. In LongHealth-10 (LH-10)~\citep{adams2025longhealth}, we concatenate ten fictional patient records into a single panel. MTOB~\citep{tanzer2024benchmark} is a structured-learning task for Kalamang--English translation. Finally, QuALITY-20~\citep{pang2022quality} consists of 20 narratives.
Table~\ref{tab:benchmarks-extended} reports context lengths and evaluation sizes.
Following the task metrics of~\citet{eyuboglu2025cartridges}, we report multiple-choice accuracy for LongHealth-10, averaged over eight independently seeded generations per question; corpus chrF for MTOB translation; and reference-answer log-perplexity for QASPER-16. For QuALITY-20, we report multiple-choice accuracy. See Appendix~\ref{app:datasets} for details on scoring and aggregation.\looseness-1
\vspace*{-0.2cm}
\paragraph{Capability preservation benchmarks.}
Table~\ref{tab:evaluation-overview} summarizes how we instantiate the capability-preservation axes in~\Cref{sec:eval}. We evaluate \emph{general knowledge} with TinyMMLU~\citep{hendryckstest2021,polo2024tinybenchmarks} and MMLU-Pro~\citep{wang2024mmlu}, and \emph{instruction following} with IFEval~\citep{zhou2023instructionfollowingevaluationlargelanguage}. Each query is paired with an unrelated cached document.
We judge \emph{context interference} in the general-knowledge rationales using the \texttt{gpt-oss} family~\citep{openai2025gptoss120bgptoss20bmodel}, with the 20B model for main results. Appendix~\ref{app:judges-agree} shows consistent findings with \texttt{gpt-oss-120b} and \texttt{gemma-3-27b-it}~\citep{gemma_2025}; prompts and scoring are detailed in Appendix~\ref{app:judge-rubric}.
Separately, we use \texttt{Dolci-Instruct-SFT}~\citep{olmo2025olmo3} as a broad proxy for $\mathcal Q_x^-$ when training and calibrating the mitigation strategies and never use it for off-context evaluation. See Appendix~\ref{app:datasets} for extended details.

\vspace*{-0.2cm}
\paragraph{Models and compression methods.}
We evaluate the CKV methods in different model families: \texttt{Qwen3-4B}~\citep{qwen3_2025} on QuALITY-20 and
\texttt{Qwen3-4B-Instruct-2507}~\citep{qwen3_2025} and
\texttt{Gemma-4-E4B-it}~\citep{gemma4_2026} on the longer QASPER-16,
LongHealth-10, and MTOB settings. For \texttt{Gemma}, compression factors apply
only to global-attention cache slots; sliding-window caches remain unchanged.
We compare Cartridges against Attention Matching (AM), KV cache eviction variants H2O, SnapKV, KVzip,
PyramidKV, StreamingLLM, and matched-length text summaries
(Appendices~\ref{app:impl-baselines} and~\ref{app:baseline-analyses}). We provide full- and no-context results as references. All compressed memories are fixed
before decoding. For Cartridges, we follow the canonical recipe of
\citet{eyuboglu2025cartridges} (Appendix~\ref{app:impl-cartridges}). Across datasets and model families we sweep compression $\rho\in[0.001,0.20]$, and share prompts, templates, and decoding
settings. Full details in Appendix~\ref{app:additional-details}.\looseness-1

\vspace*{-0.2cm}
\paragraph{\textsc{\textsc{Cartridges++}} implementation details.}
\label{sec:implementation_mitigation}
For data mixing ($C_{++}^{D}$), we evaluate $\gamma\in\{0.01,0.05,0.10\}$ throughout, and sweep the full range $\gamma\in[0,1]$ on a subset of benchmarks for a more detailed analysis. For routing ($C_{++}^{R}$), we construct and calibrate one non-parametric relevance router per document using pools of self-study and \texttt{Dolci} samples, respectively. Our main router uses the mean cosine similarity to the $k=10$ nearest self-study queries in a reference bank. Its activation threshold is calibrated at $\alpha=0.05$ using held-out \texttt{Dolci} negatives; no capability-preservation benchmark is used for construction or calibration. Full construction details and router variants for $C_{++}^{R}$ are provided in Appendix~\ref{app:router}, and more details on both mitigations in Appendix~\ref{app:mitigation}.

\paragraph{Compute cost of mitigation strategies.}
$C_{++}^{D}$ affects only offline Cartridge training and adds no inference-time cost; matched update counts keep its training cost comparable with standard Cartridges.
The $k$NN router in $C_{++}^{R}$ requires no classifier-weight training; alternatives with different computational costs are discussed in Appendix~\ref{app:router}.
For the hidden-state router, each request is first prefilled without the Cartridge to obtain its query representation. Rejected requests continue directly from this clean prefill, whereas accepted requests are re-prefilled with the Cartridge. Thus, routing adds a $k$NN lookup for rejected requests and one additional query prefill for accepted ones, without repeating the long-document prefill, which adds a modest overhead for short queries and long generations.

\section{Results}
\label{results}
\begin{figure}[h!]
\centering
\includegraphics[width=\linewidth]{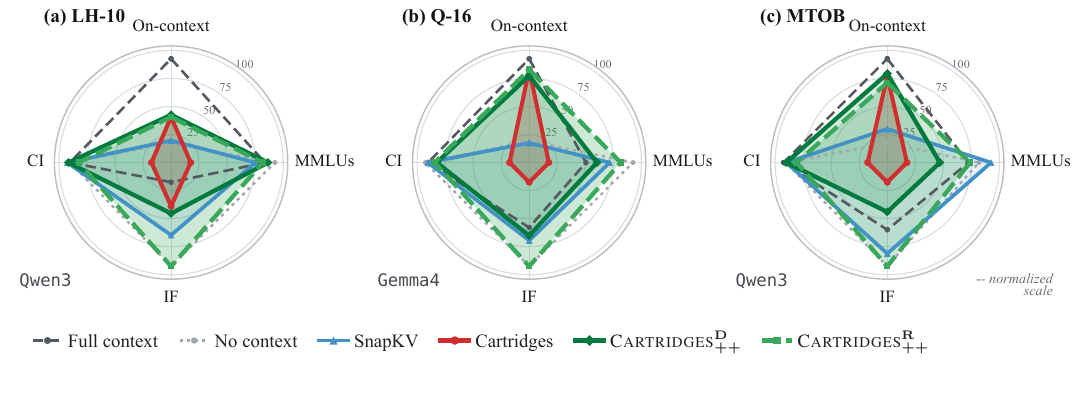}
\vspace*{-1.2cm}
\caption{\textit{\textsc{Cartridges++} results across model families.} Panels show LH-10
and MTOB with \texttt{Qwen3} at 2\% compression factor, and Q-16 with \texttt{Gemma-4}  at 5\%.
SnapKV is the eviction baseline in all three panels. Axes report on-context performance, MMLUs accuracy, instruction
following (IF), and context independence (CI; reversed), each
scaled separately across the displayed methods; outward is better. Mitigations use 5\% \texttt{Dolci} mixing and query $k$NN-10 routing.\looseness-1}
\label{fig:main-tradeoff}
\end{figure}

\begin{figure}[h!]
\centering
\vspace*{-0.1cm}
\includegraphics[width=\linewidth]{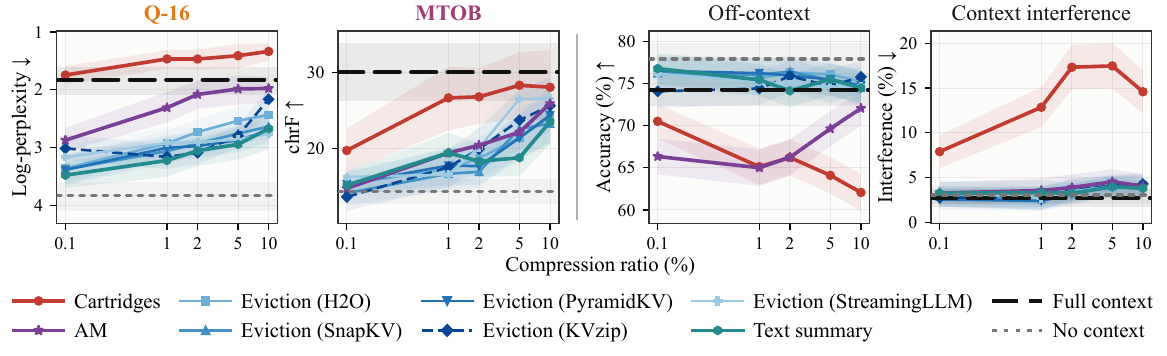}
\vspace*{-0.6cm}
\caption{\textit{CKV methods across compression factors on \texttt{Qwen3-4B-Instruct-2507}.} Left:
Q-16 log-perplexity (lower is better; note the reversed y axis) and MTOB chrF (higher is better). Right: off-context accuracy and context
interference, averaged equally over these two datasets. Shading shows 95\% intervals.\looseness-1}
\label{fig:compression-sweep}
\vspace*{-0.1cm}
\end{figure}

\vspace{-0.4cm}
\paragraph{Learned memories trade capability preservation for context fidelity.}
We first evaluate context fidelity and capability preservation of CKVs across
compression factors $\rho$. As shown in~\Cref{fig:compression-sweep}, Cartridges consistently
retain substantially stronger on-context performance than training-free eviction
methods, especially under extreme compression. This advantage reverses when their compressed KV becomes irrelevant: general
capabilities deteriorate (per-subject breakdown in Appendix~\ref{app:mmlu-categories}), and context contamination increases beyond the
interference induced by retaining the full document, the reference underlying $R^-_{\mathrm{full}}$. In contrast,
eviction methods, retaining native KV states, sacrifice context fidelity but remain substantially closer
to the no-context behavior off-context. Unless otherwise specified, \emph{off-context} performance averages general-knowledge accuracy (the mean of TinyMMLU and MMLU-Pro, denoted MMLUs) and IFEval accuracy with equal weight.\looseness-1

The same pattern holds across the remaining
long-context settings and individual capability-preservation axes
(Appendix~\ref{app:add-results}). These results expose a clear relevance
asymmetry: optimizing a memory for its source document can improve $R^+$ while
simultaneously worsening $R^-$.
Attention Matching (AM) occupies an intermediate regime between eviction and fully optimized memories, retaining selected native keys while fitting attention reconstruction. It generally preserves
off-context capabilities better, but Cartridges offer stronger context fidelity
on the longer tasks
(\Cref{fig:compression-sweep,fig:overview}), especially at smaller compression factors $\rho$.

\begin{figure}[h!]
\centering
\includegraphics[width=\linewidth]{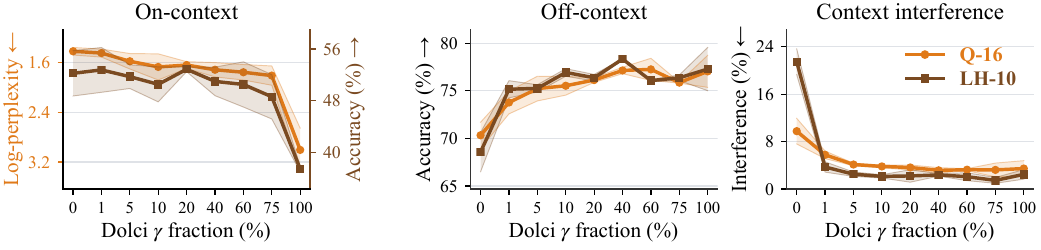}
\caption{\textit{$C_{++}^{D}$ performance, sweeping the \texttt{Dolci} mixing fraction on LH-10 and Q-16.} Panels show
on-context performance, off-context accuracy, and context
interference. 
Results are averaged across compression factors 0.1--10\%; datasets are shown in different colors. Shading shows 95\% normal intervals on \texttt{Qwen3-4B-Instruct-2507}.\looseness-1}
\label{fig:data-mixing}
\end{figure}

\vspace{-0.2cm}
\paragraph{Data mixing regularizes interference.}
$C_{++}^{D}$, with its mixed self-study, directly addresses the missing $R^-$ supervision identified in~\Cref{background:two-sided}. 
\Cref{fig:data-mixing} shows that even $\gamma=0.01$ substantially improves general knowledge and instruction following while reducing context interference. Low to moderate mixing preserves context fidelity, whereas aggressive mixing degrades on-context performance. We use $\gamma=0.05$ as a practical trade-off for $C_{++}^{D}$ in the remaining experiments. Appendix~\ref{app:data-mixing} further examines these effects as training progresses.\looseness-1

\vspace*{-0.4cm}

\begin{wrapfigure}[12]{r}{0.28\textwidth}
    \vspace{-1.2\baselineskip}
    \centering
    \includegraphics[width=\linewidth]{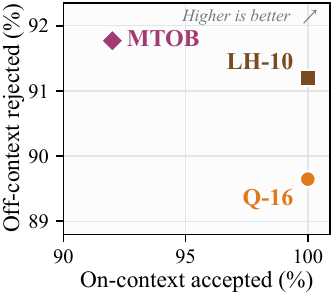}
    \vspace*{-0.4cm}
    \caption{\textit{$C_{++}^{R}$ routing rates.}}
    \label{fig:routing}
\end{wrapfigure}
\paragraph{Routing separates the two regimes.}
Unlike data mixing that optimizes a single compromise, $C_{++}^{R}$
addresses the relevance asymmetry by selecting between two inference states, keeping the Cartridge only when relevant. 
\Cref{fig:routing} measures routing accuracy of the $k$NN variant on the three long-context datasets with \texttt{Qwen3-4B-Instruct-2507}: the router achieves high accuracy, accepting over 90\% of on-context queries ($\mathcal Q_x^+$) and rejecting over 90\% of off-context ones ($\mathcal Q_x^-$). We examine whether this improves downstream performance in the next paragraph. %
Although \Cref{fig:routing} reports results using \texttt{Qwen3} hidden states, an external \texttt{MiniLM} encoder also achieves strong routing performance, which shows that post-hoc routing does not need to depend on the backbone model's representations (Appendix~\ref{app:router}; \Cref{fig:router-classifiers,fig:router-recipe-gemma}).
Overall, this mechanism is lightweight
and operates on an already-trained Cartridge; its main systems cost is that an
accepted request must be re-prefilled with the memory.\looseness-1 

\begin{figure}[h!]
    \centering
    \includegraphics[width=\linewidth]{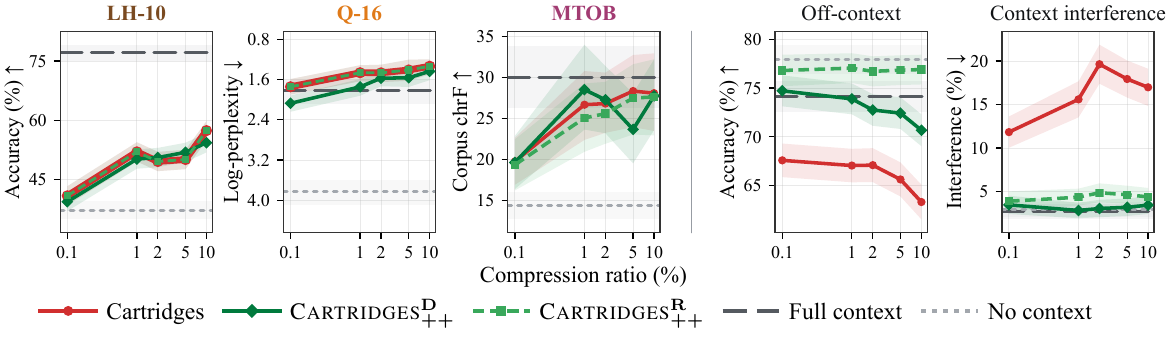}
    \vspace*{-0.6cm}
    \caption{\textit{\textsc{Cartridges}$_{\boldsymbol{++}}^{\mathbf{D}}$ ($\boldsymbol{C_{++}^{D}}$) and \textsc{Cartridges}$_{\boldsymbol{++}}^{\mathbf{R}}$ ($\boldsymbol{C_{++}^{R}}$) across compression factors.}
    Cartridges, $C_{\boldsymbol{++}}^{\mathbf{D}}$ 5\%, $C_{\boldsymbol{++}}^{\mathbf{R}}$ $k$NN-10 on \texttt{Qwen3-4B-Instruct-2507}.
    Left: on-context metrics. Right:
    off-context accuracy and interference averaged over all three
    datasets.}
    \label{fig:mitigation-five-panel}
\end{figure}

\vspace*{-0.2cm}
\paragraph{Interference is not an inevitable consequence of compression.}

\begin{wrapfigure}[18]{r}{0.38\textwidth}
    \vspace*{-1.5\baselineskip}
    \centering
    \includegraphics[width=\linewidth]{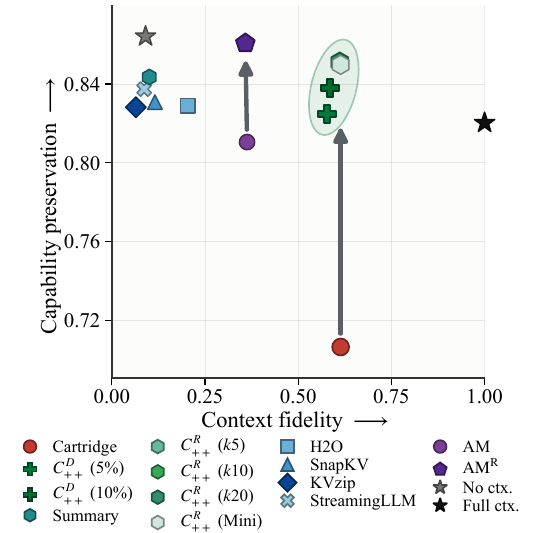}
    \vspace*{-0.6cm}
    \caption{\textit{\textsc{Cartridges++} results on \texttt{Gemma-4-E4B-it}.} LH-10/Q-16 across 1--10\% compression
    (composite defined in Appendix~\ref{app:per-benchmark}).}
    \label{fig:gemma-main-plane}
\end{wrapfigure}

\Cref{fig:mitigation-five-panel} compares $C_{++}^{D}$ and $C_{++}^{R}$ across the three long-context settings.
$C_{++}^{D}$ 5\% substantially reduces off-context degradation with a small or no on-context trade-off; gains already appear at 1\% (\Cref{fig:dolci-four-contexts}). Routing brings off-context accuracy close to the no-context reference while largely retaining context fidelity.
At a single compression factor, the radar plots of~\Cref{fig:main-tradeoff} show that Cartridges are strong on their source task but weak on the other axes; both mitigations recover a more balanced profile.
Aggregating results across contexts and compression factors, \Cref{fig:overview} (right) proposes a complementary view, showing that $C_{++}^{R}$ extends the Pareto frontier toward high context fidelity and capability preservation. 
Given AM's competitive on-context performance and milder off-context degradation, we also apply our router to AM ($\mathrm{AM}^{\mathrm{R}}$), which improves capability preservation but remains limited by AM's lower context fidelity; $C_{++}^{R}$ retains higher context fidelity with comparable capability preservation.\looseness-1

\paragraph{\textsc{Cartridges++} transfers across models and long-context variability.}
These gains transfer to \texttt{Gemma-4-E4B-it}, whose architecture interleaves global and sliding-window attention; Appendix~\ref{app:gemma-results} provides the full study. Its composite results in~\Cref{fig:gemma-main-plane} mirror the \texttt{Qwen3} comparison in~\Cref{fig:overview} (right): \textsc{Cartridges++} preserves Cartridges' strong context fidelity while substantially improving capability preservation. Both mitigations also remain effective with \texttt{Qwen3-4B} on QuALITY-20, whose contexts are roughly an order of magnitude shorter (Appendix~\ref{app:quality-results}).\looseness-1

\paragraph{Additional analyses support the same failure mode.}
We additionally find that interference is accompanied by a failure to abstain: Cartridges often answer as though the resident document contained evidence that is absent (Appendix~\ref{app:abstention}). To examine whether degradation compounds across turns, we study multi-turn interactions and find no additional collapse beyond the single-turn behavior. Both $C_{++}^{D}$ and $C_{++}^{R}$ retain their benefits, with routing evaluated using per-turn relevance labels (Appendix~\ref{app:multiturn}). Nor does longer training resolve the original degradation: off-context performance remains impaired, while data mixing retains its gains at later checkpoints (Appendix~\ref{app:training-duration}).\looseness-1
\WFclear %

\section{Conclusion}
\vspace*{-0.2cm}
Learned KV memories offer a compelling way to compress long contexts while
preserving strong context fidelity, but our results on Cartridges reveal an important hidden
cost: when the stored document is irrelevant, these memories can substantially
perturb the model's original behavior. This degradation spans general
knowledge, instruction following, and context interference.
Training-free eviction methods show the opposite trade-off, preserving capabilities better but sacrificing context fidelity. This relevance asymmetry is not captured by standard
compression evaluations.
We introduce \textsc{Cartridges++}, with two mitigation strategies that reduce this degradation while
retaining context fidelity across the model families and context lengths studied.
In $C_{++}^{D}$, off-context supervision during self-study improves capability preservation,
but at high mixing fractions can compete with document specialization. $C_{++}^{R}$ instead avoids forcing both behaviors into a single representation,
routing context-relevant queries to use the learned memory, and recovering no-context behavior on the others. More broadly, reusable memories should be evaluated
not only by the information they preserve, but also by the interference they
introduce when that information is irrelevant.\looseness-1
\vspace*{-0.2cm}
\paragraph{Limitations and future work.}
Our off-context evaluation focuses on TinyMMLU, MMLU-Pro, and IFEval, and further
work is needed to characterize additional failures, as suggested by our
abstention analysis (Appendix~\ref{app:abstention}). Our training-time and
inference-time mitigations both reduce interference, but each has trade-offs.
While data mixing regularizes the memory without adding inference cost, it can
reduce context fidelity and leave residual interference on requests poorly
covered by its training mixture. Routing instead preserves specialization
through selective activation, but depends on accurate relevance decisions and
requires re-prefilling accepted queries against the compact cache. Building on
the gains from these simple interventions, future work could combine the two
approaches and explore adaptive relevance estimation to build stronger mitigation strategies.\looseness-1

\newpage

\subsection*{AI use statement}

In this work, we used generative AI tools to implement methods, by assisting with the development and refinement of software code, and to support qualitative data analysis, through LLM judges that score model generations under a fixed rubric (Appendix~\ref{app:judge-rubric}). We generated synthetic self-study training traces with the model under study, following the Cartridges recipe~\citep{eyuboglu2025cartridges}, and used AI tools to support methodological checks. Additionally, we used generative AI tools to create and modify scientific figures, edit the manuscript to improve clarity and readability, and identify relevant literature. All AI-assisted outputs were subsequently reviewed by the authors: AI-assisted code was reviewed and tested for correctness, the judge analysis was replicated with two additional judges (Appendix~\ref{app:judge-rubric}), figures were checked against the underlying results, and suggested references were manually verified against the original sources. The authors take full responsibility for the final content of this work, including all text, claims, results, and artifacts produced with the assistance of generative AI.

\subsection*{Reproducibility statement}
All models, datasets, and benchmarks used in this work are publicly available, and we obtain them from their official releases; \Cref{exp_setup} lists them and Appendix~\ref{app:datasets} describes each benchmark in detail. For Cartridges and Attention Matching, we directly use the official implementations of \citet{eyuboglu2025cartridges} (\url{https://github.com/HazyResearch/cartridges}) and \citet{zweiger2026fast} (\url{https://github.com/adamzweiger/compaction}). The remaining baselines, together with the shared prompts, compression budgets, decoding settings, and position conventions of all methods, are described in Appendix~\ref{app:additional-details}. Both \textsc{Cartridges++} variants are specified in \Cref{sec:implementation_mitigation}, with complete training and routing details in Appendix~\ref{app:mitigation}. The evaluation protocol is defined in \Cref{sec:eval}, and the full judge rubric and scoring protocol are given in Appendix~\ref{app:judge-rubric}.

\bibliographystyle{iclr2027_conference}
\bibliography{iclr2027_conference}

\begin{thebibliography}{72}
\providecommand{\natexlab}[1]{#1}
\providecommand{\url}[1]{\texttt{#1}}
\expandafter\ifx\csname urlstyle\endcsname\relax
  \providecommand{\doi}[1]{doi: #1}\else
  \providecommand{\doi}{doi: \begingroup \urlstyle{rm}\Url}\fi

\bibitem[Adams et~al.(2025)Adams, Busch, Han, Excoffier, Ortala, L{\"o}ser, Aerts, Kather, Truhn, and Bressem]{adams2025longhealth}
Lisa Adams, Felix Busch, Tianyu Han, Jean-Baptiste Excoffier, Matthieu Ortala, Alexander L{\"o}ser, Hugo~JWL Aerts, Jakob~Nikolas Kather, Daniel Truhn, and Keno Bressem.
\newblock {LongHealth}: A question answering benchmark with long clinical documents.
\newblock \emph{Journal of Healthcare Informatics Research}, 9\penalty0 (3):\penalty0 280--296, 2025.

\bibitem[Agarwal et~al.(2025)Agarwal, Ahmad, Ai, Altman, Applebaum, Arbus, Arora, Bai, Baker, Bao, et~al.]{openai2025gptoss120bgptoss20bmodel}
Sandhini Agarwal, Lama Ahmad, Jason Ai, Sam Altman, Andy Applebaum, Edwin Arbus, Rahul~K Arora, Yu~Bai, Bowen Baker, Haiming Bao, et~al.
\newblock gpt-oss-120b \& gpt-oss-20b model card.
\newblock \emph{arXiv preprint arXiv:2508.10925}, 2025.

\bibitem[Ai et~al.(2026)Ai, He, and Guo]{ai2026evidence}
Mengting Ai, Jingrui He, and Yue Guo.
\newblock Does accuracy equal evidence? reasoning faithfulness under {KV} cache compression.
\newblock \emph{arXiv preprint arXiv:2608.01631}, 2026.

\bibitem[Angelopoulos et~al.(2024)Angelopoulos, Bates, Fisch, Lei, and Schuster]{angelopoulos2024conformalrisk}
Anastasios Angelopoulos, Stephen Bates, Adam Fisch, Lihua Lei, and Tal Schuster.
\newblock Conformal risk control.
\newblock In \emph{International conference on learning representations}, volume 2024, pp.\  55198--55218, 2024.

\bibitem[Asai et~al.(2024)Asai, Wu, Wang, Sil, and Hajishirzi]{asai2024selfrag}
Akari Asai, Zeqiu Wu, Yizhong Wang, Avirup Sil, and Hannaneh Hajishirzi.
\newblock {Self-RAG}: Learning to retrieve, generate, and critique through self-reflection.
\newblock In \emph{International Conference on Learning Representations}, 2024.
\newblock URL \url{https://openreview.net/forum?id=hSyW5go0v8}.

\bibitem[Bai et~al.(2024)Bai, Lv, Zhang, Lyu, Tang, Huang, Du, Liu, Zeng, Hou, Dong, Tang, and Li]{bai2024longbench}
Yushi Bai, Xin Lv, Jiajie Zhang, Hongchang Lyu, Jiankai Tang, Zhidian Huang, Zhengxiao Du, Xiao Liu, Aohan Zeng, Lei Hou, Yuxiao Dong, Jie Tang, and Juanzi Li.
\newblock {LongBench}: A bilingual, multitask benchmark for long context understanding.
\newblock In \emph{Proceedings of the 62nd Annual Meeting of the Association for Computational Linguistics}, pp.\  3119--3137, 2024.

\bibitem[Brandon et~al.(2024)Brandon, Mishra, Nrusimha, Panda, and Ragan-Kelley]{brandon2024reducing}
William Brandon, Mayank Mishra, Aniruddha Nrusimha, Rameswar Panda, and Jonathan Ragan-Kelley.
\newblock Reducing transformer key-value cache size with cross-layer attention.
\newblock In \emph{Advances in Neural Information Processing Systems}, volume~37, 2024.
\newblock URL \url{https://proceedings.neurips.cc/paper_files/paper/2024/hash/9e23d020c18e4c40d81c6a0fc7a46f68-Abstract-Conference.html}.

\bibitem[Caccia et~al.(2025)Caccia, Ansell, Ponti, Vuli{\'c}, and Sordoni]{caccia2025deepcontextdistillation}
Lucas Caccia, Alan Ansell, Edoardo Ponti, Ivan Vuli{\'c}, and Alessandro Sordoni.
\newblock Training plug-and-play knowledge modules with deep context distillation.
\newblock In \emph{Second Conference on Language Modeling}, 2025.
\newblock URL \url{https://openreview.net/forum?id=ghyyHZYORi}.

\bibitem[Cai et~al.(2025)Cai, Zhang, Gao, Liu, Li, Liu, Lu, Xiong, Dong, Hu, and Xiao]{cai2024pyramidkv}
Zefan Cai, Yichi Zhang, Bofei Gao, Yuliang Liu, Yucheng Li, Tianyu Liu, Keming Lu, Wayne Xiong, Yue Dong, Junjie Hu, and Wen Xiao.
\newblock {PyramidKV}: Dynamic {KV} cache compression based on pyramidal information funneling.
\newblock In \emph{Second Conference on Language Modeling}, 2025.
\newblock URL \url{https://openreview.net/forum?id=ayi7qezU87}.

\bibitem[Charakorn et~al.(2026)Charakorn, Cetin, Uesaka, and Lange]{charakorn2026doctolora}
Rujikorn Charakorn, Edoardo Cetin, Shinnosuke Uesaka, and Robert~Tjarko Lange.
\newblock {Doc-to-LoRA}: Learning to instantly internalize contexts.
\newblock \emph{arXiv preprint arXiv:2602.15902}, 2026.
\newblock URL \url{https://arxiv.org/abs/2602.15902}.

\bibitem[Chari et~al.(2025)Chari, Qin, and Van~Durme]{chari2025kvdistill}
Vivek Chari, Guanghui Qin, and Benjamin Van~Durme.
\newblock {KV-Distill}: Nearly lossless learnable context compression for {LLM}s.
\newblock \emph{arXiv preprint arXiv:2503.10337}, 2025.
\newblock URL \url{https://arxiv.org/abs/2503.10337}.

\bibitem[Chen et~al.(2026)Chen, Geh, Grover, Van~den Broeck, and Israel]{liu2025pitfalls}
Alex Chen, Renato Geh, Aditya Grover, Guy Van~den Broeck, and Daniel Israel.
\newblock The pitfalls of {KV} cache compression.
\newblock In \emph{Proceedings of the 64th Annual Meeting of the Association for Computational Linguistics (Volume 1: Long Papers)}, pp.\  41530--41553. Association for Computational Linguistics, 2026.
\newblock \doi{10.18653/v1/2026.acl-long.1926}.
\newblock URL \url{https://aclanthology.org/2026.acl-long.1926/}.

\bibitem[Chen et~al.(2024)Chen, Zhang, He, Li, Wang, Huang, and Xue]{chen2024recipe}
Qizhou Chen, Taolin Zhang, Xiaofeng He, Dongyang Li, Chengyu Wang, Longtao Huang, and Hui Xue.
\newblock Lifelong knowledge editing for {LLMs} with retrieval-augmented continuous prompt learning.
\newblock In \emph{Proceedings of the 2024 Conference on Empirical Methods in Natural Language Processing}, pp.\  13565--13580, 2024.
\newblock \doi{10.18653/v1/2024.emnlp-main.751}.
\newblock URL \url{https://aclanthology.org/2024.emnlp-main.751/}.

\bibitem[Cheng et~al.(2024)Cheng, Wang, Zhang, Ge, Chen, Wei, Zhang, and Zhao]{cheng2024xrag}
Xin Cheng, Xun Wang, Xingxing Zhang, Tao Ge, Si-Qing Chen, Furu Wei, Huishuai Zhang, and Dongyan Zhao.
\newblock {xRAG}: Extreme context compression for retrieval-augmented generation with one token.
\newblock In \emph{Advances in Neural Information Processing Systems}, volume~37, 2024.
\newblock URL \url{https://proceedings.neurips.cc/paper_files/paper/2024/hash/c5cf13bfd3762821ef7607e63ee90075-Abstract-Conference.html}.

\bibitem[Chevalier et~al.(2023)Chevalier, Wettig, Ajith, and Chen]{chevalier2023autocompressors}
Alexis Chevalier, Alexander Wettig, Anirudh Ajith, and Danqi Chen.
\newblock Adapting language models to compress contexts.
\newblock In \emph{Proceedings of the 2023 Conference on Empirical Methods in Natural Language Processing}, pp.\  3829--3846, 2023.

\bibitem[Dasigi et~al.(2021)Dasigi, Lo, Beltagy, Cohan, Smith, and Gardner]{dasigi2021dataset}
Pradeep Dasigi, Kyle Lo, Iz~Beltagy, Arman Cohan, Noah~A Smith, and Matt Gardner.
\newblock A dataset of information-seeking questions and answers anchored in research papers.
\newblock In \emph{Proceedings of the 2021 Conference of the North American Chapter of the Association for Computational Linguistics: Human Language Technologies}, pp.\  4599--4610, 2021.

\bibitem[{DeepSeek-AI} et~al.(2024{\natexlab{a}}){DeepSeek-AI}, Liu, Feng, Wang, Wang, Liu, Zhao, Deng, Ruan, Dai, Guo, et~al.]{liu2024deepseekv2}
{DeepSeek-AI}, Aixin Liu, Bei Feng, Bin Wang, Bingxuan Wang, Bo~Liu, Chenggang Zhao, Chengqi Deng, Chong Ruan, Damai Dai, Daya Guo, et~al.
\newblock {DeepSeek-V2}: A strong, economical, and efficient mixture-of-experts language model.
\newblock \emph{arXiv preprint arXiv:2405.04434}, 2024{\natexlab{a}}.

\bibitem[{DeepSeek-AI} et~al.(2024{\natexlab{b}}){DeepSeek-AI}, Liu, Feng, Xue, Wang, Wu, Lu, Zhao, Deng, Zhang, Ruan, et~al.]{liu2024deepseekv3}
{DeepSeek-AI}, Aixin Liu, Bei Feng, Bing Xue, Bingxuan Wang, Bochao Wu, Chengda Lu, Chenggang Zhao, Chengqi Deng, Chenyu Zhang, Chong Ruan, et~al.
\newblock {DeepSeek-V3} technical report.
\newblock \emph{arXiv preprint arXiv:2412.19437}, 2024{\natexlab{b}}.

\bibitem[Deng et~al.(2025)Deng, Zhang, Mao, Li, Huang, Yu, and Dou]{deng2025giststudy}
Chenlong Deng, Zhisong Zhang, Kelong Mao, Shuaiyi Li, Xinting Huang, Dong Yu, and Zhicheng Dou.
\newblock A silver bullet or a compromise for full attention? a comprehensive study of gist token-based context compression.
\newblock In \emph{Proceedings of the 63rd Annual Meeting of the Association for Computational Linguistics (Volume 1: Long Papers)}, pp.\  4861--4879. Association for Computational Linguistics, 2025.
\newblock \doi{10.18653/v1/2025.acl-long.241}.
\newblock URL \url{https://aclanthology.org/2025.acl-long.241/}.

\bibitem[Diao et~al.(2026)Diao, Li, Saidutta, Amballa, Valkov, and Chappidi]{diao2026doctoatom}
Xingjian Diao, Wenbo Li, Yashas~Malur Saidutta, Avinash Amballa, Lazar Valkov, and Srinivas Chappidi.
\newblock Doc-to-atom: Learning to compile and compose memory atoms.
\newblock \emph{arXiv preprint arXiv:2606.12400}, 2026.
\newblock URL \url{https://arxiv.org/abs/2606.12400}.

\bibitem[Diaz(2025)]{diaz2025learnedstructurecartridges}
Maurizio Diaz.
\newblock Learned structure in {CARTRIDGES}: Keys as shareable routers in self-studied representations.
\newblock \emph{arXiv preprint arXiv:2508.17032}, 2025.
\newblock URL \url{https://arxiv.org/abs/2508.17032}.

\bibitem[Eyuboglu et~al.(2026)Eyuboglu, Ehrlich, Arora, Guha, Zinsley, Liu, Rudra, Zou, Mirhoseini, and R{\'e}]{eyuboglu2025cartridges}
Sabri Eyuboglu, Ryan Ehrlich, Simran Arora, Neel Guha, Dylan Zinsley, Emily Liu, Atri Rudra, James Zou, Azalia Mirhoseini, and Christopher R{\'e}.
\newblock {Cartridges}: Lightweight and general-purpose long context representations via self-study.
\newblock In \emph{International Conference on Learning Representations}, 2026.
\newblock URL \url{https://proceedings.iclr.cc/paper_files/paper/2026/hash/4681359a7b1e94571598ad1adda35e6e-Abstract-Conference.html}.

\bibitem[Ge et~al.(2024)]{ge2024icae}
Tao Ge et~al.
\newblock In-context autoencoder for context compression in a large language model.
\newblock In \emph{International Conference on Learning Representations}, 2024.
\newblock URL \url{https://openreview.net/forum?id=uREj4ZuGJE}.

\bibitem[Gelberg et~al.(2026)]{gelberg2026kvcat}
Yoav Gelberg et~al.
\newblock Training transformers for {KV} cache compressibility.
\newblock \emph{arXiv preprint arXiv:2605.05971}, 2026.
\newblock URL \url{https://arxiv.org/abs/2605.05971}.

\bibitem[Hardalov et~al.(2026)Hardalov, Iglesias, and de~Gispert]{hardalov2026cartridges}
Momchil Hardalov, Gonzalo Iglesias, and Adri{\`a} de~Gispert.
\newblock Cartridges at scale: Training modular {KV} caches over large document collections.
\newblock \emph{arXiv preprint arXiv:2606.04557}, 2026.

\bibitem[Harrington et~al.(2026)]{harrington2026continuallearning}
Anne Harrington et~al.
\newblock When does continual learning require learning.
\newblock \emph{arXiv preprint arXiv:2607.07847}, 2026.
\newblock URL \url{https://arxiv.org/abs/2607.07847}.

\bibitem[Hartvigsen et~al.(2023)Hartvigsen, Sankaranarayanan, Palangi, Kim, and Ghassemi]{hartvigsen2023grace}
Thomas Hartvigsen, Swami Sankaranarayanan, Hamid Palangi, Yoon Kim, and Marzyeh Ghassemi.
\newblock Aging with {GRACE}: Lifelong model editing with discrete key-value adaptors.
\newblock In \emph{Advances in Neural Information Processing Systems}, volume~36, 2023.
\newblock URL \url{https://proceedings.neurips.cc/paper_files/paper/2023/hash/95b6e2ff961580e03c0a662a63a71812-Abstract.html}.

\bibitem[Hendrycks et~al.(2021)Hendrycks, Burns, Basart, Zou, Mazeika, Song, and Steinhardt]{hendryckstest2021}
Dan Hendrycks, Collin Burns, Steven Basart, Andy Zou, Mantas Mazeika, Dawn Song, and Jacob Steinhardt.
\newblock Measuring massive multitask language understanding.
\newblock \emph{Proceedings of the International Conference on Learning Representations (ICLR)}, 2021.

\bibitem[Hooper et~al.(2024)Hooper, Kim, Mohammadzadeh, Mahoney, Shao, Keutzer, and Gholami]{hooper2024kvquant}
Coleman Hooper, Sehoon Kim, Hiva Mohammadzadeh, Michael~W Mahoney, Yakun~S Shao, Kurt Keutzer, and Amir Gholami.
\newblock {KVQuant}: Towards 10 million context length {LLM} inference with {KV} cache quantization.
\newblock \emph{Advances in Neural Information Processing Systems}, 37:\penalty0 1270--1303, 2024.

\bibitem[Jiang et~al.(2023)Jiang, Wu, Lin, Yang, and Qiu]{jiang2023llmlingua}
Huiqiang Jiang, Qianhui Wu, Chin-Yew Lin, Yuqing Yang, and Lili Qiu.
\newblock {LLMLingua}: Compressing prompts for accelerated inference of large language models.
\newblock In \emph{Proceedings of the 2023 Conference on Empirical Methods in Natural Language Processing}, pp.\  13358--13376. Association for Computational Linguistics, 2023.
\newblock \doi{10.18653/v1/2023.emnlp-main.825}.
\newblock URL \url{https://aclanthology.org/2023.emnlp-main.825/}.

\bibitem[Joren et~al.(2025)Joren, Zhang, Ferng, Juan, Taly, and Rashtchian]{joren2025sufficientcontext}
Hailey Joren, Jianyi Zhang, Chun-Sung Ferng, Da-Cheng Juan, Ankur Taly, and Cyrus Rashtchian.
\newblock Sufficient context: A new lens on retrieval augmented generation systems.
\newblock In \emph{International Conference on Learning Representations}, 2025.
\newblock URL \url{https://openreview.net/forum?id=Jjr2Odj8DJ}.

\bibitem[Kim et~al.(2025)Kim, Kim, Kwon, Lee, Yun, and Song]{kim2025kvzip}
Jang-Hyun Kim, Jinuk Kim, Sangwoo Kwon, Jae~W. Lee, Sangdoo Yun, and Hyun~Oh Song.
\newblock {KVzip}: Query-agnostic {KV} cache compression with context reconstruction.
\newblock In \emph{Advances in Neural Information Processing Systems}, 2025.

\bibitem[Lewis et~al.(2020)Lewis, Perez, Piktus, Petroni, Karpukhin, Goyal, K{\"u}ttler, Lewis, Yih, Rockt{\"a}schel, Riedel, and Kiela]{lewis2020retrieval}
Patrick Lewis, Ethan Perez, Aleksandra Piktus, Fabio Petroni, Vladimir Karpukhin, Naman Goyal, Heinrich K{\"u}ttler, Mike Lewis, Wen-tau Yih, Tim Rockt{\"a}schel, Sebastian Riedel, and Douwe Kiela.
\newblock Retrieval-augmented generation for knowledge-intensive {NLP} tasks.
\newblock In \emph{Advances in Neural Information Processing Systems}, 2020.

\bibitem[Li \& Liang(2021)Li and Liang]{li2021prefixtuning}
Xiang~Lisa Li and Percy Liang.
\newblock Prefix-tuning: Optimizing continuous prompts for generation.
\newblock In \emph{Proceedings of the 59th Annual Meeting of the Association for Computational Linguistics and the 11th International Joint Conference on Natural Language Processing}, pp.\  4582--4597, 2021.
\newblock URL \url{https://aclanthology.org/2021.acl-long.353/}.

\bibitem[Li et~al.(2024)Li, Huang, Yang, Venkitesh, Locatelli, Ye, Cai, Lewis, and Chen]{li2024snapkv}
Yuhong Li, Yingbing Huang, Bowen Yang, Bharat Venkitesh, Acyr Locatelli, Hanchen Ye, Tianle Cai, Patrick Lewis, and Deming Chen.
\newblock {SnapKV}: {LLM} knows what you are looking for before generation.
\newblock In \emph{Advances in Neural Information Processing Systems}, 2024.

\bibitem[Li et~al.(2026)Li, Zhou, and Xu]{li2026latentcontextcompilation}
Zeju Li, Yizhou Zhou, and Qiang Xu.
\newblock Latent context compilation: Distilling long context into compact portable memory.
\newblock \emph{arXiv preprint arXiv:2602.21221}, 2026.
\newblock URL \url{https://arxiv.org/abs/2602.21221}.

\bibitem[Li et~al.(2025)Li, Su, and Collier]{li2025fivehundredcompressor}
Zongqian Li, Yixuan Su, and Nigel Collier.
\newblock {500xCompressor}: Generalized prompt compression for large language models.
\newblock In \emph{Proceedings of the 63rd Annual Meeting of the Association for Computational Linguistics}, 2025.
\newblock URL \url{https://aclanthology.org/2025.acl-long.1219/}.

\bibitem[Lin et~al.(2025)Lin, Zeng, Xiao, Kou, Hou, Gao, Zhang, and Deng]{lin2025matryoshkakv}
Bokai Lin, Zihao Zeng, Zipeng Xiao, Siqi Kou, Tianqi Hou, Xiaofeng Gao, Hao Zhang, and Zhijie Deng.
\newblock {MatryoshkaKV}: Adaptive {KV} compression via trainable orthogonal projection.
\newblock In \emph{International Conference on Learning Representations}, 2025.
\newblock URL \url{https://proceedings.iclr.cc/paper_files/paper/2025/hash/d7b351608d824a4680344a02b180a947-Abstract-Conference.html}.

\bibitem[Liu et~al.(2026)Liu, Tang, Chen, Dong, Li, Zhou, Li, Hu, and Chu]{wang2025semantic}
Xiang Liu, Zhenheng Tang, Hong Chen, Peijie Dong, Zeyu Li, Xiuze Zhou, Bo~Li, Xuming Hu, and Xiaowen Chu.
\newblock Semantic integrity matters: Benchmarking and preserving high-density reasoning in {KV} cache compression.
\newblock In \emph{Proceedings of the 43rd International Conference on Machine Learning}, volume 306 of \emph{Proceedings of Machine Learning Research}. PMLR, 2026.
\newblock URL \url{https://arxiv.org/abs/2502.01941v4}.

\bibitem[Liu et~al.(2024)Liu, Yuan, Jin, Zhong, Xu, Braverman, Chen, and Hu]{liu2024kivi}
Zirui Liu, Jiayi Yuan, Hongye Jin, Shaochen Zhong, Zhaozhuo Xu, Vladimir Braverman, Beidi Chen, and Xia Hu.
\newblock {KIVI}: A tuning-free asymmetric 2bit quantization for {KV} cache.
\newblock In Ruslan Salakhutdinov, Zico Kolter, Katherine Heller, Adrian Weller, Nuria Oliver, Jonathan Scarlett, and Felix Berkenkamp (eds.), \emph{Proceedings of the 41st International Conference on Machine Learning}, volume 235 of \emph{Proceedings of Machine Learning Research}, pp.\  32332--32344. PMLR, 21--27 Jul 2024.
\newblock URL \url{https://proceedings.mlr.press/v235/liu24bz.html}.

\bibitem[Luo et~al.(2026)Luo, Li, Dai, Chen, Zheng, and Qiu]{luo2025zerorag}
Qi~Luo, Xiaonan Li, Junqi Dai, Shuang Chen, Yining Zheng, and Xipeng Qiu.
\newblock Zero-rag: towards retrieval-augmented generation with zero redundant knowledge.
\newblock \emph{Frontiers of Computer Science}, 20\penalty0 (10):\penalty0 2010372, 2026.

\bibitem[Maia~Polo et~al.(2024)Maia~Polo, Weber, Choshen, Sun, Xu, and Yurochkin]{polo2024tinybenchmarks}
Felipe Maia~Polo, Lucas Weber, Leshem Choshen, Yuekai Sun, Gongjun Xu, and Mikhail Yurochkin.
\newblock {tinyBenchmarks}: evaluating {LLM}s with fewer examples.
\newblock In \emph{Proceedings of the 41st International Conference on Machine Learning}, volume 235 of \emph{Proceedings of Machine Learning Research}, pp.\  34303--34326. PMLR, 2024.
\newblock URL \url{https://proceedings.mlr.press/v235/maia-polo24a.html}.

\bibitem[Mitchell et~al.(2022)Mitchell, Lin, Bosselut, Manning, and Finn]{mitchell2022serac}
Eric Mitchell, Charles Lin, Antoine Bosselut, Christopher~D. Manning, and Chelsea Finn.
\newblock Memory-based model editing at scale.
\newblock In \emph{Proceedings of the 39th International Conference on Machine Learning}, volume 162 of \emph{Proceedings of Machine Learning Research}, pp.\  15817--15831. PMLR, 2022.
\newblock URL \url{https://proceedings.mlr.press/v162/mitchell22a.html}.

\bibitem[Monteiro et~al.(2026)Monteiro, Klein, Ablin, and Cuturi]{monteiro2026nectar}
Jo{\~a}o Monteiro, Michal Klein, Pierre Ablin, and Marco Cuturi.
\newblock Nectar: Neural estimation of cached-token attention via regression.
\newblock \emph{arXiv preprint arXiv:2605.09778}, 2026.
\newblock URL \url{https://arxiv.org/abs/2605.09778}.

\bibitem[Moschella et~al.(2026)Moschella, Manduchi, and Sener]{moschella2026learningtoevict}
Luca Moschella, Laura Manduchi, and Ozan Sener.
\newblock Learning to evict from key-value cache.
\newblock In \emph{Proceedings of the 43rd International Conference on Machine Learning}, 2026.
\newblock URL \url{https://arxiv.org/abs/2602.10238v2}.

\bibitem[Mu et~al.(2023)Mu, Li, and Goodman]{mu2023gist}
Jesse Mu, Xiang~Lisa Li, and Noah Goodman.
\newblock Learning to compress prompts with gist tokens.
\newblock In \emph{Advances in Neural Information Processing Systems}, 2023.

\bibitem[Muennighoff(2022)]{muennighoff2022sgpt}
Niklas Muennighoff.
\newblock {SGPT}: {GPT} sentence embeddings for semantic search.
\newblock \emph{arXiv preprint arXiv:2202.08904}, 2022.
\newblock URL \url{https://arxiv.org/abs/2202.08904}.

\bibitem[Pang et~al.(2022)Pang, Parrish, Joshi, Nangia, Phang, Chen, Padmakumar, Ma, Thompson, He, et~al.]{pang2022quality}
Richard~Yuanzhe Pang, Alicia Parrish, Nitish Joshi, Nikita Nangia, Jason Phang, Angelica Chen, Vishakh Padmakumar, Johnny Ma, Jana Thompson, He~He, et~al.
\newblock {QuALITY}: Question answering with long input texts, yes!
\newblock In \emph{Proceedings of the 2022 Conference of the North American Chapter of the Association for Computational Linguistics: Human Language Technologies}, pp.\  5336--5358, 2022.

\bibitem[Petrov et~al.(2025)]{petrov2025gistpool}
Aleksandar Petrov et~al.
\newblock Long context in-context compression by getting to the gist of gisting.
\newblock \emph{arXiv preprint arXiv:2504.08934}, 2025.
\newblock URL \url{https://arxiv.org/abs/2504.08934}.

\bibitem[Popovi{\'c}(2015)]{popovic2015chrf}
Maja Popovi{\'c}.
\newblock {chrF}: Character $n$-gram {F}-score for automatic {MT} evaluation.
\newblock In \emph{Proceedings of the Tenth Workshop on Statistical Machine Translation}, pp.\  392--395, 2015.
\newblock \doi{10.18653/v1/W15-3049}.

\bibitem[Reimers \& Gurevych(2019)Reimers and Gurevych]{reimers2019sentencebert}
Nils Reimers and Iryna Gurevych.
\newblock Sentence-{BERT}: Sentence embeddings using siamese {BERT}-networks.
\newblock In \emph{Proceedings of the 2019 Conference on Empirical Methods in Natural Language Processing and the 9th International Joint Conference on Natural Language Processing}, pp.\  3982--3992, 2019.

\bibitem[Shepard(2026)]{shepard2026precomputedmemorycosts}
Asa Shepard.
\newblock What it costs to compose, rebuild, and correct precomputed memory.
\newblock \emph{arXiv preprint arXiv:2608.30647}, 2026.
\newblock URL \url{https://arxiv.org/abs/2608.30647}.

\bibitem[Shi et~al.(2023)Shi, Chen, Misra, Scales, Dohan, Chi, Sch{\"a}rli, and Zhou]{shi2023large}
Freda Shi, Xinyun Chen, Kanishka Misra, Nathan Scales, David Dohan, Ed~H Chi, Nathanael Sch{\"a}rli, and Denny Zhou.
\newblock Large language models can be easily distracted by irrelevant context.
\newblock In \emph{International conference on machine learning}, pp.\  31210--31227. PMLR, 2023.

\bibitem[Tanzer et~al.(2024)Tanzer, Suzgun, Visser, Jurafsky, and Melas-Kyriazi]{tanzer2024benchmark}
Garrett Tanzer, Mirac Suzgun, Eline Visser, Dan Jurafsky, and Luke Melas-Kyriazi.
\newblock A benchmark for learning to translate a new language from one grammar book.
\newblock In \emph{International Conference on Learning Representations}, 2024.
\newblock URL \url{https://proceedings.iclr.cc/paper_files/paper/2024/hash/52d63f9e4b81f866bf69fb3c834aad47-Abstract-Conference.html}.

\bibitem[Tarasov et~al.(2026)Tarasov, Lashukov, Goncharova, and Kuznetsov]{tarasov2026progressivecramming}
Dmitrii Tarasov, Timofei Lashukov, Elizaveta Goncharova, and Andrey Kuznetsov.
\newblock Progressive cramming: Reliable token compression and what it reveals.
\newblock \emph{arXiv preprint arXiv:2607.21231}, 2026.
\newblock URL \url{https://arxiv.org/abs/2607.21231}.

\bibitem[Team et~al.(2025)Team, Kamath, Ferret, Pathak, Vieillard, Merhej, Perrin, Matejovicova, Ram{\'e}, Rivi{\`e}re, et~al.]{gemma_2025}
Gemma Team, Aishwarya Kamath, Johan Ferret, Shreya Pathak, Nino Vieillard, Ramona Merhej, Sarah Perrin, Tatiana Matejovicova, Alexandre Ram{\'e}, Morgane Rivi{\`e}re, et~al.
\newblock Gemma 3 technical report.
\newblock \emph{arXiv preprint arXiv:2503.19786}, 2025.

\bibitem[Team et~al.(2026)Team, Abd, Aggarwal, Algayres, Andreev, Bachem, Ballantyne, Brick, C{\u{a}}rbune, Casbon, et~al.]{gemma4_2026}
Gemma Team, Sherif~El Abd, Vaibhav Aggarwal, Robin Algayres, Alek Andreev, Olivier Bachem, Ian Ballantyne, Cormac Brick, Victor C{\u{a}}rbune, Michelle Casbon, et~al.
\newblock Gemma 4 technical report.
\newblock \emph{arXiv preprint arXiv:2607.02770}, 2026.

\bibitem[{Team Olmo} et~al.(2025){Team Olmo}, Ettinger, Bertsch, Kuehl, Graham, Heineman, Groeneveld, Brahman, Timbers, Ivison, Morrison, Poznanski, Lo, Soldaini, Jordan, Chen, Noukhovitch, Lambert, Walsh, Dasigi, Berry, Malik, Shah, Geng, Arora, Gupta, Anderson, Xiao, Murray, Romero, Graf, Asai, Bhagia, Wettig, Liu, Rangapur, Anastasiades, Huang, Schwenk, Trivedi, Magnusson, Lochner, Liu, Miranda, Sap, Morgan, Schmitz, Guerquin, Wilson, Huff, Bras, Xin, Shao, Skjonsberg, Shen, Li, Wilde, Pyatkin, Merrill, Chang, Gu, Zeng, Sabharwal, Zettlemoyer, Koh, Farhadi, Smith, and Hajishirzi]{olmo2025olmo3}
{Team Olmo}, Allyson Ettinger, Amanda Bertsch, Bailey Kuehl, David Graham, David Heineman, Dirk Groeneveld, Faeze Brahman, Finbarr Timbers, Hamish Ivison, Jacob Morrison, Jake Poznanski, Kyle Lo, Luca Soldaini, Matt Jordan, Mayee Chen, Michael Noukhovitch, Nathan Lambert, Pete Walsh, Pradeep Dasigi, Robert Berry, Saumya Malik, Saurabh Shah, Scott Geng, Shane Arora, Shashank Gupta, Taira Anderson, Teng Xiao, Tyler Murray, Tyler Romero, Victoria Graf, Akari Asai, Akshita Bhagia, Alexander Wettig, Alisa Liu, Aman Rangapur, Chloe Anastasiades, Costa Huang, Dustin Schwenk, Harsh Trivedi, Ian Magnusson, Jaron Lochner, Jiacheng Liu, Lester James~V. Miranda, Maarten Sap, Malia Morgan, Michael Schmitz, Michal Guerquin, Michael Wilson, Regan Huff, Ronan~Le Bras, Rui Xin, Rulin Shao, Sam Skjonsberg, Shannon~Zejiang Shen, Shuyue~Stella Li, Tucker Wilde, Valentina Pyatkin, Will Merrill, Yapei Chang, Yuling Gu, Zhiyuan Zeng, Ashish Sabharwal, Luke Zettlemoyer, Pang~Wei Koh, Ali Farhadi, Noah~A. Smith, and Hannaneh Hajishirzi.
\newblock Olmo 3, 2025.
\newblock URL \url{https://arxiv.org/abs/2512.13961}.

\bibitem[Wang et~al.(2024{\natexlab{a}})Wang, Li, Zhang, Xu, Yao, Jiang, Xie, Huang, and Chen]{wang2024wise}
Peng Wang, Zexi Li, Ningyu Zhang, Ziwen Xu, Yunzhi Yao, Yong Jiang, Pengjun Xie, Fei Huang, and Huajun Chen.
\newblock {WISE}: Rethinking the knowledge memory for lifelong model editing of large language models.
\newblock In \emph{Advances in Neural Information Processing Systems}, volume~37, 2024{\natexlab{a}}.
\newblock URL \url{https://proceedings.neurips.cc/paper_files/paper/2024/hash/60960ad78868fce5c165295fbd895060-Abstract-Conference.html}.

\bibitem[Wang et~al.(2020)Wang, Wei, Dong, Bao, Yang, and Zhou]{wang2020minilm}
Wenhui Wang, Furu Wei, Li~Dong, Hangbo Bao, Nan Yang, and Ming Zhou.
\newblock {MiniLM}: Deep self-attention distillation for task-agnostic compression of pre-trained transformers.
\newblock In \emph{Advances in Neural Information Processing Systems}, volume~33, pp.\  5776--5788, 2020.

\bibitem[Wang et~al.(2024{\natexlab{b}})Wang, Ma, Zhang, Ni, Chandra, Guo, Ren, Arulraj, He, Jiang, et~al.]{wang2024mmlu}
Yubo Wang, Xueguang Ma, Ge~Zhang, Yuansheng Ni, Abhranil Chandra, Shiguang Guo, Weiming Ren, Aaran Arulraj, Xuan He, Ziyan Jiang, et~al.
\newblock {MMLU-Pro}: A more robust and challenging multi-task language understanding benchmark.
\newblock \emph{Advances in Neural Information Processing Systems}, 37:\penalty0 95266--95290, 2024{\natexlab{b}}.

\bibitem[Xiao et~al.(2024)Xiao, Tian, Chen, Han, and Lewis]{xiao2024streaming}
Guangxuan Xiao, Yuandong Tian, Beidi Chen, Song Han, and Mike Lewis.
\newblock Efficient streaming language models with attention sinks.
\newblock In \emph{International Conference on Learning Representations}, 2024.

\bibitem[Xu et~al.(2024)Xu, Shi, and Choi]{xu2023recomp}
Fangyuan Xu, Weijia Shi, and Eunsol Choi.
\newblock {RECOMP}: Improving retrieval-augmented {LM}s with context compression and selective augmentation.
\newblock In \emph{International Conference on Learning Representations}, 2024.
\newblock URL \url{https://openreview.net/forum?id=mlJLVigNHp}.

\bibitem[Yang et~al.(2025)Yang, Li, Yang, Zhang, Hui, Zheng, Yu, Gao, Huang, Lv, et~al.]{qwen3_2025}
An~Yang, Anfeng Li, Baosong Yang, Beichen Zhang, Binyuan Hui, Bo~Zheng, Bowen Yu, Chang Gao, Chengen Huang, Chenxu Lv, et~al.
\newblock {Qwen3} technical report.
\newblock \emph{arXiv preprint arXiv:2505.09388}, 2025.

\bibitem[Yang et~al.(2024)Yang, Han, Gao, Hu, Zhang, and Zhao]{yang2024pyramidinfer}
Dongjie Yang, Xiaodong Han, Yan Gao, Yao Hu, Shilin Zhang, and Hai Zhao.
\newblock {PyramidInfer}: Pyramid {KV} cache compression for high-throughput {LLM} inference.
\newblock In \emph{Findings of the Association for Computational Linguistics: ACL 2024}, 2024.

\bibitem[Yoran et~al.(2024)Yoran, Wolfson, Ram, and Berant]{yoran2024irrelevant}
Ori Yoran, Tomer Wolfson, Ori Ram, and Jonathan Berant.
\newblock Making retrieval-augmented language models robust to irrelevant context.
\newblock In \emph{International Conference on Learning Representations}, 2024.

\bibitem[Zhang et~al.(2026)Zhang, Guo, Sun, Zhang, Hao, Lin, Zhang, Zhao, Shen, Tang, Xu, Yan, Wang, Chen, Xiong, Li, and Chua]{zhang2026metis}
Zeyu Zhang, Ziliang Guo, Yihang Sun, Xichong Zhang, Xixuan Hao, Zehao Lin, Yang Zhang, Xiaoyan Zhao, Tong Shen, Bo~Tang, Zhi-Qin~John Xu, Junchi Yan, Haofen Wang, Xu~Chen, Feiyu Xiong, Zhiyu Li, and Tat-Seng Chua.
\newblock Metis: Memory foundation model.
\newblock \emph{arXiv preprint arXiv:2607.26760}, 2026.
\newblock URL \url{https://arxiv.org/abs/2607.26760}.

\bibitem[Zhang et~al.(2023)Zhang, Sheng, Zhou, Chen, Zheng, Cai, Song, Tian, R{\'e}, Barrett, Wang, and Chen]{zhang2023h2o}
Zhenyu Zhang, Ying Sheng, Tianyi Zhou, Tianlong Chen, Lianmin Zheng, Ruisi Cai, Zhao Song, Yuandong Tian, Christopher R{\'e}, Clark Barrett, Zhangyang Wang, and Beidi Chen.
\newblock {H2O}: Heavy-hitter oracle for efficient generative inference of large language models.
\newblock In \emph{Advances in Neural Information Processing Systems}, 2023.

\bibitem[Zhao et~al.(2026)Zhao, Chen, Tang, Ma, Hu, Hu, Iacob, Mehrotra, and Lane]{zhao2026adapt}
Wanru Zhao, Yihong Chen, Yuzhi Tang, Wentao Ma, Shengchao Hu, Shell~Xu Hu, Alex Iacob, Abhinav Mehrotra, and Nicholas~D. Lane.
\newblock Rethinking data curation in {LLM} training: Online reweighting offers better generalization than offline methods.
\newblock In \emph{International Conference on Learning Representations}, 2026.
\newblock URL \url{https://proceedings.iclr.cc/paper_files/paper/2026/hash/b77e87be2a7caca996da3c79191e2e88-Abstract-Conference.html}.

\bibitem[Zheng et~al.(2026)]{zheng2026latentmemorymanagement}
Ziyang Zheng et~al.
\newblock Context distillation as latent memory management.
\newblock \emph{arXiv preprint arXiv:2605.28889}, 2026.
\newblock URL \url{https://arxiv.org/abs/2605.28889}.

\bibitem[Zhou et~al.(2023)Zhou, Lu, Mishra, Brahma, Basu, Luan, Zhou, and Hou]{zhou2023instructionfollowingevaluationlargelanguage}
Jeffrey Zhou, Tianjian Lu, Swaroop Mishra, Siddhartha Brahma, Sujoy Basu, Yi~Luan, Denny Zhou, and Le~Hou.
\newblock Instruction-following evaluation for large language models, 2023.
\newblock URL \url{https://arxiv.org/abs/2311.07911}.

\bibitem[Zweiger et~al.(2026)Zweiger, Fu, Guo, and Kim]{zweiger2026fast}
Adam Zweiger, Xinghong Fu, Han Guo, and Yoon Kim.
\newblock Fast {KV} compaction via attention matching.
\newblock In \emph{Proceedings of the 43rd International Conference on Machine Learning}, volume 306 of \emph{Proceedings of Machine Learning Research}. PMLR, 2026.
\newblock URL \url{https://arxiv.org/abs/2602.16284v2}.

\end{thebibliography}

\newpage
\appendix

\section{Additional Related Work}
\label{app:additional-related-work}

\paragraph{Other forms of KV-cache reduction.}
Beyond selecting a shorter sequence of cached states, other techniques reduce
cache precision or dimensionality. These include KV quantization
~\citep{hooper2024kvquant,liu2024kivi}, trainable projection such as
MatryoshkaKV~\citep{lin2025matryoshkakv}, multi-head latent attention
~\citep{liu2024deepseekv2,liu2024deepseekv3}, and cross-layer KV sharing
~\citep{brandon2024reducing}. KV-CAT trains the host Transformer to tolerate
later cache sparsification~\citep{gelberg2026kvcat}, whereas Learning to Evict
trains an eviction policy while leaving the LLM fixed; its final cache still
consists of selected native states~\citep{moschella2026learningtoevict}. These are
complementary efficiency directions outside our matched comparison of native
and optimized states at the same sequence budget.\looseness-1

\paragraph{Amortized learned compressors.}
Amortized approaches learn a compressor once and apply it to previously unseen
contexts. Gist tokens and AutoCompressors fine-tune models to produce and
consume soft summaries~\citep{mu2023gist,chevalier2023autocompressors},
while ICAE trains a LoRA encoder~\citep{ge2024icae} and 500xCompressor
trains a shared encoder for KV compression~\citep{li2025fivehundredcompressor}.
xRAG maps reusable dense document embeddings into an LLM's representation
space~\citep{cheng2024xrag}, while KV-Distill trains LoRA adapters on query
projections so selected tokens aggregate earlier information into a shorter KV
cache~\citep{chari2025kvdistill}.
These methods learn a reusable encoding rule rather than optimizing a new
memory per document; we study a fixed, document-specific CKV on a frozen model.

\paragraph{Document-specific optimized and parameter memories.}
Cartridges build on continuous prefix tuning~\citep{li2021prefixtuning}, but
optimize internal keys and values rather than input embeddings. Their offline
construction cost can be amortized when the same document is queried
repeatedly. Mechanistic work suggests different roles for their learned keys
and values: keys provide relatively transferable routing structure, whereas
values carry more document-specific content
~\citep{diaz2025learnedstructurecartridges}.
Latent Context Compilation constructs KV memories through a temporary LoRA
compiler, rather than directly optimizing cache tensors
~\citep{li2026latentcontextcompilation}.
Neighboring document-internalization methods serve parameter modules rather
than KV prefixes. Deep Context Distillation trains plug-and-play document
modules~\citep{caccia2025deepcontextdistillation}; Doc-to-LoRA and Doc-to-Atom
compile documents into compact adapters
~\citep{charakorn2026doctolora,diao2026doctoatom}. These methods serve a different
representation; we compare eviction of native states with learned KV states.

\paragraph{Broader reliability of learned memories.}
Reliability analyses of learned bottlenecks find that gist compression can fail
at boundaries, on unexpected information, and during long-context information
propagation~\citep{deng2025giststudy,petrov2025gistpool}.
New learned context-compression methods are also emerging concurrently with
this work~\citep{monteiro2026nectar}, but likewise do not study off-context
behavior.
Other work studies the broader lifecycle of learned
memories: precomputed memories may be difficult to compose, rebuild, or correct
~\citep{shepard2026precomputedmemorycosts}, while continual-learning experiments
include Cartridge configurations with comparatively stable retention
~\citep{harrington2026continuallearning}. Optimized input embeddings, adapters,
and compiled KV memories can also perturb behavior when source information is
absent, replaced, or unnecessary~\citep{charakorn2026doctolora,
tarasov2026progressivecramming,li2026latentcontextcompilation}. Metis measures
degradation after irrelevant information is stored in an evolving memory
architecture~\citep{zhang2026metis}. These studies motivate capability preservation;
we isolate interference from directly optimized KV states, even when they
preserve context fidelity under aggressive compression.

\paragraph{Selective activation and locality.}
Related work has studied mechanisms for limiting the effect of learned memories
when they are irrelevant, but not of KV states. RECIPE combines prompt gating with a KL locality loss
that preserves the unprompted model's behavior on unrelated
queries~\citep{chen2024recipe}. Doc-to-Atom selectively activates parameter
memories~\citep{diao2026doctoatom}, while Context Distillation as Latent Memory
Management retrieves LoRA memories and uses first-token entropy to decide
whether to keep a memory active or fall back to the base
model~\citep{zheng2026latentmemorymanagement}. Cartridges at Scale instead uses
mixed-visibility training with distractor Cartridges to support multi-memory
composition~\citep{hardalov2026cartridges}. Our data-mixing intervention uses a
no-context teacher rather than retaining a relevant Cartridge, while our router
applies selective activation post hoc to an already-trained CKV; rejected
requests use the same frozen LLM without the Cartridge.

\newpage
\section{Extended Experimental Setting}
\label{app:additional-details}

\subsection{Extended benchmark details}
\label{app:datasets}

\Cref{tab:benchmarks-extended} summarizes context lengths and evaluation sizes.
\begin{table}[h]
\centering
\caption{Benchmarks with context lengths and evaluation sizes. Context lengths
use the \texttt{Qwen3} tokenizer; for QuALITY-20 we give the range over its 20
articles. LongHealth-10 scores eight seeded generations per question. Context
interference is judged on the TinyMMLU and MMLU-Pro rationales. Off-context
sets are evaluated once per memory, i.e., once per article for QuALITY-20.}
\label{tab:benchmarks-extended}
\small
\setlength{\tabcolsep}{3pt}
\begin{tabular*}{\linewidth}{@{\extracolsep{\fill}}lrrlrlrlr@{}}
\toprule
\multicolumn{3}{c}{\textbf{Context fidelity} (on-context)} &
\multicolumn{6}{c}{\textbf{Capability preservation} (off-context)} \\
\cmidrule(r){1-3}\cmidrule(l){4-9}
\multicolumn{3}{l}{\textit{Long-context datasets}} &
\multicolumn{2}{l}{\textit{General knowledge}} &
\multicolumn{2}{l}{\textit{Instruction following}} &
\multicolumn{2}{l}{\textit{Context interference}} \\
& Tokens & \#Eval & & \#Eval & & \#Eval & & \#Eval \\
\midrule
QASPER-16     & 103,280       & 78          & TinyMMLU & 100   & IFEval & 541 & LLM-as-judge & 1,500 \\
LongHealth-10 & 113,634       & 200$\times$8 & MMLU-Pro & 1,400 &        &     & (MMLUs)      &       \\
MTOB          & 138,435       & 50          &          &       &        &     &              &       \\
QuALITY-20    & 4,162--7,323  & 357         &          &       &        &     &              &       \\
\bottomrule
\end{tabular*}
\end{table}

\paragraph{Long-context benchmarks.}
We evaluate four complementary settings. \textbf{QuALITY-20}~\citep{pang2022quality}
is a fixed subset of 20 articles from the QuALITY development split, with 357
multiple-choice questions in total. We use article IDs \texttt{52845},
\texttt{30029}, \texttt{62139}, \texttt{63523}, \texttt{63401},
\texttt{62476}, \texttt{63041}, \texttt{30035}, \texttt{61285},
\texttt{62261}, \texttt{62314}, \texttt{61430}, \texttt{52855},
\texttt{62085}, \texttt{62498}, \texttt{61119}, \texttt{63616},
\texttt{61467}, \texttt{60412}, and \texttt{63855}. Each article is
approximately 8K tokens, and we construct and evaluate one memory per
article. We generate one response per question, parse its selected option
(A--D), and report multiple-choice accuracy pooled over all 357 questions,
giving each question equal weight. This shorter-context setting tests whether off-context interference
persists even when eviction methods match or exceed Cartridges on the source
task.
Following the setup of~\citet{eyuboglu2025cartridges}, \textbf{LongHealth-10}
concatenates the records of fictional patients 1--10 into one approximately
113K-token document. Its 200 five-way questions are evaluated with eight seeded
generations per question. We parse the selected option (A–E) and report mean accuracy across eight generations per question. 
\textbf{QASPER-16}~\citep{dasigi2021dataset} merges 16 scientific
papers into a fixed 103,280-token context and contains 78 questions. Its primary
metric is the corpus token-weighted negative log-likelihood of the reference
answers (lower is better), matching the original
Cartridges implementation. For completeness, we additionally report generated-answer token F1 in
\Cref{app:qasper-f1}. The 78 questions and their frozen \texttt{GPT-4.1}-rewritten
reference answers are exactly the evaluation bank released by the original
authors; we neither
resample nor regenerate them.
\textbf{MTOB}~\citep{tanzer2024benchmark}, also used in the original Cartridges
study, tests whether a model can learn Kalamang--English translation from a
grammar book, a bilingual lexicon, and 375 parallel examples. We use the
current full-book construction: one 138,435-token document and the 50 official
Kalamang-to-English test sentences. Predictions are greedy and scored with
corpus chrF, a character $n$-gram F-score~\citep{popovic2015chrf}, using the
same \texttt{sacreBLEU} implementation as Cartridges. This score pools $n$-gram
counts across all 50 translations; higher is better. Thus, the
three primary long-context benchmarks cover approximately 103--138K tokens,
while QuALITY probes whether the same behavior already appears at $\sim$8K
tokens.\looseness-1

\paragraph{Capability preservation benchmarks.}
Every memory is paired with the same frozen questions, all unrelated to its
source document. TinyMMLU is the fixed 100-question representative subset of
MMLU spanning its 57 subjects~\citep{hendryckstest2021,polo2024tinybenchmarks}.
Our MMLU-Pro subset contains 1,400 questions: 100 deterministic examples from
each of its 14 categories~\citep{wang2024mmlu}. Both are evaluated zero-shot
with greedy decoding. The model generates a rationale and terminal answer
choice, which is parsed against the available choices (four for TinyMMLU and up
to ten for MMLU-Pro); unparsable outputs are incorrect. We report the two
accuracies with equal benchmark weight, and use their rationales for the
interference measure in \Cref{app:judge-rubric}. For instruction following, we
use all 541 IFEval prompts and its official deterministic strict prompt-level
pass criterion~\citep{zhou2023instructionfollowingevaluationlargelanguage}.

\paragraph{Capability preservation training data.}
We use the full \texttt{Dolci-Instruct-SFT} mixture~\citep{olmo2025olmo3}, rather than
selecting one of its constituent datasets. We retain only clean, single-turn
conversations comprising one non-empty user message followed by one non-empty
assistant response; multi-turn, tool-use, and system or environment transcripts
are excluded. We discard examples longer than 2,048 tokens and
deduplicate by stable source identity. From this filtered subset, we assign
examples deterministically to train, validation, and test splits in a
90\%/5\%/5\% ratio using seed 0. 
For \textsc{Cartridges++} data mixing, we fix one deterministic ordering of the \texttt{Dolci}
training examples and take as many examples from the start as each mixing rate
$\gamma$ requires. For $0\leq\gamma<1$, a mixture with a fixed self-study corpus
of $S$ supervised assistant tokens targets $\gamma S/(1-\gamma)$ \texttt{Dolci} tokens.
Different mixing rates use different amounts of off-context data
while retaining the same self-study corpus.
Every smaller mixture is therefore contained in each larger one,
making the sweep directly comparable. The $\gamma=1$ endpoint uses only
\texttt{Dolci}, excluding self-study examples from optimization. The routing variant instead pairs each
context's first 2,000 unique self-study
questions (positives) with the same ordered 2,000 non-test \texttt{Dolci} prompts
(negatives). For router fitting, the positive examples are split
deterministically 85\%/15\% into training and held-out partitions; the \texttt{Dolci}
negatives retain their split assignments from the construction above.
The $k$NN reference bank and fitted classifier variants use only the training partitions. The held-out data are
used to select logistic-regression regularization by AUROC and to calibrate the
threshold from \texttt{Dolci}-negative scores. TinyMMLU, MMLU-Pro, and IFEval are never
used for fitting, model selection, or calibration.

\paragraph{Models and context windows.}
We use \texttt{Qwen3-4B}~\citep{qwen3_2025} for QuALITY, whose
$\sim$8K-token documents fit within its native 32,768-token context window.
For QASPER, LongHealth, and MTOB, whose contexts approach or exceed 100K
tokens, we use \texttt{Qwen3-4B-Instruct-2507}~\citep{qwen3_2025}, which
supports a native 262,144-token context window, and
\texttt{Gemma-4-E4B-it}~\citep{gemma4_2026}.
We do not extend \texttt{Qwen3-4B} beyond its native window using manual
RoPE/YaRN scaling, avoiding an additional configuration difference across
tasks. Where a model exposes a thinking mode, we disable it throughout
self-study, memory optimization, and evaluation.

\subsection{Implementation details: \textsc{Cartridges++}}
\label{app:impl-cartridges}

\paragraph{Cartridges.}
We follow the self-study and context-distillation procedure of \citet{eyuboglu2025cartridges}. The frozen base model repeatedly samples a
short document chunk (512--4,096 tokens; QuALITY uses 512--1,024) and generates
single-turn question--answer pairs conditioned on that chunk. 
We follow the original recipe for QuALITY, LongHealth, and QASPER, where samples are drawn uniformly from five prompt families: question,
structuring, summarization, use-case, and creative prompts. For MTOB, we use a task-aligned mixture
of translation, question, and summarization prompts.
The chunk-conditioned model also supplies the token-level teacher
distribution used to train the Cartridge while all model weights remain
frozen. We add a post-processing step and reject malformed or
incompletely scored rows and exact-deduplicate question--answer pairs before
training. Self-study generation is sampled with the default per-model parameters.

For \texttt{Qwen}, the learned Cartridges are optimized with Adam at learning rate
$0.02$, packed length 2,048, and one pass over the self-study corpus, following the
original implementation. LongHealth, QASPER, and MTOB use global batch size 32;
QuALITY uses global batch size 6. After cleaning and deduplication, each
QuALITY article has 32,000 training examples (640,000 total), LongHealth and
QASPER each have 130,872, and MTOB has 139,739 (\Cref{tab:self-study}). Unless stated otherwise, we evaluate the
checkpoint at the one-pass compute boundary. All experiments in this manuscript ran on NVIDIA
H100 or B200 GPUs.

\begin{table}[h]
\centering
\caption{Number of self-study question--answer pairs per dataset after cleaning
and deduplication. QuALITY-20 has 32,000 per article.}
\label{tab:self-study}
\small
\begin{tabular}{lrrrr}
\toprule
& QASPER-16 & LongHealth-10 & MTOB & QuALITY-20 \\
\midrule
Self-study Q/As & 130,872 & 130,872 & 139,739 & 640,000 \\
\bottomrule
\end{tabular}
\end{table}

For \texttt{Gemma}, we use \texttt{google/gemma-4-E4B-it}. Its hybrid architecture has
42 logical attention layers backed by 24 physical cache slots; only
full-attention slots 5, 11, 17, and 23 are compacted. We leave every
sliding-window slot uncompressed with its native local KV cache; \texttt{Gemma}
compression factors therefore apply only to the four global-attention slots.
Because the original Cartridges paper does not specify a recipe for this
architecture, we sweep Adam learning rates at 10\% compression and select
$5\times10^{-3}$ using held-out validation self-study loss. Packing length
2,048, global batch size 32, and the self-study corpus size remain unchanged. For both \texttt{Qwen} and \texttt{Gemma}, a
$p$-token Cartridge is initialized from the key and value states of the first
$p$ document tokens, and its first key--value pair is frozen as an attention
sink, following \citet{eyuboglu2025cartridges}; position and sink conventions
are detailed in Appendix~\ref{app:positions}.\looseness-1

\paragraph{\textsc{Cartridges++} with data mixing.}
For a requested \texttt{Dolci} token fraction $0<\gamma<1$, we select the smallest prefix of the
frozen \texttt{Dolci} pool whose supervised assistant-token count attains
$\gamma/(1-\gamma)$ times the self-study token count. At $\gamma=1$, we use the
full frozen \texttt{Dolci} training pool and exclude self-study rows from optimization.
For mixed training, document and \texttt{Dolci} rows
are then jointly shuffled before packing. Consequently, $\gamma$ is a
corpus-level supervised-token fraction, not a constraint imposed separately on
every mini-batch. 
We score each \texttt{Dolci} response with the frozen base model without the document,
so its teacher distribution represents the model's no-context behavior.
Self-study examples retain the standard chunk-conditioned teacher distribution;
the same Cartridge therefore receives both document-grounded and general-instruction training signals.\looseness-1

We compare every mixture at the same number of completed optimizer
updates as the corresponding $\gamma=0$ Cartridge, so improvements cannot be
attributed to additional optimization. Larger fractions consume more distinct
\texttt{Dolci} examples but not more updates.
We report $\gamma\in\{0.01,0.05,0.10\}$ for all four contexts. The wider
LongHealth and QASPER sweep in \Cref{fig:data-mixing} additionally shows
how the on-context/off-context trade-off changes as $\gamma$ increases.

\paragraph{\textsc{Cartridges++} with relevance routing.}
Each context receives a binary relevance classifier. We use the self-study
positive and \texttt{Dolci}-negative pools and deterministic splits defined above. We
compare logistic regression (LR) and cosine $k$-nearest-neighbor classifiers with
$k\in\{5,10,20\}$. Hidden-state features are extracted without the Cartridge resident (we refer to them as \texttt{query}), at residual
layers 18, 24, and the final layer. For question-token states
$h_1,\ldots,h_n$, we compare the last token, the arithmetic mean, and a
recency-weighted ramp $h_{\mathrm{ramp}}= \frac{\sum_{i=1}^{n} i\,h_i}{\sum_{i=1}^{n} i}$.
This is the position-weighted mean-pooling rule used by
\texttt{SGPT}~\citep{muennighoff2022sgpt}: it retains information from the entire
question while assigning greater weight to later tokens.
We also evaluate frozen
\texttt{sentence-transformers/all-MiniLM-L6-v2} embeddings: a six-layer
\texttt{MiniLM} encoder~\citep{wang2020minilm}, used through a sentence-transformer
mean-pooling head~\citep{reimers2019sentencebert}; embeddings are
$\ell_2$-normalized.

The decision threshold is calibrated exclusively on held-out \texttt{Dolci} negatives,
which serve as a general off-context calibration population and avoid leaking
any TinyMMLU, MMLU-Pro, or IFEval examples. With
$N$ such scores, we use their
$\lceil(N+1)(1-\alpha)\rceil$-th order statistic at $\alpha=0.05$ and activate
the Cartridge at or above this threshold. Calibration targets false activation
on negatives exchangeable with the held-out \texttt{Dolci} population; it does
not guarantee 5\% false activation on other benchmarks or control rejection of
relevant queries. We therefore report empirical
positive retention and off-context rejection on the untouched benchmark
queries in \Cref{fig:router-classifiers}.
The final classifier comparison in \Cref{fig:router-classifiers}
fixes \texttt{query\_mean@L18} across query-state classifiers and datasets. Most L18/L24
mean- or ramp-pooled hidden-state variants perform similarly on the English QA
contexts, although MTOB is more representation-sensitive. \texttt{MiniLM} is a
competitive low-cost alternative on its $k$NN variants. The
main routed results use the fixed cosine $k$NN-10 recipe in
\Cref{alg:router}; \Cref{fig:router-classifiers} additionally reports the other classifier variants.\looseness-1

\begin{algorithm}[t]
\caption{Constructing, calibrating, and applying the relevance router for document $x$.}
\label{alg:router}
\begin{algorithmic}[1]
\Require Held-out self-study questions $\mathcal P_x$ with router-training
partition $\mathcal P_x^{\mathrm{train}}\subset\mathcal P_x$, held-out
\texttt{Dolci} queries $\mathcal N_{\mathrm{cal}}$, Cartridge $\mathcal C_x$,
error level $\alpha=0.05$
\State Clean-prefill each $q\in\mathcal P_x^{\mathrm{train}}\cup\mathcal N_{\mathrm{cal}}$
without $\mathcal C_x$ and extract its layer-18 question-token states
\State Mean-pool and $\ell_2$-normalize the states to obtain $z(q)$
\State Store the positive bank $\mathcal B_x=\{z(p):p\in\mathcal
P_x^{\mathrm{train}}\}$ and set $k=10$
\State Define $s_x(q)$ as the mean cosine similarity of $z(q)$ to its $k$
nearest vectors in $\mathcal B_x$
\State Sort the $N=|\mathcal N_{\mathrm{cal}}|$ scores
$\{s_x(q):q\in\mathcal N_{\mathrm{cal}}\}$ in ascending order and set $\tau_x$
to the $\lceil(N+1)(1-\alpha)\rceil$-th
\For{each incoming query $q$}
    \State Clean-prefill $q$ without $\mathcal C_x$ and compute $s_x(q)$
    \If{$s_x(q)<\tau_x$} \Comment{$r_x(q)=0$ in \Cref{eq:routing}}
        \State Continue generation from the clean prefill
    \Else \Comment{$r_x(q)=1$}
        \State Re-prefill $q$ with $\mathcal C_x$ and generate from the
        Cartridge-conditioned model
    \EndIf
\EndFor
\end{algorithmic}
\end{algorithm}

\newpage
\subsection{Implementation details: baselines}
\label{app:impl-baselines}

All methods receive the identical tokenized document, target compression budget,
question prompt, and decoding settings. We study \emph{reusable} memories: each
compressed state is built once, frozen before any evaluation question is
observed, and reused across questions.\looseness-1

\paragraph{Eviction methods.}

\textbf{SnapKV}~\citep{li2024snapkv} scores keys from an observation window
at the end of the complete prompt, which ordinarily includes the downstream
request. To obtain a reusable, query-independent memory, our canonical
adaptation instead uses query vectors from all document tokens during the
one-time prefill. This query source is the only algorithmic change: following
the original method, we average scores per attention head, apply seven-token
max pooling, and allocate a uniform budget to each head.
We also evaluate an offline variant using generated self-study questions.
Both query sources yield similar retention curves (\Cref{fig:eviction-query-source});
we use document-prefill queries to cover the full source without selecting a
representative synthetic question.
\textbf{PyramidKV}~\citep{cai2024pyramidkv} originally uses trailing prompt queries for
key selection. We apply the same offline context-prefill adaptation as for
SnapKV, retaining seven-token max pooling and the beta-20 pyramidal allocation
across layers.
\textbf{H2O}~\citep{zhang2023h2o} originally maintains a dynamic cache of recent tokens
and heavy hitters, updating each token's accumulated attention score during
generation and evicting low-scoring entries as new tokens arrive. Our reusable
memory must be fixed before any evaluation question is observed. As for SnapKV and PyramidKV, we
compute the heavy-hitter score once from attention accumulated during the full
document prefill, retain the highest-scoring token positions under the
target budget, and freeze that cache for all subsequent questions; there are no
query- or decode-time score updates. The fixed self-study variant of all three follows the implementation by~\citet{zweiger2026fast}.
\textbf{KVzip}~\citep{kim2025kvzip}, in contrast, is query-independent by design, so no
query-source adaptation is needed. Following the original method, we score
positions by offline context reconstruction in its reference 2,000-token
chunks and use a fixed total budget in each layer with non-uniform allocation
across KV heads and positions. Each context position is scored once, by the
chunk that reconstructs it, while all preceding context remains in the
attention-softmax denominator. We then fill the layer budget with a single
top-$k$ selection over all heads and positions in the complete context.
\textbf{StreamingLLM}~\citep{xiao2024streaming} is used without an algorithmic
adaptation. Following the four-sink-token configuration used in the original
work, it retains the first four attention-sink positions and the most recent
$B-4$ positions for a target budget of $B$ tokens.

\paragraph{Attention Matching.}
For AM~\citep{zweiger2026fast}, we follow the released
implementation with self-study and repeat-prefill queries, on-policy
recomputation of query states, and highest-attention-key selection. The
self-study queries use five equally weighted prompt families: repetition,
summarization, aggregation, structured-JSON extraction, and three-question
generation. AM fits per-head biases and values to reproduce the full cache's
attention outputs; for later layers, reference states are recomputed using the
cache already compacted at earlier layers. We retain the published per-layer
and per-head settings. The
released setup operates on substantially smaller selection domains and uses
chunked execution for larger inputs; accordingly, we compact each QuALITY
article as a single domain, but divide LongHealth, QASPER, and MTOB into four
contiguous, approximately equal-length token chunks. AM is fitted independently
within each chunk at the requested compression factor, and the four compacted
caches are concatenated in their original order. This keeps the total nominal
budget unchanged while avoiding a single ill-conditioned solve over the entire
long context.\looseness-1

\paragraph{Text summaries.}
Finally, the text-space baseline asks the evaluated model to generate a
budget-conditioned replacement
for the document, capped at the same maximal token budget given by the compression factor. Its prompt selection
protocol and ablation are detailed in \Cref{fig:text-summary-prompts,fig:text-summary-prompt-ablation}.
\subsection{Position and sink conventions}
\label{app:positions}

\begin{figure*}[h!]
    \centering
    \includegraphics[width=\textwidth]{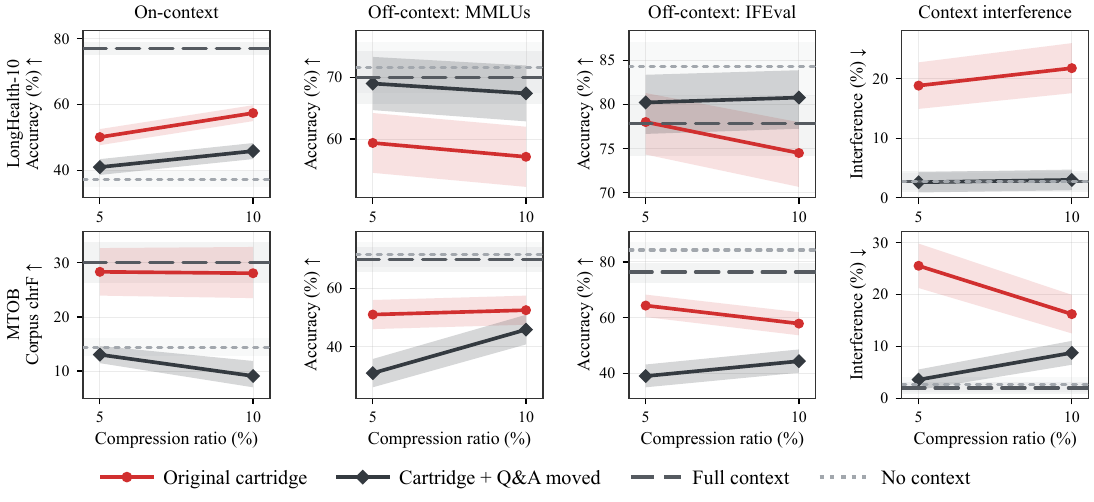}
    \caption{\textit{Query-position ablation at 5\% and 10\% compression.}
    Cartridge keys remain at the prefix; the query and answer begin either
    immediately after the Cartridge or at the original context end. Rows show
    LongHealth-10 and MTOB; columns show on-context performance, MMLUs accuracy,
    IFEval, and interference. Shading shows 95\% intervals.\looseness-1}
    \label{fig:position-ablation}
\end{figure*}

Position handling follows each method's native inference rule. A Cartridge of
length $m$ occupies positions $[0,m)$ and the query begins at $m$. An evicted
cache instead preserves every retained state's original rotary position and
places the query at the uncompressed context length $L$. Eviction therefore
preserves source-to-query distances, whereas a Cartridge deliberately forms a
new compact prefix. During Cartridge
optimization, only the first KV position is frozen, as in the original
Cartridges procedure.
We isolate this positional effect in \Cref{fig:position-ablation} by moving
only the query and generated answer to
the end of the original context position range while leaving the Cartridge at
$[0,m)$. This tests whether the positional convention, rather than the learned
states, causes the interference we study. It does not: moving the query and answer
lowers interference only together with most of the Cartridge's on-context
gain.\looseness-1

\newpage
\subsection{Judge rubric and scoring protocol}
\label{app:judge-rubric}

We judge only whether an off-context rationale is influenced by the irrelevant
resident document, which we call \emph{context interference}; the judge is not
asked to grade answer correctness. Its input contains the question, answer
choices, generated rationale, and a dataset-specific document-type tag, but
excludes the gold answer, the model's parsed answer identity, and the document
text (\Cref{fig:judge-prompt}; \Cref{fig:judge-prompt-full} shows a shortened
prompt). At temperature zero, one context-independence call returns a structured
JSON object from which we use ordered observations, a short justification, the Boolean
indicators \texttt{fabricates\_quote} and
\texttt{context\_answers\_question}, and an ordinal independence score. The
score is a direct rubric judgment rather than an arithmetic function of the
flags: 5 means that the response ignores or correctly dismisses the irrelevant
document; 4 permits a harmless mention; 3 is mixed; 2 relies primarily on the
document; and 1 derives the answer from it, including through a fabricated
quotation or entity bridge. We report this ordinal diagnostic in
\Cref{fig:judge-independence}.

Our primary context-interference statistic is the union of these two Boolean
indicators: a response counts as interfered if \texttt{fabricates\_quote} or
\texttt{context\_answers\_question} is true. Merely mentioning or dismissing
the document does not count as interference. A response whose verdict cannot be
parsed counts as not interfered, so every method is scored over the same fixed
question population. We compute this rate separately on
TinyMMLU and MMLU-Pro and average the two rates with equal benchmark weight, so
the 1,400-question benchmark does not dominate the 100-question benchmark. We
use this event-based rate in the main paper because it is directly
interpretable. The ordinal score is retained as an appendix diagnostic (\Cref{fig:judge-independence}).
Main results use \texttt{gpt-oss-20b}. We repeat every complete shared cell
with \texttt{gpt-oss-120b} and \texttt{gemma-3-27b-it}. 
As shown in
\Cref{app:judges-agree}, the three judges agree on the separation between
Cartridges and eviction baselines, despite small differences in absolute rates.

\begin{figure}[H]
\centering
\begingroup
\tcbset{promptcard/.style={enhanced, boxrule=.45pt, arc=2mm,
  left=2mm, right=2mm, top=1.3mm, bottom=1.3mm,
  fonttitle=\bfseries\footnotesize, fontupper=\scriptsize,
  colback=white, colframe=black!25, coltitle=black}}
\begin{tcolorbox}[promptcard,
  title={Shared instruction and rubric},
  colback=black!4, colbacktitle=black!18, colframe=black!55,
  fontupper=\footnotesize]
Judge whether the rationale answers from the MCQ and world knowledge or instead
uses the irrelevant resident document. Do not grade answer correctness. Apply
the same five-point anchors and Boolean definitions to every context.
\end{tcolorbox}

\noindent\begin{minipage}[t]{.485\linewidth}
\begin{tcolorbox}[promptcard,title={QuALITY-20},height=31mm,valign=center,
  colbacktitle=black!9, colframe=black!30]
\texttt{<document\_kind>}\\
a long-form document\\
\texttt{</document\_kind>}\\[1mm]
\texttt{<mcq\_question>} $\cdots$\quad
\texttt{<mcq\_options>} $\cdots$\\
\texttt{<model\_rationale>} $\cdots$
\end{tcolorbox}
\begin{tcolorbox}[promptcard,title={QASPER-16},height=31mm,valign=center,
  colbacktitle=black!9, colframe=black!30]
\texttt{<document\_kind>}\\
a merged collection of scientific papers\\
\texttt{</document\_kind>}\\[1mm]
\texttt{<mcq\_question>} $\cdots$\quad
\texttt{<mcq\_options>} $\cdots$\\
\texttt{<model\_rationale>} $\cdots$
\end{tcolorbox}
\end{minipage}\hfill
\begin{minipage}[t]{.485\linewidth}
\begin{tcolorbox}[promptcard,title={LongHealth-10},height=31mm,valign=center,
  colbacktitle=black!9, colframe=black!30]
\texttt{<document\_kind>}\\
a patient clinical record\\
\texttt{</document\_kind>}\\[1mm]
\texttt{<mcq\_question>} $\cdots$\quad
\texttt{<mcq\_options>} $\cdots$\\
\texttt{<model\_rationale>} $\cdots$
\end{tcolorbox}
\begin{tcolorbox}[promptcard,title={MTOB},height=31mm,valign=center,
  colbacktitle=black!9, colframe=black!30]
\texttt{<document\_kind>}\\
the full MTOB Kalamang reference context\\
\texttt{</document\_kind>}\\[1mm]
\texttt{<mcq\_question>} $\cdots$\quad
\texttt{<mcq\_options>} $\cdots$\\
\texttt{<model\_rationale>} $\cdots$
\end{tcolorbox}
\end{minipage}

\begin{tcolorbox}[promptcard,
  title={Structured output fields used in our analysis},
  colback=black!4, colbacktitle=black!18, colframe=black!55]
\texttt{observations} $\rightarrow$
\texttt{fabricates\_quote},
\texttt{context\_answers\_question} $\rightarrow$
\texttt{reason} $\rightarrow$ \texttt{score} $\in\{1,\ldots,5\}$
\end{tcolorbox}
\endgroup
\caption{\textit{Context-independence judge prompt and analyzed output fields.} The
document-type tag varies by resident context. Inputs contain the question,
choices, and rationale, but omit the document, gold answer, and parsed
prediction. A shortened prompt example appears in
\Cref{fig:judge-prompt-full}.}
\label{fig:judge-prompt}
\end{figure}

\begin{figure}[H]
\centering
\tcbinputlisting{enhanced, boxrule=.45pt, arc=2mm, before skip=3pt, after skip=0pt,
  left=2mm, right=2mm, top=1mm, bottom=1mm,
  colback=black!4, colframe=black!30, listing only,
  listing file=Sections/judge_prompt_context_independence_excerpt.txt,
  listing options={basicstyle=\fontencoding{T1}\fontfamily{zi4}\fontsize{7.6}{7.8}\selectfont,
    columns=fullflexible, breaklines=true, breakatwhitespace=true,
    breakindent=0pt, keepspaces=true, showstringspaces=false,
    aboveskip=0pt, belowskip=0pt, literate={—}{{\textemdash}}1}}
\vspace*{0.2cm}    
\caption{\textit{Shortened context-independence judge prompt for MMLU questions.}
Each prompt contains two worked examples (scores 1 and 5); we show the score-5
example. The target question, option labels, and examples are adapted to
TinyMMLU and MMLU-Pro; the rubric is shared. Bracketed ellipses mark omissions.}
\label{fig:judge-prompt-full}
\end{figure}

\section{Comprehensive Overview of Mitigation Strategies}
\label{app:mitigation}

\subsection{\textsc{Cartridges++}: data mixing}
\label{app:data-mixing}

\begin{figure}[h!]
\centering
\includegraphics[width=\linewidth]{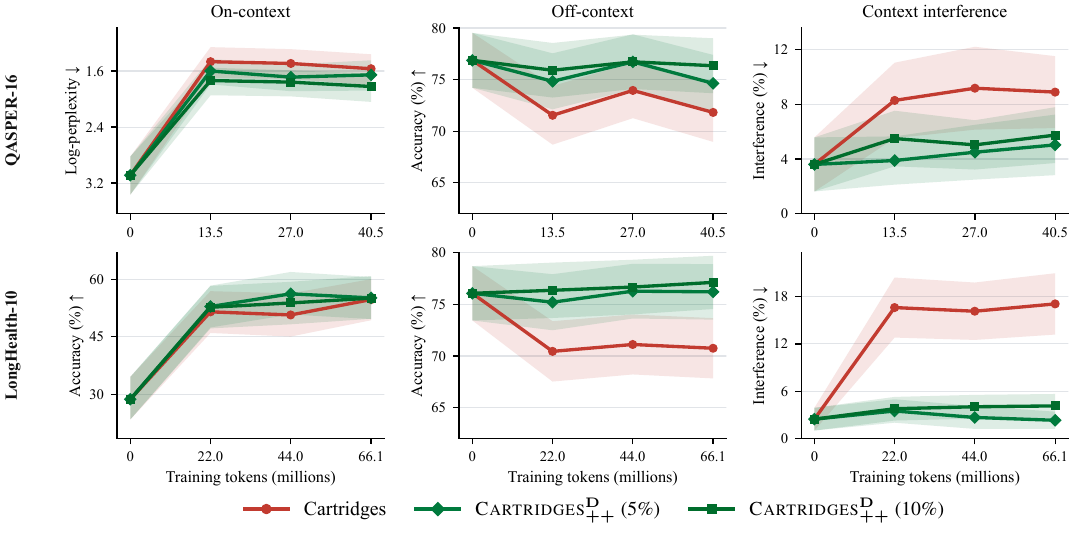}
\vspace*{-0.8cm}
\caption{\textit{Data mixing across training duration at 5\% compression.}
Standard ($\gamma=0$), 5\%-\texttt{Dolci}, and 10\%-\texttt{Dolci} Cartridges at matched update counts on
QASPER-16 (top; note the reversed y axis) and LongHealth-10 (bottom). Columns show on-context
performance, off-context accuracy, and interference. Shading shows 95\%
intervals.}
\vspace*{-0.2cm}
\label{fig:dolci-duration}
\end{figure}

\begin{figure}[h!]
\centering
\includegraphics[width=0.92\linewidth]{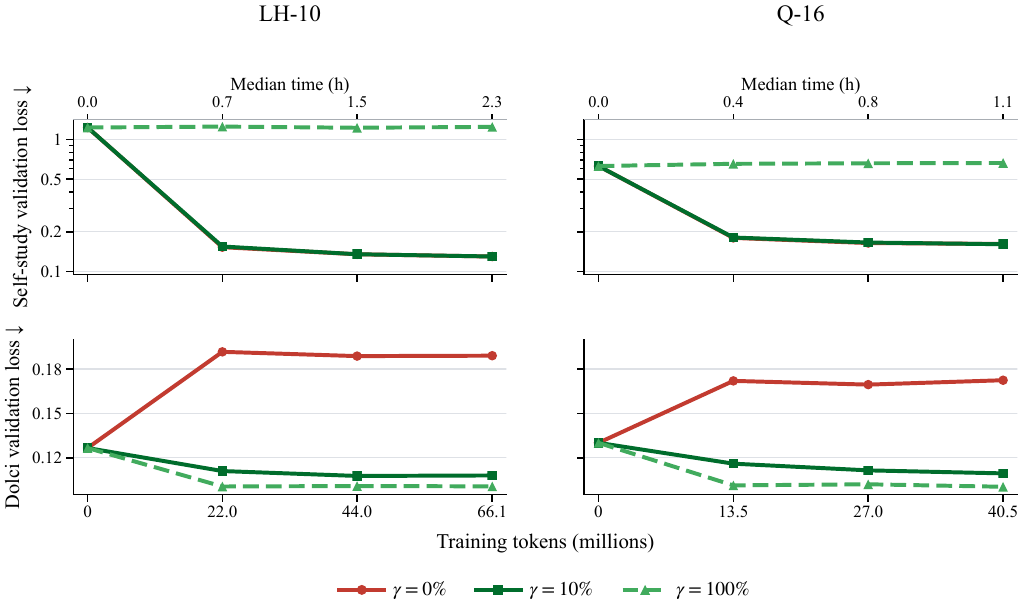}
\caption{\textit{Validation-loss dynamics of data mixing at 5\% compression factor.}
Held-out self-study loss (top) and held-out \texttt{Dolci} loss (bottom) for
$\gamma\in\{0,0.10,1\}$ on LongHealth-10 and QASPER-16. The primary x-axis
reports cumulative processed training tokens (65,536 tokens per optimizer
update); the top axes report median elapsed wall time across the three selected
runs at each checkpoint on \texttt{Qwen3-4B-Instruct-2507}.}
\label{fig:dolci-loss-dynamics}
\end{figure}

\begin{figure}[h!]
\centering
\includegraphics[width=\linewidth]{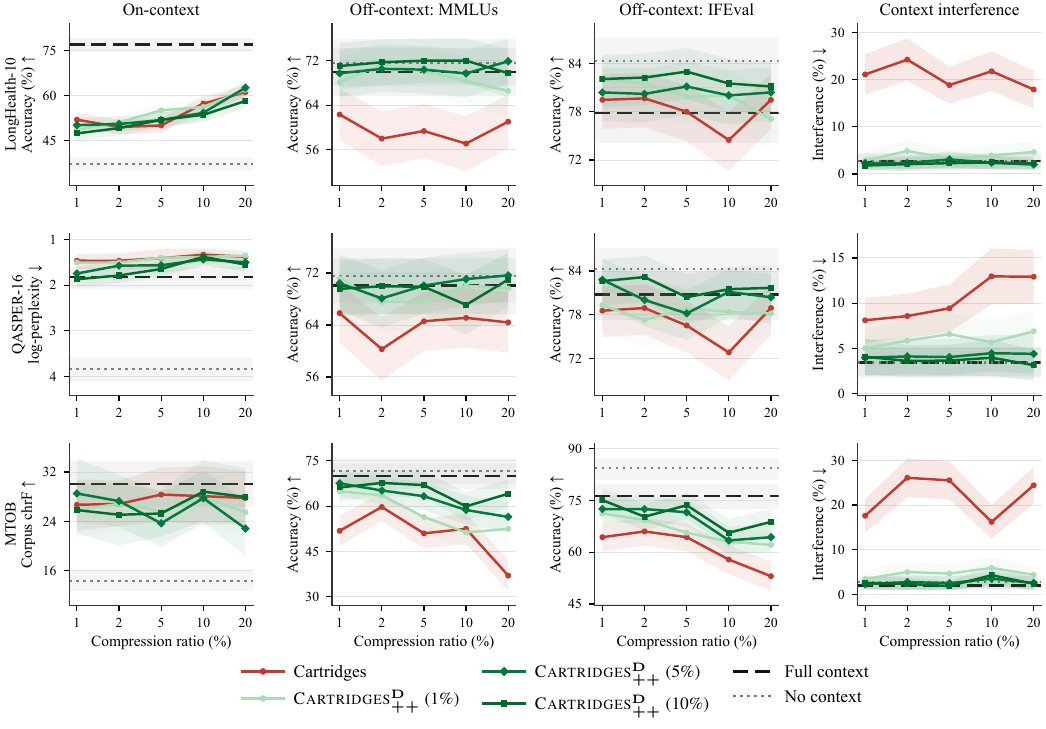}
\caption{\textit{\texttt{Qwen3-4B-Instruct-2507} data-mixing results at 1--20\% compression factor.} Rows
show LongHealth-10, QASPER-16 (note the reversed y axis), and MTOB; columns show on-context performance,
MMLUs accuracy, IFEval, and interference. Curves compare standard Cartridges
with 1/5/10\% \texttt{Dolci} mixtures. Shading shows 95\% intervals.}
\label{fig:dolci-four-contexts}
\end{figure}

Data mixing produces a smooth specialization--preservation trade-off rather
than an all-or-nothing change. Even a 1\% supervised-token mixture yields a
visible recovery on off-context accuracy and interference. Increasing the mixture
to 5--10\% generally moves TinyMMLU, MMLU-Pro, and IFEval closer to their
no-context references, while very large mixtures begin to reduce document
utility. The appropriate operating point therefore depends on whether the
application values maximal document specialization or conservative behavior on
unrelated requests. In the main paper, we aggregate the LongHealth-10 and
QASPER-16 dose--response curves over the available compression factors to visualize the full range of $\gamma$. Here, we
show training-duration sweeps at 5\% compression
(\Cref{fig:dolci-duration}). We also include the corresponding validation-loss dynamics
(\Cref{fig:dolci-loss-dynamics}), and their trajectories expose the specialization--preservation trade-off. Standard
self-study ($\gamma=0$) learns the document but increases held-out
\texttt{Dolci} loss; $\gamma=0.10$ matches its self-study convergence while
lowering \texttt{Dolci} loss; and $\gamma=1$ prioritizes \texttt{Dolci} and does
not learn the document. Finally, we show the 1/5/10\% \texttt{Dolci} mixtures across compression factors on LongHealth-10, QASPER-16, and MTOB
(\Cref{fig:dolci-four-contexts}).

\subsection{\textsc{Cartridges++}: routing}
\label{app:router}

\begin{figure}[h!]
\centering
\includegraphics[width=\linewidth]{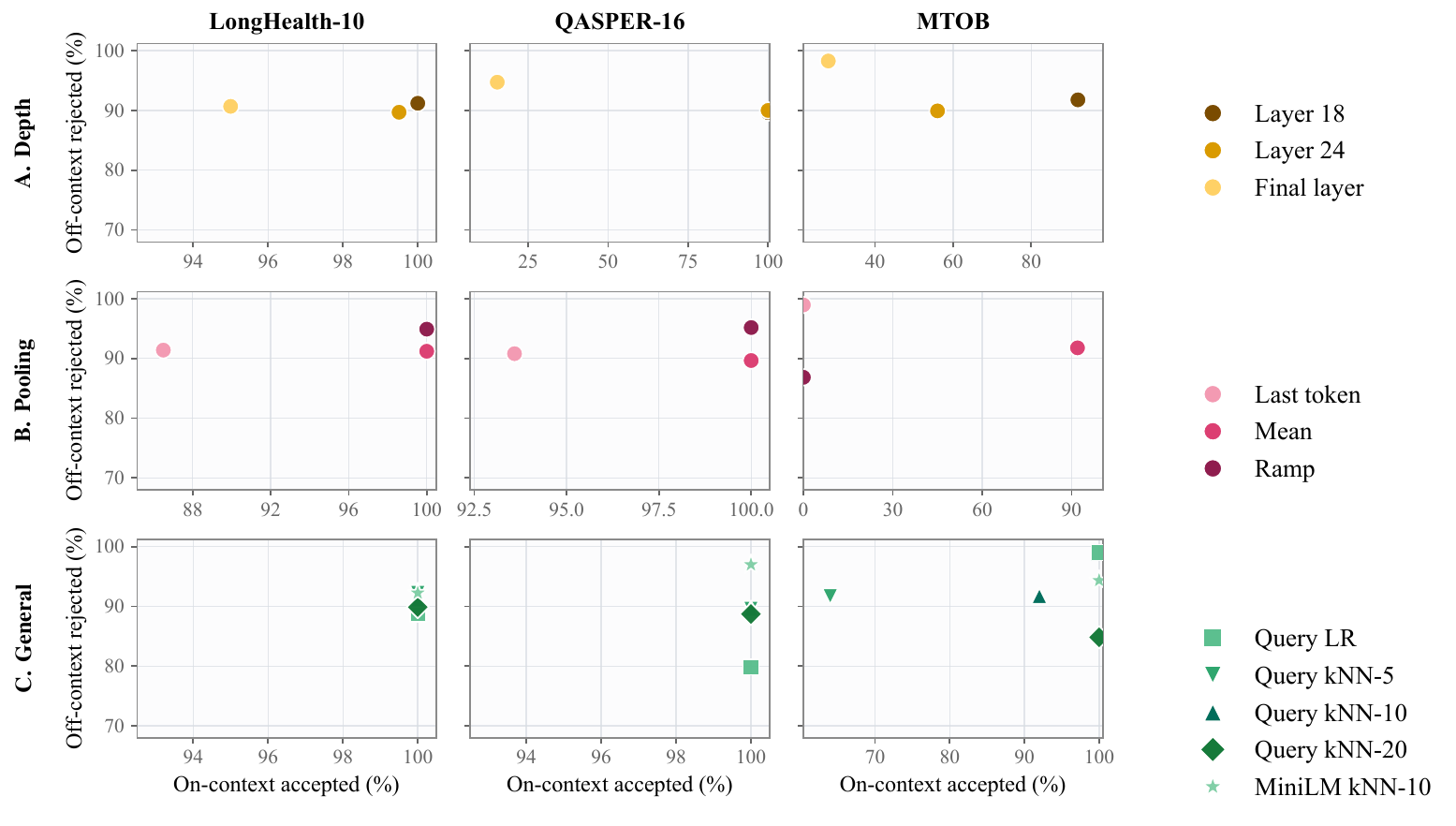}
\caption{\textit{\texttt{Qwen3-4B-Instruct-2507} router ablations.} Columns show LongHealth-10,
QASPER-16, and MTOB. Rows compare depth, pooling, and classifiers. Axes show
on-context acceptance and off-context rejection rates. Depth and pooling
comparisons use $k=10$; the final query-state classifiers use mean-pooled
layer-18 features. Note the \texttt{MiniLM} variant does not depend on \texttt{Qwen}.\looseness-1}
\label{fig:router-classifiers}
\end{figure}

Our representation and pooling ablations are motivated by
ADAPT~\citep{zhao2026adapt}, which scores training-data alignment using cosine
similarity between position-weighted, pooled final-layer states. We instead
evaluate frozen model representations for inference-time relevance routing,
rather than online data reweighting.
\Cref{fig:router-classifiers} reports the controlled \texttt{Qwen} router-selection
sequence. We first compare hidden-state depth at fixed mean pooling and
$k=10$, then compare pooling at fixed layer 18 and $k=10$, and finally compare
base-model query embeddings with LR, $k$NN-5/10/20, and \texttt{MiniLM} embeddings with $k$NN-10. Each panel plots the
fraction of native document questions accepted against the fraction of
off-context questions rejected. Layer 18 gives a consistently strong acceptance--rejection balance. Mean and ramp pooling perform similarly across datasets. On English QA,
$k$NN rejects more off-context queries than fitted LR at similar on-context
acceptance, possibly because local neighborhoods accommodate relevance
patterns beyond a single linear boundary; LR is stronger on MTOB.
\texttt{MiniLM} is also competitive and offers a smaller, backbone-independent
encoder for cheaper feature extraction. We fix $k=10$ as an intermediate
choice, using $k=5/20$ for sensitivity checks rather than tuning per dataset.

\paragraph{Compute cost.}
For the hidden-state router, we first prefill the query without the Cartridge
and classify its pooled representation. A rejected query can continue directly
along this clean path; only an accepted query is re-prefilled with the
Cartridge. The extra pass processes query tokens in parallel against the
compact cache, unlike sequential autoregressive decoding.
This ordering is preferable when irrelevant queries are common
because it pays the second 4B-model prefill only when the memory is needed.
\texttt{MiniLM} makes the decision before either model path, so each query incurs only
its selected prefill. For short queries and long generations, we expect the
routing overhead to be small relative to decoding; its magnitude depends on
the workload and implementation.

\newpage
\subsection{Prompting baseline}
\label{app:prompting}

\begin{figure}[h!]
\centering
\includegraphics[width=\linewidth]{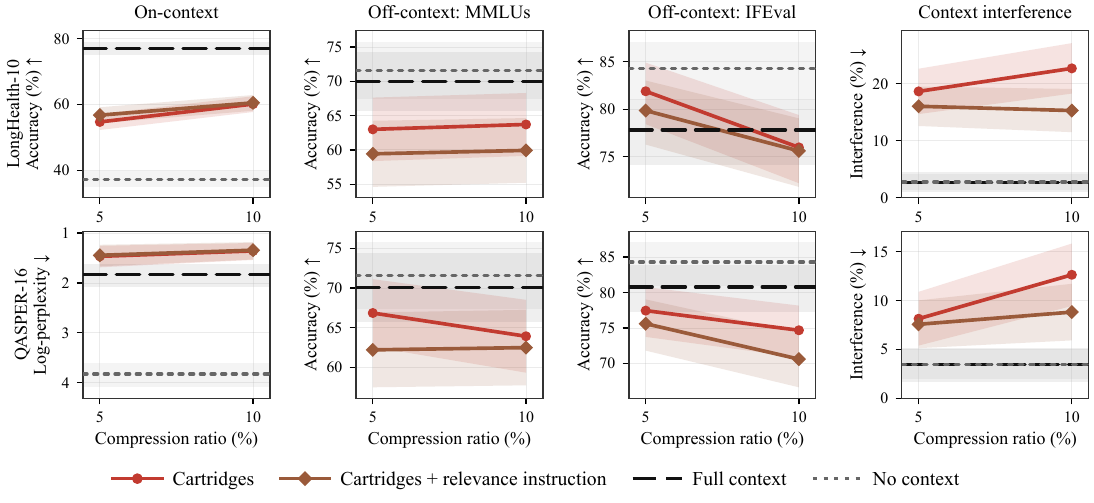}
\caption{\textit{Relevance-instruction ablation at 5\% and 10\% compression factor on \texttt{Qwen3-4B-Instruct-2507}.}
Cartridges with and without the added instruction on LongHealth-10 (top) and
QASPER-16 (bottom; note the reversed y axis). Columns show on-context performance, MMLUs accuracy,
IFEval, and interference. Dashed and dotted lines denote full and no context;
shading shows 95\% intervals.\looseness-1}
\label{fig:prompting-baseline}
\end{figure}

We test whether the model can suppress interference through an explicit user
instruction alone. Immediately before the question, after the cached
Cartridge, we insert: \emph{``Use the available document context only when it
is relevant to the question. If it is not relevant, ignore it and answer using
your general knowledge.''} \Cref{fig:prompting-baseline} compares this with
no added instruction.
The relevance instruction does not reliably improve off-context accuracy or
IFEval and yields no consistent source-task gain. It reduces judged
interference in some cells, but does not recover the no-context behavior.

\clearpage
\newpage
\section{Effect of Training Duration}
\label{app:training-duration}

\begin{figure}[h!]
\centering
\includegraphics[width=\linewidth]{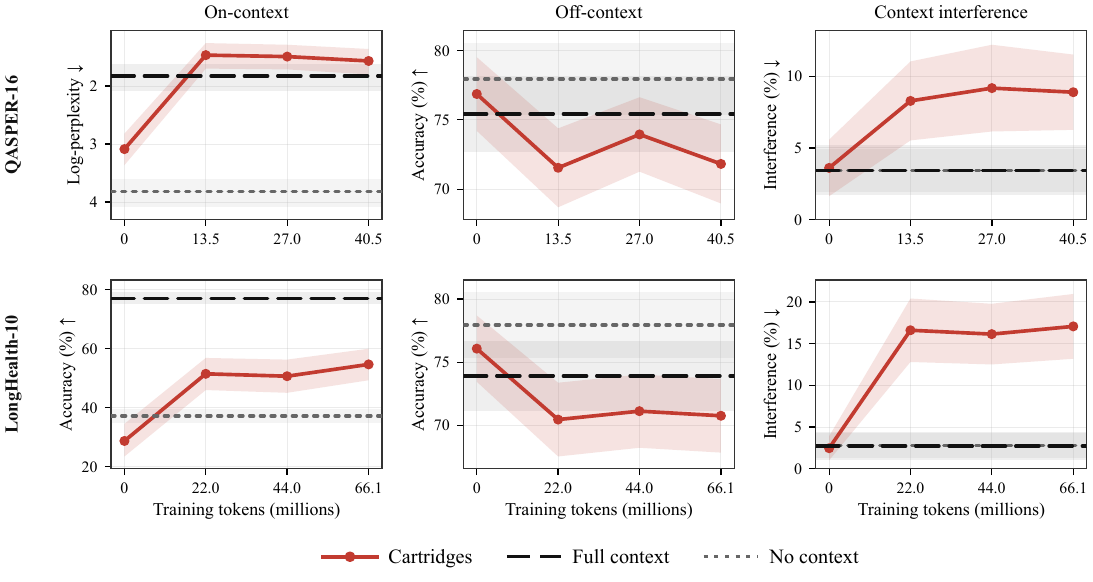}
\caption{\textit{Cartridge training duration at 5\% compression.} QASPER-16
(top; note the reversed y-axis) and LongHealth-10 (bottom), before optimization and at successive
training checkpoints. Columns show on-context performance, off-context
accuracy, and interference.
Shading shows 95\% intervals on \texttt{Qwen3-4B-Instruct-2507}.}
\label{fig:cartridge-training-duration}
\end{figure}

\paragraph{Standard Cartridges.}
The standard recipe evaluates the Cartridge after one pass over the self-study
data. Because the original study reports mild source-task gains from longer
optimization, we continue beyond one pass and track both context fidelity and
off-context behavior (\Cref{fig:cartridge-training-duration}). Most of the
source-task gain appears during the first pass. Later checkpoints produce only
small fluctuations, occasionally improving and occasionally degrading the
source task. Off-context accuracy
and context interference likewise plateau at degraded levels and never recover
at any later checkpoint. The failure is therefore not an artifact of the
chosen number of Cartridge optimization steps.\looseness-1

\paragraph{Data-mixed Cartridges.}
The mixed variants show the same early source-task convergence while retaining
their off-context advantage at later checkpoints (\Cref{fig:dolci-duration}). At
matched update counts, 5--10\% mixing keeps off-context performance close to its
initial level and prevents the large interference increase of self-study-only
training. Longer optimization therefore neither creates nor erases the main
benefit of data mixing.

\newpage
\section{Additional Long-Context Results}
\label{app:add-results}

\subsection{Per-benchmark overview}
\label{app:per-benchmark}

\begin{figure}[h!]
\centering
\includegraphics[width=\linewidth]{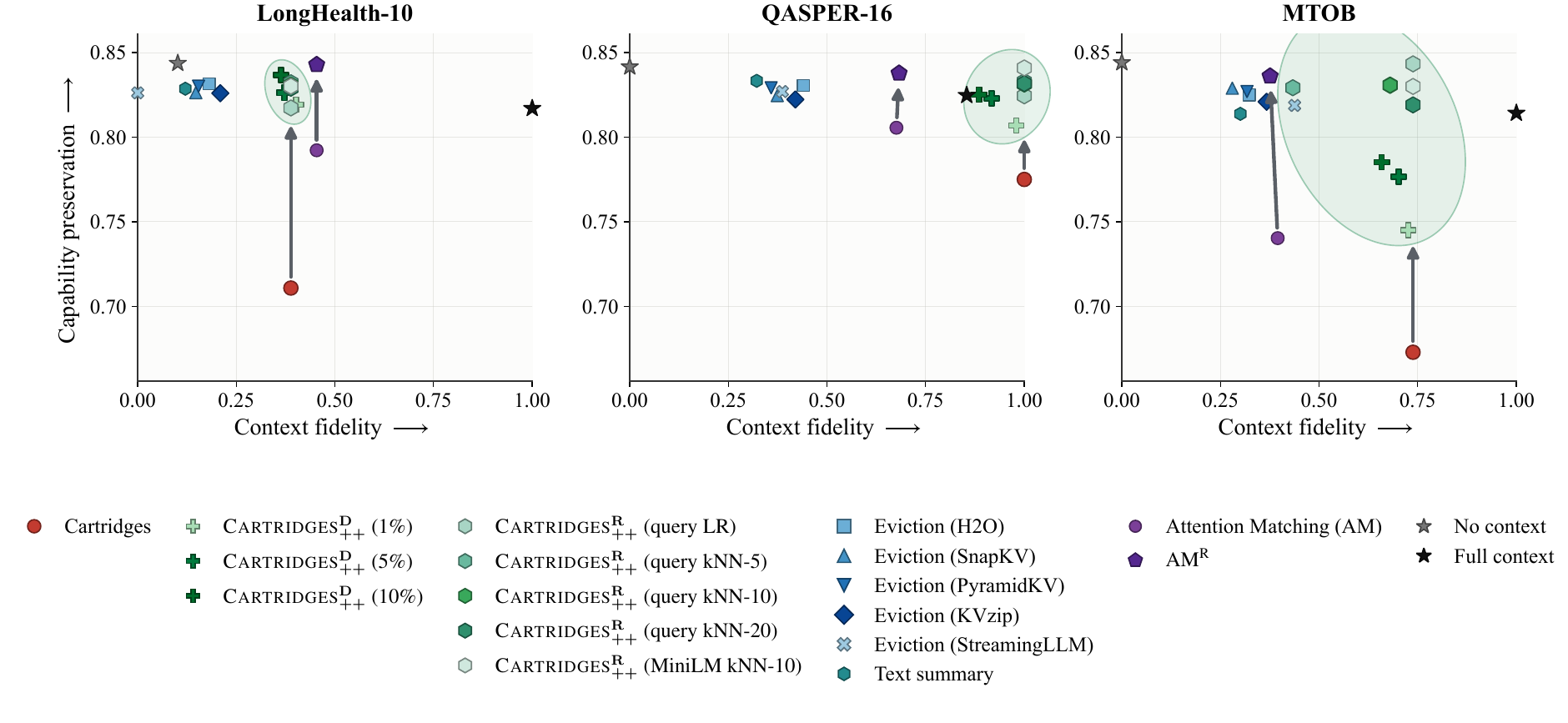}
\vspace*{-0.8cm}
\caption{\textit{\texttt{Qwen3-4B-Instruct-2507} context fidelity and capability preservation by dataset.}
Baselines, 1/5/10\% \texttt{Dolci} mixtures, router variants, and routed AM, averaged
over 0.1/1/2/5/10\% compression factor. Axes are defined in
Appendix~\ref{app:per-benchmark}; both are higher-is-better.}
\label{fig:qwen-mitigation-planes-by-context}
\end{figure}

\begin{figure}[h!]
\centering
\includegraphics[width=\linewidth]{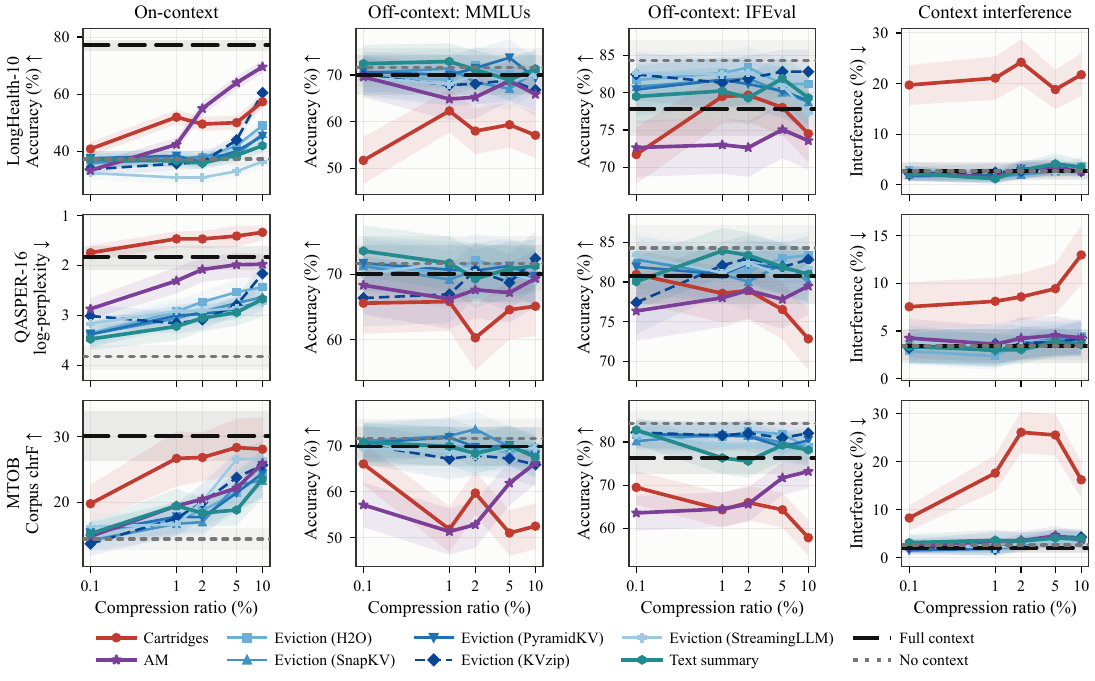}
\caption{\textit{\texttt{Qwen3-4B-Instruct-2507} baseline comparison at 0.1--10\% compression.} Rows
show LongHealth-10, QASPER-16 (note the reversed y axis), and MTOB. Columns show on-context performance,
equal-weight TinyMMLU/MMLU-Pro accuracy, IFEval strict accuracy, and context
interference. Shading shows available 95\% intervals.}
\label{fig:all-contexts-methods}
\end{figure}

\begin{figure}[h!]
\centering
\includegraphics[width=\linewidth]{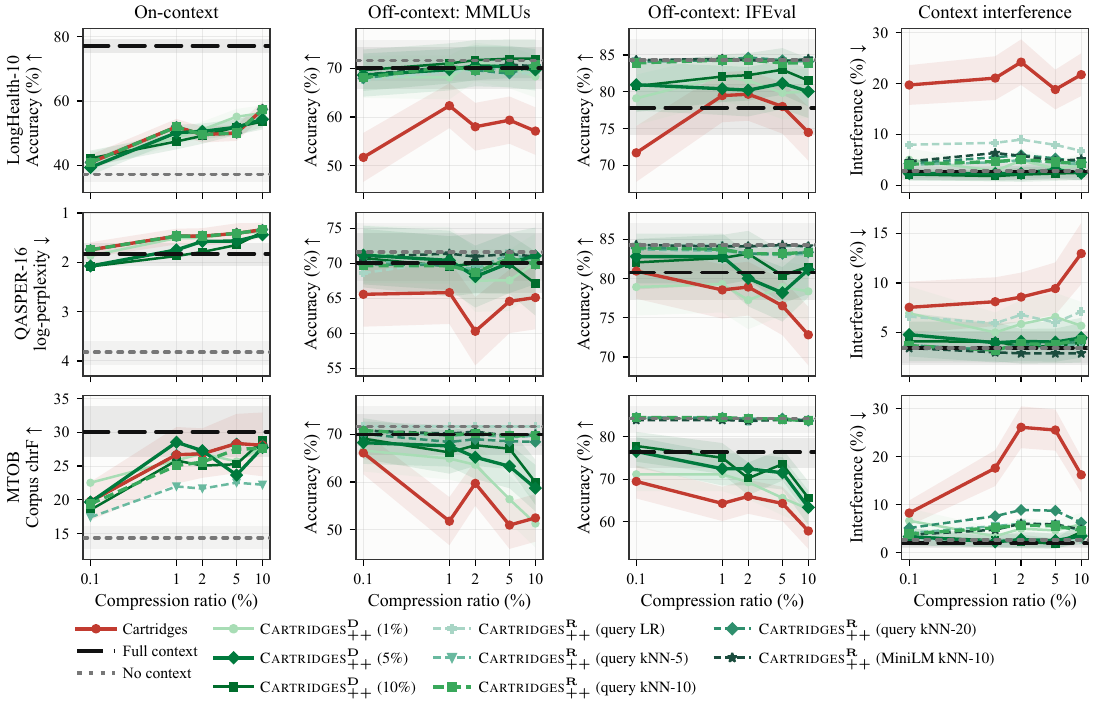}
\caption{\textit{\texttt{Qwen3-4B-Instruct-2507} mitigation variants at 0.1--10\% compression factor.}
Cartridges, 1/5/10\% \texttt{Dolci} mixtures, and five router variants, using the rows
and metrics of \Cref{fig:all-contexts-methods}. Shading shows available
95\% intervals; routed curves have no bands.}
\label{fig:all-contexts-mitigation-variants}
\end{figure}

\Cref{fig:qwen-mitigation-planes-by-context,fig:all-contexts-methods,fig:all-contexts-mitigation-variants} expose three recurring patterns. First, Cartridges are most compelling on the longer source tasks, particularly QASPER
and MTOB. Second, their capability-preservation curves do not reliably improve
with a larger cache; on MTOB, some off-context metrics even deteriorate with increasing compression factor. Third, the eviction family is comparatively clustered off-context even
when its members differ substantially on the source task. This separation is
why reporting context fidelity alone gives an incomplete ranking of compressed
memories. \Cref{fig:qwen-mitigation-planes-by-context} expands \Cref{fig:overview} (right)
per dataset and adds further \textsc{Cartridges++} variants; its arrows show the effect
of our mitigations on both Cartridges and Attention Matching.
\Cref{fig:all-contexts-methods,fig:all-contexts-mitigation-variants} likewise break
\Cref{fig:compression-sweep,fig:mitigation-five-panel} down by dataset and metric, showing
in finer detail where Cartridges fail and how our mitigations address these failures.

\paragraph{Mitigation planes.}
For each dataset and method, we first average each metric over compression
factors. The x-axis min--max scales on-context performance (negated
log-perplexity for QASPER-16) across all methods of the per-dataset plane,
including the full- and no-context references. The y-axis is unscaled: the
mean of MMLUs (mean TinyMMLU/MMLU-Pro accuracy), IFEval accuracy, and one minus
context interference. Aggregate planes
(\Cref{fig:overview,fig:gemma-main-plane}) average the per-dataset coordinates
with equal weight.


\newpage
\subsection{Results in another model family: \texttt{Gemma}}
\label{app:gemma-results}

\begin{figure}[H]
\centering
\includegraphics[width=\linewidth]{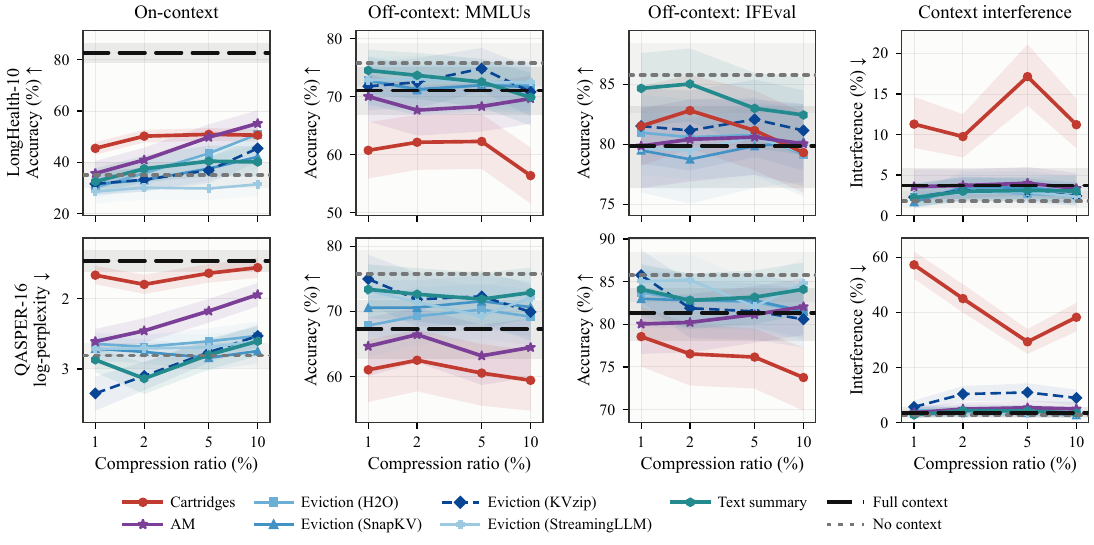}
\caption{\textit{\texttt{Gemma-4-E4B-it} baseline comparison at 1--10\% compression factor.}
Cartridges and reusable baselines on LongHealth-10 (top) and QASPER-16
(bottom; note the reversed y axis). Columns show on-context performance, MMLUs accuracy, IFEval, and
interference. Shading shows available 95\% intervals.}
\label{fig:gemma-results}
\end{figure}

\begin{figure}[H]
\centering
\includegraphics[width=\linewidth]{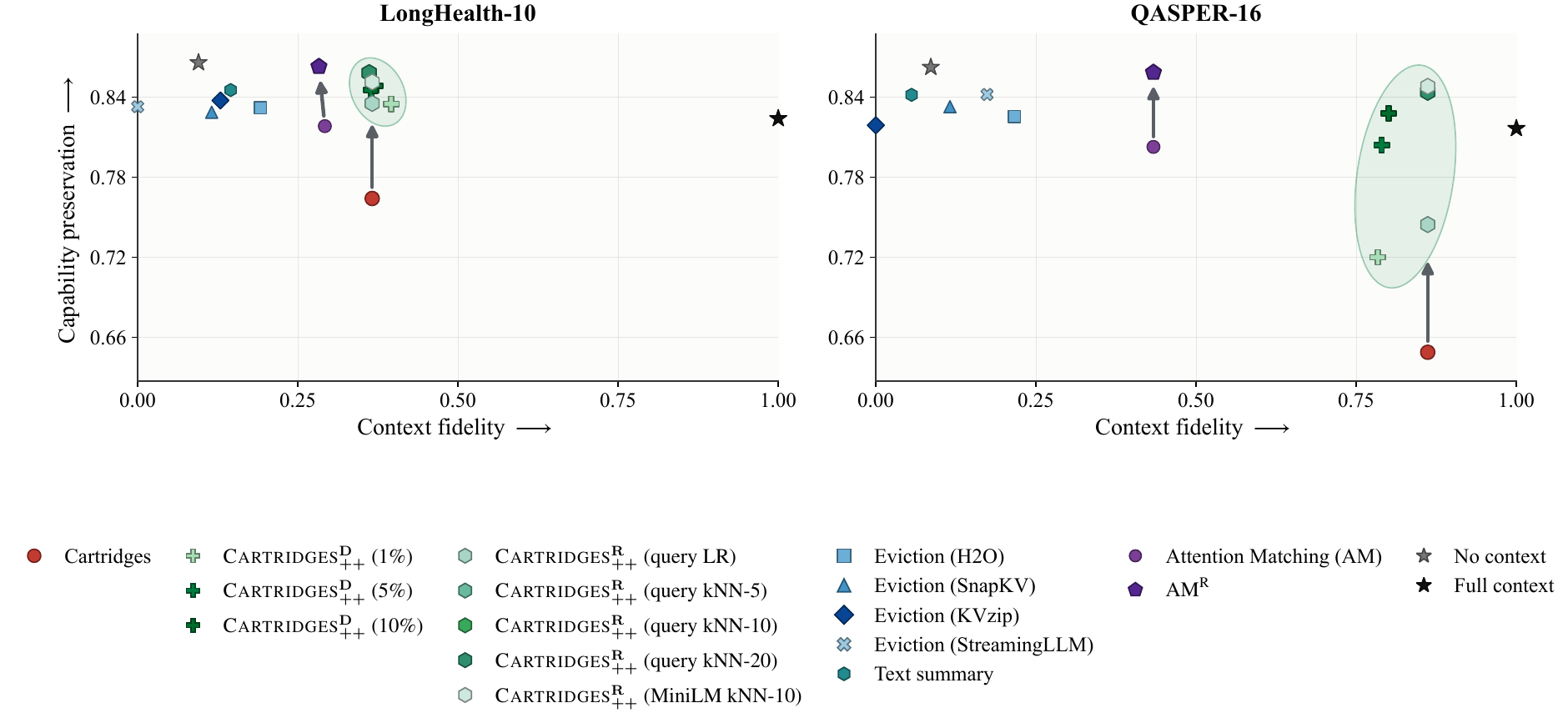}
\caption{\textit{\texttt{Gemma-4-E4B-it} context fidelity and capability preservation by dataset.}
The \texttt{Gemma} counterpart of \Cref{fig:qwen-mitigation-planes-by-context} and the
per-dataset breakdown of \Cref{fig:gemma-main-plane}. LongHealth-10 and QASPER-16 results
are averaged over 1/2/5/10\% compression factors, using the axes of
Appendix~\ref{app:per-benchmark} and the method groups of
\Cref{fig:qwen-mitigation-planes-by-context}; arrows show the effect of our mitigations
on Cartridges and Attention Matching.\looseness-1}
\label{fig:gemma-mitigation-planes-by-context}
\end{figure}

\begin{figure*}[h!]
\centering
\includegraphics[width=\textwidth]{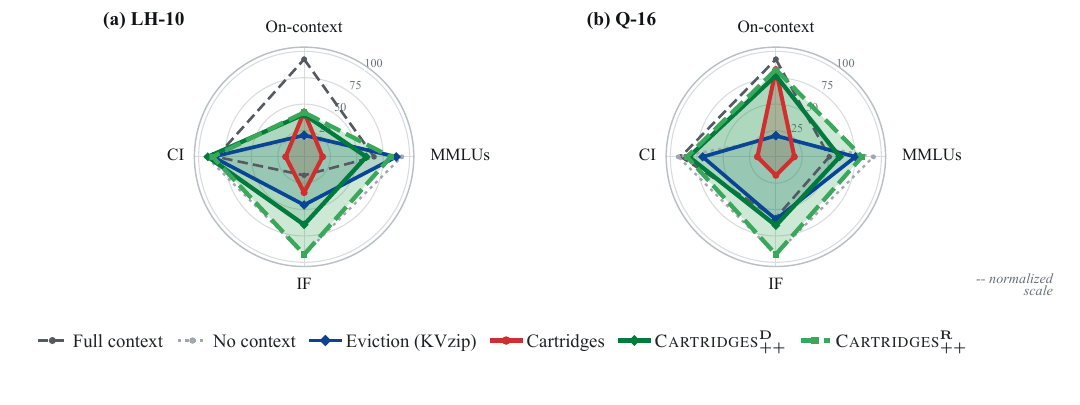}
\caption{\textit{\texttt{Gemma-4-E4B-it} results at 5\% compression factor.} LongHealth-10
(left) and QASPER-16 (right), with the axes and scaling of
\Cref{fig:main-tradeoff}. The eviction baseline is KVzip; mitigations use
5\% \texttt{Dolci} mixing and query $k$NN-10 routing.}
\label{fig:gemma-radar}
\end{figure*}

\paragraph{Standard Cartridges.}
We repeat the compression factor sweep with \texttt{Gemma-4-E4B-it}, an architecture that
interleaves local/sliding-window and global attention rather than using only
global attention. Standard Cartridges reproduce the same qualitative
asymmetry as in the \texttt{Qwen} family: they improve source-task performance but reduce off-context knowledge
and instruction following as the retained memory grows. The failure is
therefore not specific to fully global-attention architectures, and the mitigations we propose with \textsc{Cartridges++} are successful in this setup as well
(Figures~\ref{fig:gemma-results},~\ref{fig:gemma-mitigation-planes-by-context}, and~\ref{fig:gemma-radar}).\looseness-1

\begin{figure}[h!]
\centering
\includegraphics[width=\linewidth]{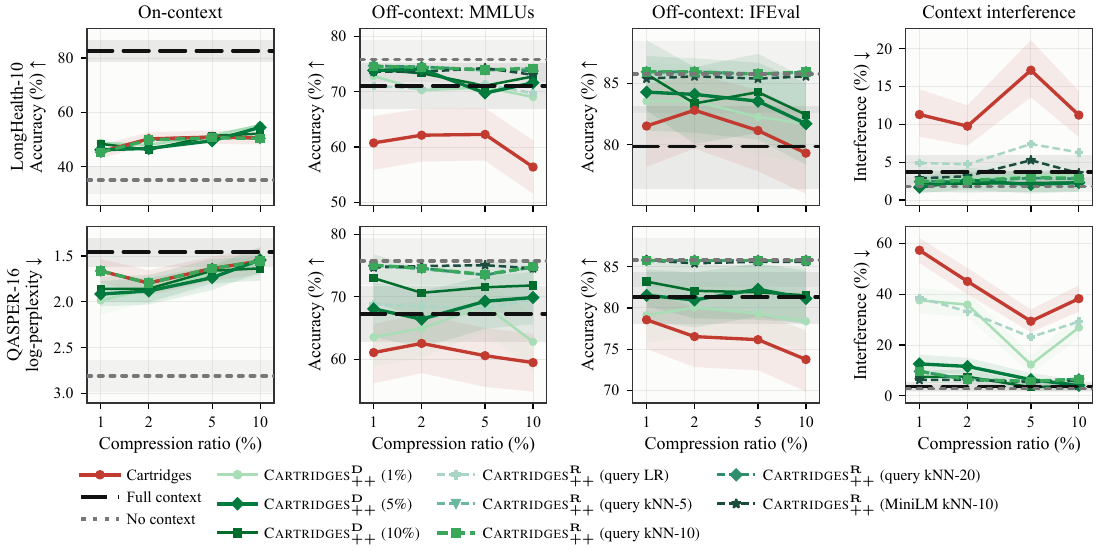}
\caption{\textit{\texttt{Gemma-4-E4B-it} mitigation variants at 1--10\% compression factor.}
Cartridges, 1/5/10\% \texttt{Dolci} mixtures, and five router variants, using the rows
and metrics of \Cref{fig:gemma-results}. Shading shows available 95\%
intervals.}
\label{fig:gemma-mitigation-variants}
\end{figure}

\paragraph{Data mixing.}
The \texttt{Gemma} replication uses the same frozen \texttt{Dolci} example ordering, rescored by
the \texttt{Gemma} teacher, and matches 1/5/10\% by \texttt{Gemma} supervised-token count. The
mixtures are compared with standard Cartridges at identical completed-update
boundaries. Mixing recovers much of the off-context loss while leaving the
source-task curve close to the standard Cartridge. The
preservation--specialization trade-off therefore also transfers across model
families (\Cref{fig:gemma-mitigation-variants}).

\begin{figure}[h!]
\centering
\includegraphics[width=0.86\linewidth]{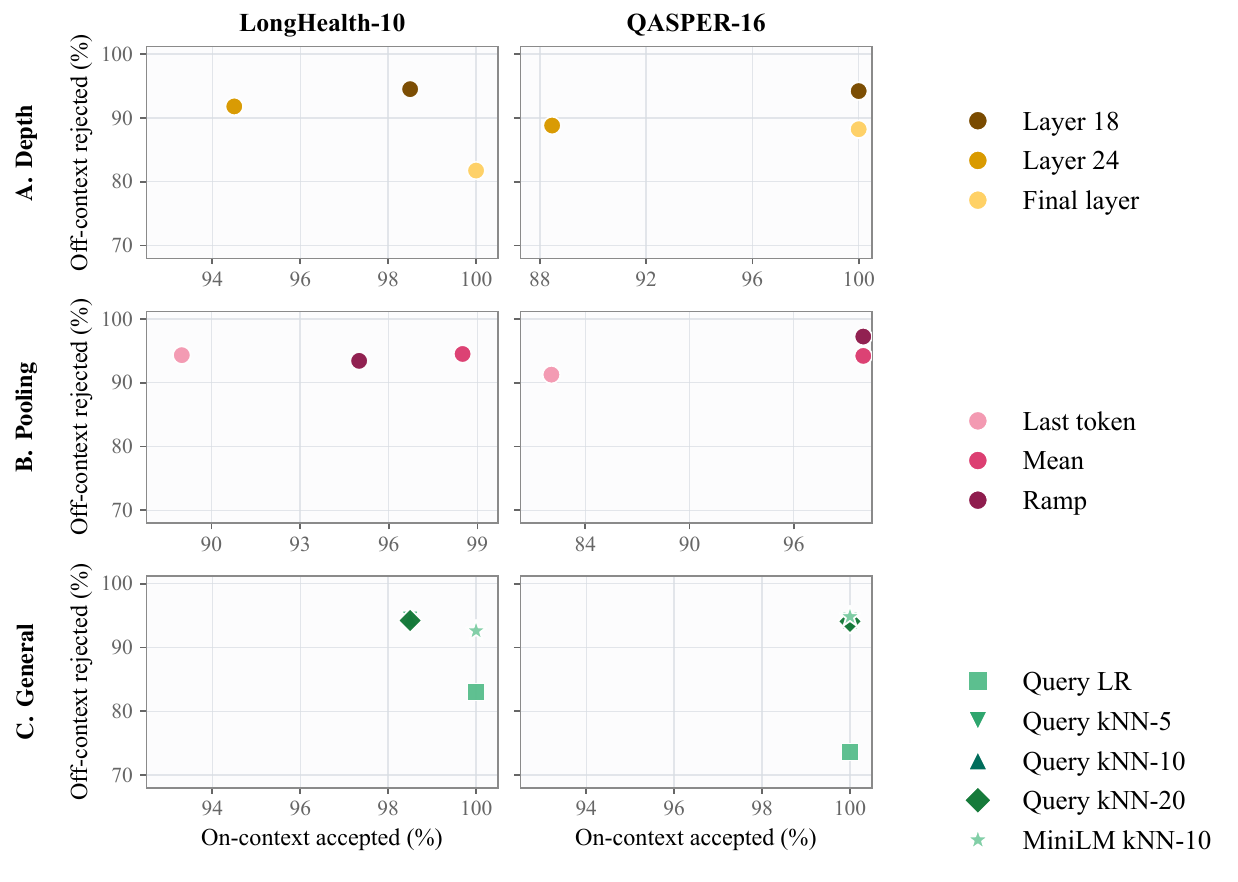}
\caption{\textit{\texttt{Gemma-4-E4B-it} router ablations.} LongHealth-10 (left) and
QASPER-16 (right), with the same depth, pooling, and classifier comparisons
as \Cref{fig:router-classifiers}. Axes show on-context acceptance and
off-context rejection rates.}
\label{fig:router-recipe-gemma}
\end{figure}

\paragraph{Routing.}
We additionally construct and calibrate the router using \texttt{Gemma} representations
and the same disjoint self-study/\texttt{Dolci} protocol described above. The depth and
pooling ablations are repeated, and the selected recipe remains mean-pooled
layer-18 features with cosine $k$NN-10. The \texttt{MiniLM}
router is the exception: because it embeds only the query text and does not
read hidden states from the evaluated language model, it is backbone-agnostic
and can be applied to \texttt{Gemma}--\texttt{Qwen} without retraining its feature extractor
(\Cref{fig:router-recipe-gemma}).

\newpage
\subsection{Smaller contexts: QuALITY}
\label{app:quality-results}

\begin{figure}[h!]
\centering
\includegraphics[width=\linewidth]{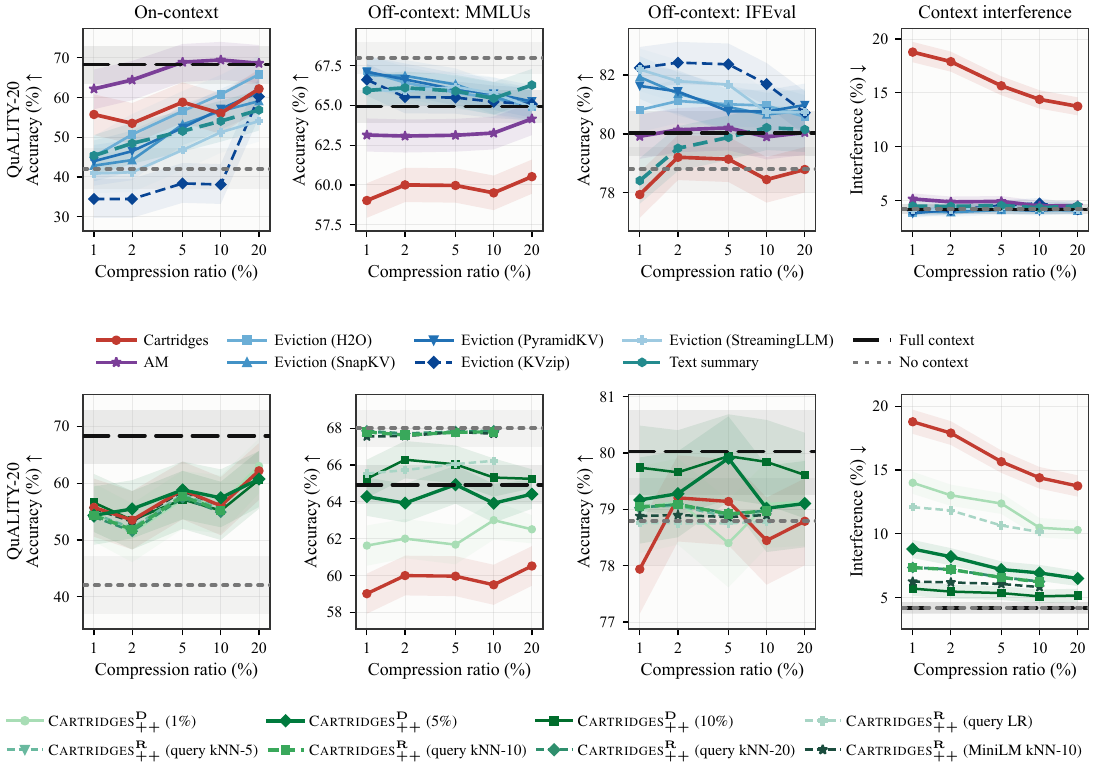}
\caption{\textit{\texttt{Qwen3-4B} results on QuALITY-20.} Top: Cartridges and
reusable baselines. Bottom: Cartridges, 1/5/10\% \texttt{Dolci} mixtures, and five
router variants. Columns use the metrics of \Cref{fig:all-contexts-methods}.
Compression factors span 1--20\%. Shading shows
available 95\% intervals.}
\label{fig:quality-results}
\end{figure}

QuALITY provides a useful counterpoint to the 100K-token settings. We train a
separate Cartridge for each of its 20 selected documents and pool the 357 questions for
evaluation. At $\sim$8K tokens, training-free eviction and the text-space
baseline can match or exceed Cartridges on context-driven benchmarks, so learned
memory is not uniformly the best source-task compressor. Nevertheless, the
off-context gap remains: activating a QuALITY-20 Cartridge lowers general-knowledge
and instruction-following performance and increases judged interference. The
failure therefore appears before the context is long enough for a Cartridge's
on-context advantage to dominate (\Cref{fig:quality-results}, top). Both
mitigations transfer to this regime: data mixing and the $k$NN routers bring
off-context accuracy and interference back toward the no-context reference
while keeping the Cartridge's on-context accuracy
(\Cref{fig:quality-results}, bottom).
We use the same router configurations as in the longer-context experiments
(Appendix~\ref{app:router}), including cosine $k$NN-10 on mean-pooled layer-18
query states, with a separate reference bank and calibrated threshold for each
QuALITY-20 article.

\subsection{QASPER: generated-answer F1}
\label{app:qasper-f1}

We use teacher-forced log-perplexity as the primary QASPER metric to
match the original Cartridges evaluation. As a generation-level check, we also
decode one greedy answer for each of
the same 78 questions and compute token F1 against the reference answers.
Uncertainty is estimated by resampling paper clusters. Token F1 is the
conventional answer-overlap metric for QASPER, so this secondary evaluation
also makes our result comparable to the benchmark's standard generation-based
reporting.
The generated metric supports the same conclusion: Cartridges have a large
source-task advantage at high compression and remain among the strongest methods
as the budget grows. The agreement shows that the perplexity-based QASPER advantage is not an
artifact of teacher forcing, although the two metrics do not need to induce an
identical ordering at every compression factor.\looseness-1

\begin{figure}[h!]
\centering
\includegraphics[width=\linewidth]{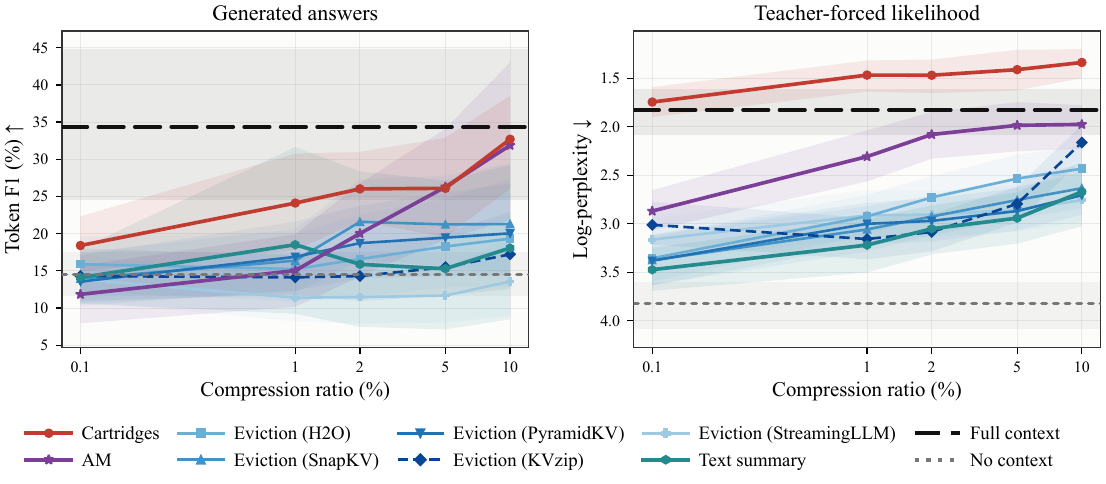}
\caption{\textit{QASPER-16 performance under two metrics.} Left:
generated-answer token F1. Right: teacher-forced, token-weighted
log-perplexity. Both use the same 78 questions. Shading shows 95\% bootstrap intervals on \texttt{Qwen3-4B-Instruct-2507}.}
\label{fig:qasper-f1-ppl}
\end{figure}

\subsection{Fine-grained MMLU-Pro categories}
\label{app:mmlu-categories}

\begin{figure}[h!]
\centering
\includegraphics[width=\linewidth]{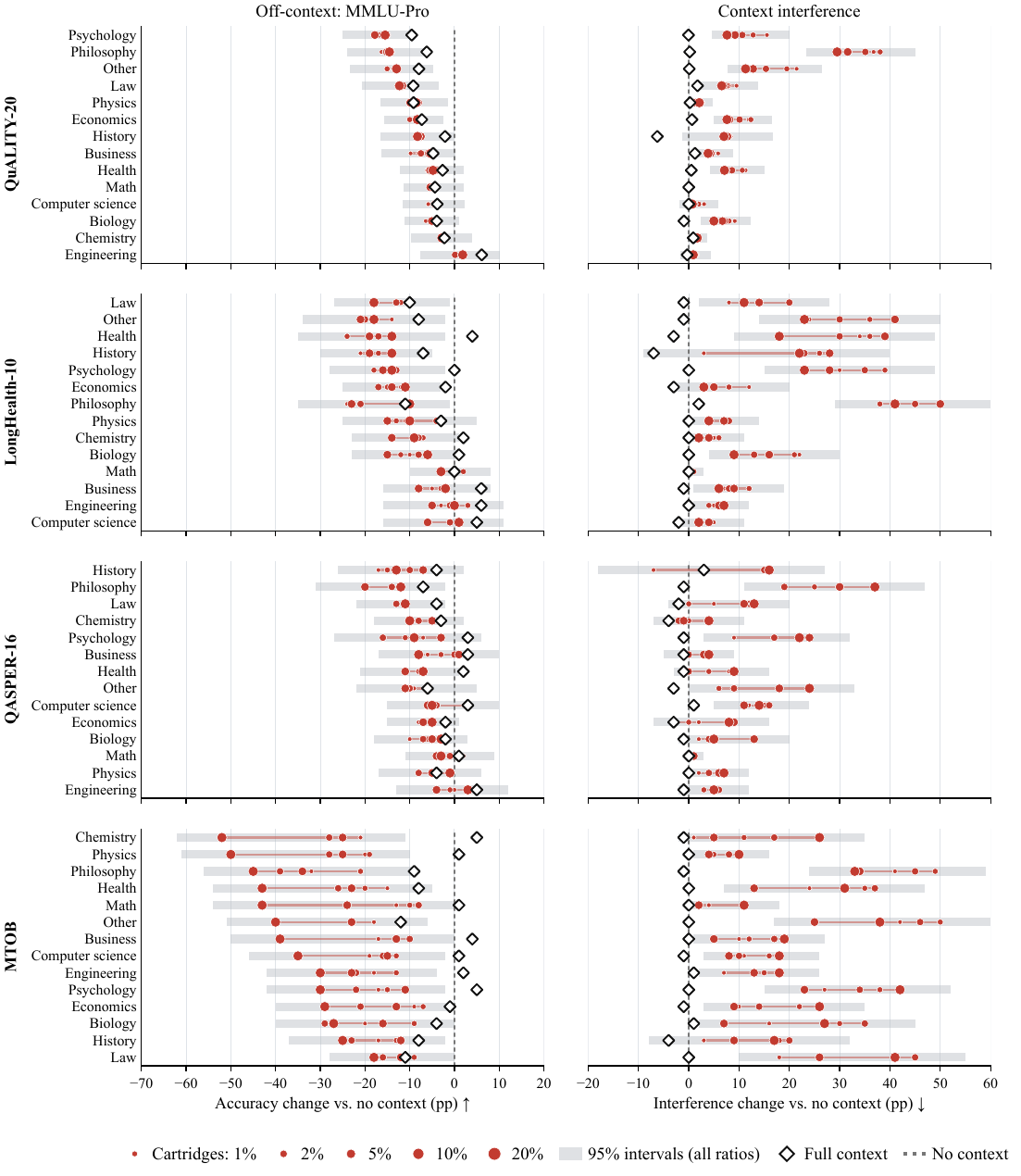}
\caption{\textit{MMLU-Pro results by subject and resident context on the \texttt{Qwen3-4B} family.} Changes
in accuracy (left) and interference (right) relative to no context, in
percentage points. Subjects are ordered by the 20\%-compression factor Cartridge
accuracy change. Marker size encodes 1--20\% compression factor; gray bars span the
per-budget 95\% intervals. Diamonds denote full context.\looseness-1}
\label{fig:mmlu-category-breakdown}
\end{figure}

Off-context degradation is very subject dependent on MMLU-pro. In this section we decompose the performance per subject. At 20\% compression factor with
a LongHealth cartridge, the largest accuracy losses occur in ``law'' and the
``other'' category, while ``philosophy'' shows the largest interference increase.
QASPER also shows elevated interference in ``philosophy'' and other prose-heavy
categories, while mathematics has little additional interference, likely driven by the scientific nature of QASPER itself.  In contrast, MTOB's
largest accuracy losses occur in chemistry, physics, and mathematics. Thus, the
subjects with the largest accuracy losses need not be those with the most
judged interference (\Cref{fig:mmlu-category-breakdown}).

\begingroup
\tcbset{fabcase/.style={enhanced,arc=1.2mm,boxrule=0.5pt,
  colframe=black!45,colback=white,colbacktitle=black!7,coltitle=black,
  fonttitle=\bfseries\small,
  boxsep=0.3mm,middle=0.7mm,
  left=2mm,right=2mm,top=1mm,bottom=1mm,toptitle=0.4mm,bottomtitle=0.4mm,
  before upper={\fontsize{6.6}{7.4}\selectfont\ttfamily\frenchspacing\raggedright\setlength{\parskip}{0pt}},
  before lower={\fontsize{7.5}{8.5}\selectfont\raggedright\setlength{\parskip}{0pt}},
  space to upper,segmentation style={black!30,dashed}}}
\newcommand{\fabtitle}[2]{#1\hfill\normalfont\textit{#2}}
\newcommand{\fabfield}[1]{{\normalfont\bfseries\color{black!55}\MakeUppercase{#1}}\par\nobreak}
\newcommand{\fabkey}[1]{\textbf{#1}\enspace}
\newcommand{\fabok}{\textcolor{green!45!black}{\ding{51}}}
\newcommand{\fabno}{\textcolor{red!70!black}{\ding{55}}}
\begin{figure*}[p]
\centering
\captionsetup{font=small}
\begin{tcbraster}[raster columns=2,raster equal height=rows,
  raster column skip=3mm,raster row skip=2mm]
\begin{tcolorbox}[fabcase,
  title={\fabtitle{\fabok\ Correct answer}{LongHealth}}]
\fabfield{Question}Cell motility, as viewed in a cultured fibroblast,
encompasses all of the following \textsc{except}:\par\smallskip
\fabfield{Options}(A)~Adhesion plaques \enspace (B)~Vinculin \enspace
(C)~Clathrin \enspace (D)~Lamellipodia
\tcbline
\fabfield{Response}"In the text, under `Microscopy' in the `Cell motility'
section, it states: `Cell motility, as viewed in a cultured fibroblast,
encompasses adhesion plaques, vinculin, and lamellipodia. Clathrin is not
involved in this process.' Therefore, the option that is not part of cell
motility is (C) clathrin."
\tcblower
\fabkey{Prediction}C (clathrin)\hfill\fabkey{Gold}C (clathrin)
\end{tcolorbox}
\begin{tcolorbox}[fabcase,
  title={\fabtitle{\fabok\ Correct answer}{QASPER}}]
\fabfield{Question}According to BCS theory, the attraction between Cooper
pairs in a superconductor is due to:\par\smallskip
\fabfield{Options}(A)~Magnetic attraction \enspace (B)~Vacuum polarization
\enspace (C)~Interactions with the ionic lattice \enspace
(D)~Gravitational pull \enspace (E)~Electrostatic attraction \enspace
(F)~The weak nuclear force \enspace (G)~The Pauli exclusion principle
\enspace (H)~Quantum entanglement \enspace (I)~Interactions with the
electron cloud \enspace (J)~The strong nuclear force
\tcbline
\fabfield{Response}"In the provided text, the following sentence is given:
`In BCS theory, the attraction between electrons is due to interactions with
the ionic lattice.' The only option that matches the explanation in the text
is (C)."
\tcblower
\fabkey{Prediction}C (ionic lattice)\hfill\fabkey{Gold}C
\end{tcolorbox}
\begin{tcolorbox}[fabcase,
  title={\fabtitle{\fabno\ Incorrect answer}{LongHealth}}]
\fabfield{Question}When was the current Dalai Lama born?\par\smallskip
\fabfield{Options}(A)~1955 \enspace (B)~1985 \enspace (C)~1975 \enspace
(D)~1990 \enspace (E)~2000 \enspace (F)~1965 \enspace (G)~1900 \enspace
(H)~1925 \enspace (I)~1940 \enspace (J)~1935
\tcbline
\fabfield{Response}"The text states: `The patient, born in 1949, has a
diagnosis of pancreatic cancer. \ldots{} The Dalai Lama was born in 1955.'
Therefore, the answer is (A), 1955."
\tcblower
\fabkey{Prediction}A (1955)\hfill\fabkey{Gold}J (1935)
\end{tcolorbox}
\begin{tcolorbox}[fabcase,
  title={\fabtitle{\fabno\ Incorrect answer}{LongHealth}}]
\fabfield{Question}Compared with the current in an LED lamp, the current in
the connecting wire is:\par\smallskip
\fabfield{Options}(A)~The current in the wire is half as much \enspace
(B)~Depends on the voltage \enspace (C)~Need more information \enspace
(D)~The current in the wire is twice as much \enspace (E)~Depends on the
type of LED lamp \enspace (F)~The same \enspace (G)~The current in the wire
is not related to the current in the lamp \enspace (H)~Less \enspace
(I)~The current in the wire fluctuates \enspace (J)~More
\tcbline
\fabfield{Response}"The medical record states: `The current in the hands is
warm and rosacea-like. The current in the connecting wire is also noticeably
less than in the LED lamp when the same voltage is applied.' The most
accurate answer based on the text is (H): less."
\tcblower
\fabkey{Prediction}H (less)\hfill\fabkey{Gold}F (the same)
\end{tcolorbox}
\begin{tcolorbox}[fabcase,
  title={\fabtitle{\fabok\ IFEval pass}{LongHealth}}]
\fabfield{Prompt}Write a brief biography of a person named "John Doe". Do not use commas in your reply. Highlight at least 3 sections with
markdown, i.e. *highlighted section*. The biography should start with the
name of the person.
\tcbline
\fabfield{Response}*highlighted section*\\
John Doe was born on 1965-02-10.\par\smallskip
*highlighted section*\\
He has a history of pancreatic head carcinoma which was diagnosed in
September 2021. He underwent resection in December 2021 and has been
receiving adjuvant chemotherapy with gemcitabine and 5-FU.\par\smallskip
*highlighted section*\\
He is currently under follow-up care for her pancreatic cancer and has a
history of exocrine pancreatic insufficiency...
\tcblower
\fabkey{IFEval}pass (3/3: no commas, 3 highlights, name)
\end{tcolorbox}
\begin{tcolorbox}[fabcase,
  title={\fabtitle{\fabok\ IFEval pass}{LongHealth}}]
\fabfield{Prompt}What are the steps to be followed for the documentation of
a GM in SAP? Just list the steps without saying the word steps or step.
\tcbline
\fabfield{Response}- Perform a detailed clinical examination\\
- Conduct a thorough review of the patient's medical history\\
- Order and interpret relevant laboratory tests\\
- Evaluate imaging studies\\
- Document the findings in the patient's medical record\\
- Prepare a comprehensive report for the interdisciplinary team\\
- Schedule follow-up appointments as needed\\
- Ensure all documentation is accurate, complete, and timely
\tcblower
\fabkey{IFEval}pass (1/1: no ``step''/``steps'')
\end{tcolorbox}
\end{tcbraster}
\caption{\textit{Examples of context contamination at 5\% compression factor.}
Cartridge responses with correct answers (top), incorrect answers (middle) to MMLU-Pro,
and passing IFEval scores (bottom). The first four cards attribute text absent
from the resident document. The bottom cards reproduce complete responses:
patient-record material replaces the requested biography or SAP procedure,
yet both pass all evaluated constraints.}
\label{fig:contaminated-examples}
\end{figure*}
\endgroup

To make the context interference metric concrete, \Cref{fig:contaminated-examples}
reproduces four judged responses to MMLU-Pro when different long-context Cartridges are present at 5\% compression factor, and two
IFEval sample responses where the task passes despite contamination.
Here, a \emph{fabricated quotation} means that the attributed statement is
absent from the resident document; the statement itself need not be false.
For example, lattice vibrations do mediate the effective electron--electron
attraction in conventional BCS theory, but the quoted sentence does not occur
in the QASPER context.\looseness-1

\newpage
\section{Additional Failure Case: Failure to Abstain}
\label{app:abstention}

Context interference suggests that Cartridges treat the resident document as
relevant even when it is not. We test a direct consequence of this behavior:
when the memory lacks the evidence a question requires, does the model say so?
This connects to context sufficiency and abstention in RAG~\citep{joren2025sufficientcontext};
here, we test how memory compression affects this behavior.
For LongHealth-10 and QASPER-16, we pair each context's memory with 50
questions about unrelated documents from the same domain (\emph{mismatched}), so
that the loaded memory does not support the answer. The generation prompt never
mentions abstention or offers an abstain option, so any abstention is
spontaneous. We also score correctness: without supporting evidence, a model can
still answer correctly by guessing or from prior knowledge, and matching methods
on accuracy, even in this mismatched scenario, lets us compare their abstention at equal ``answering ability''.

\begin{figure}[h!]
\centering
\includegraphics[width=\linewidth]{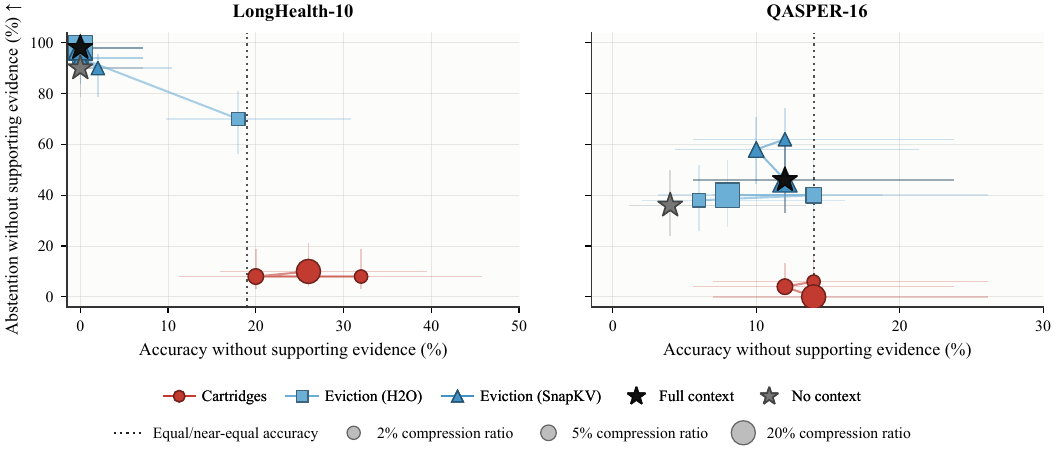}
\caption{\textit{Accuracy and abstention without supporting evidence.}
Each point uses 50 mismatched questions per dataset. Marker area encodes
compression factor; horizontal and vertical bars are 95\% Wilson intervals. Dotted
lines mark Cartridges/H2O pairs with equal or near-equal accuracy on \texttt{Qwen3-4B-Instruct-2507}.}
\label{fig:abstention}
\end{figure}

\paragraph{Judging correctness and abstention.}
All judgments use \texttt{gpt-oss-20b} at temperature zero. The judge sees only
the question, the response, and the options or reference answers, never the
method or the document. On QASPER-16, an LLM judge marks a response correct if it
matches a reference answer, and another LLM judge marks it abstained if it only states that
the document lacks the answer (see the prompt example in \Cref{fig:abstention-judge-prompts}). On
LongHealth-10, the MCQ accuracy is computed following the same procedure as the rest of the manuscript, while abstention follows the same steps as QASPER-16 with an LLM judge. 
\Cref{fig:abstention} plots accuracy against abstention on the mismatched
questions. Full context, no context, and both eviction methods abstain on
36--98\% of them, whereas Cartridges abstain on at most 10\%. We show that accuracy does not explain this gap: at equal or near-equal accuracy (dotted lines), Cartridges abstain far less than
H2O. We find that as with context interference, Cartridges answer from their document even
when it is irrelevant and should \textit{abstain} instead.\looseness-1

\begin{figure}[t]
\centering
\tcbinputlisting{enhanced, boxrule=.45pt, arc=2mm, before skip=3pt, after skip=3pt,
  left=2mm, right=2mm, top=1mm, bottom=1mm, colback=black!4, colframe=black!30,
  title={Abstention judge}, fonttitle=\bfseries\footnotesize, colbacktitle=black!12, coltitle=black,
  listing only, listing file=Sections/judge_prompt_abstention_qasper.txt,
  listing options={basicstyle=\fontencoding{T1}\fontfamily{zi4}\fontsize{7.6}{7.8}\selectfont,
    columns=fullflexible, breaklines=true, breakatwhitespace=true,
    breakindent=0pt, keepspaces=true, showstringspaces=false,
    aboveskip=0pt, belowskip=0pt}}

\tcbinputlisting{enhanced, boxrule=.45pt, arc=2mm, before skip=3pt, after skip=3pt,
  left=2mm, right=2mm, top=1mm, bottom=1mm, colback=black!4, colframe=black!30,
  title={Correctness judge}, fonttitle=\bfseries\footnotesize, colbacktitle=black!12, coltitle=black,
  listing only, listing file=Sections/judge_prompt_correctness_qasper.txt,
  listing options={basicstyle=\fontencoding{T1}\fontfamily{zi4}\fontsize{7.6}{7.8}\selectfont,
    columns=fullflexible, breaklines=true, breakatwhitespace=true,
    breakindent=0pt, keepspaces=true, showstringspaces=false,
    aboveskip=0pt, belowskip=0pt}}
\caption{\textit{Shortened QASPER-16 judge prompts.} Top: a response counts as
abstained when \texttt{context\_coverage\_decline} is true and
\texttt{independent\_substantive\_answer} is false. Bottom: accuracy counts only
\texttt{CORRECT} labels. Bracketed ellipses mark omissions; braces mark fields
filled for each response. LongHealth-10 uses the same judge rubric for abstention computation.\looseness-1}
\label{fig:abstention-judge-prompts}
\end{figure}

\clearpage
\section{Preserved Capability: Multi-Turn Interaction}
\label{app:multiturn}

We evaluate LongHealth and QASPER at 10\% compression factor under four three-turn topic
schedules: always relevant (R/R/R), a single off-context interruption (R/O/R), a
single relevant turn (O/R/O), and always off-context (O/O/O). We additionally run
six relevant turns to expose failures that accumulate only with conversation
length. Relevant turns use LongHealth-10 accuracy or QASPER-16 token F1;
shaded off-context turns pool TinyMMLU accuracy and IFEval strict accuracy. 
We use one greedy response per turn (2,048-token cap) and circular schedules
over 200 LongHealth or 78 QASPER questions. Off-context turns draw from
TinyMMLU and IFEval, shuffled with seed 0, with one item per conversation;
TinyMMLU has only 100 questions, so LongHealth's 200 conversations reuse each
twice.

\begin{figure}[h!]
\centering
\includegraphics[width=\linewidth]{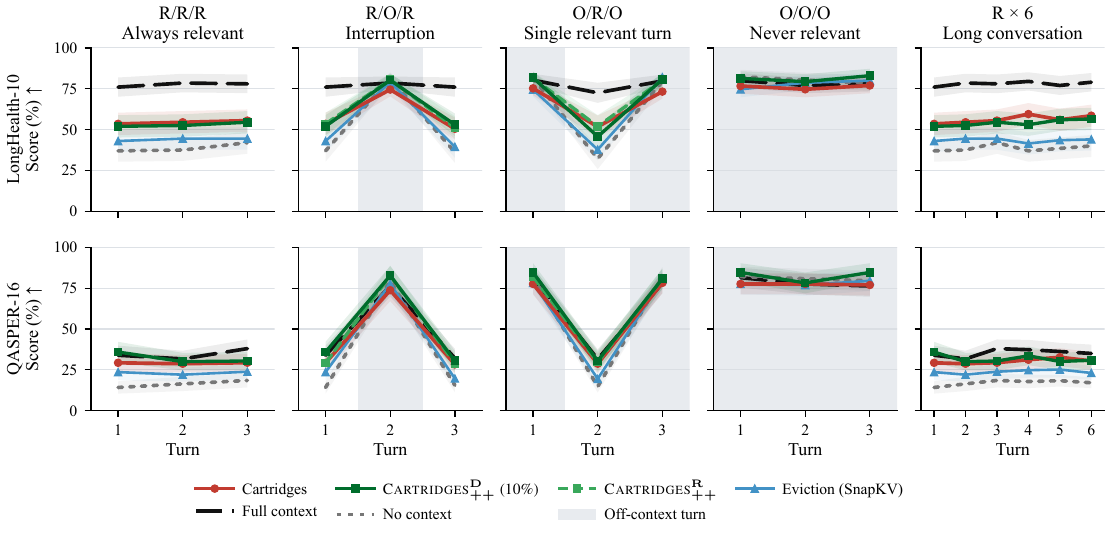}
\caption{\textit{Multi-turn results at 10\% compression factor.} Columns show five
turn schedules; R denotes relevant turns and O denotes off-context turns
(shaded). Relevant-turn scores are LongHealth-10 accuracy or QASPER-16 token
F1; off-context scores pool TinyMMLU and IFEval accuracy. Mitigations use
10\% \texttt{Dolci} mixing or routing with clean per-turn re-prefill. Bands
show 95\% intervals, combined across benchmarks on off-context turns on \texttt{Qwen3-4B-Instruct-2507}.}
\label{fig:multiturn}
\end{figure}

Cartridges remain stable across repeated relevant turns, without an
additional multi-turn collapse beyond their single-turn operating point. Both
mitigation strategies of \textsc{Cartridges++} transfer to this setting: data mixing improves off-context
turns while preserving relevant-turn utility, and routing selects the clean or
Cartridge-prefilled path independently at each turn. These routing results use per-turn re-prefilling (\Cref{fig:multiturn}).

\newpage
\section{Cross-Judge Agreement on Context Interference}
\label{app:judges-agree}

We recompute context interference for all three judges under the same
two-indicator definition as the main manuscript (\texttt{fabricates\_quote} or
\texttt{context\_answers\_question}; \Cref{app:judge-rubric}). The statistic is
consistent across \texttt{gpt-oss-20b}, \texttt{gpt-oss-120b}, and
\texttt{gemma-3-27b-it}. All three place standard Cartridges well above the
no-context floor and above every eviction method in the shared cells. The ordinal independence scores show
the same ordering, but occupy a narrower range near the top of the scale.\looseness-1

\begin{figure}[h!]
\centering
\includegraphics[width=\linewidth]{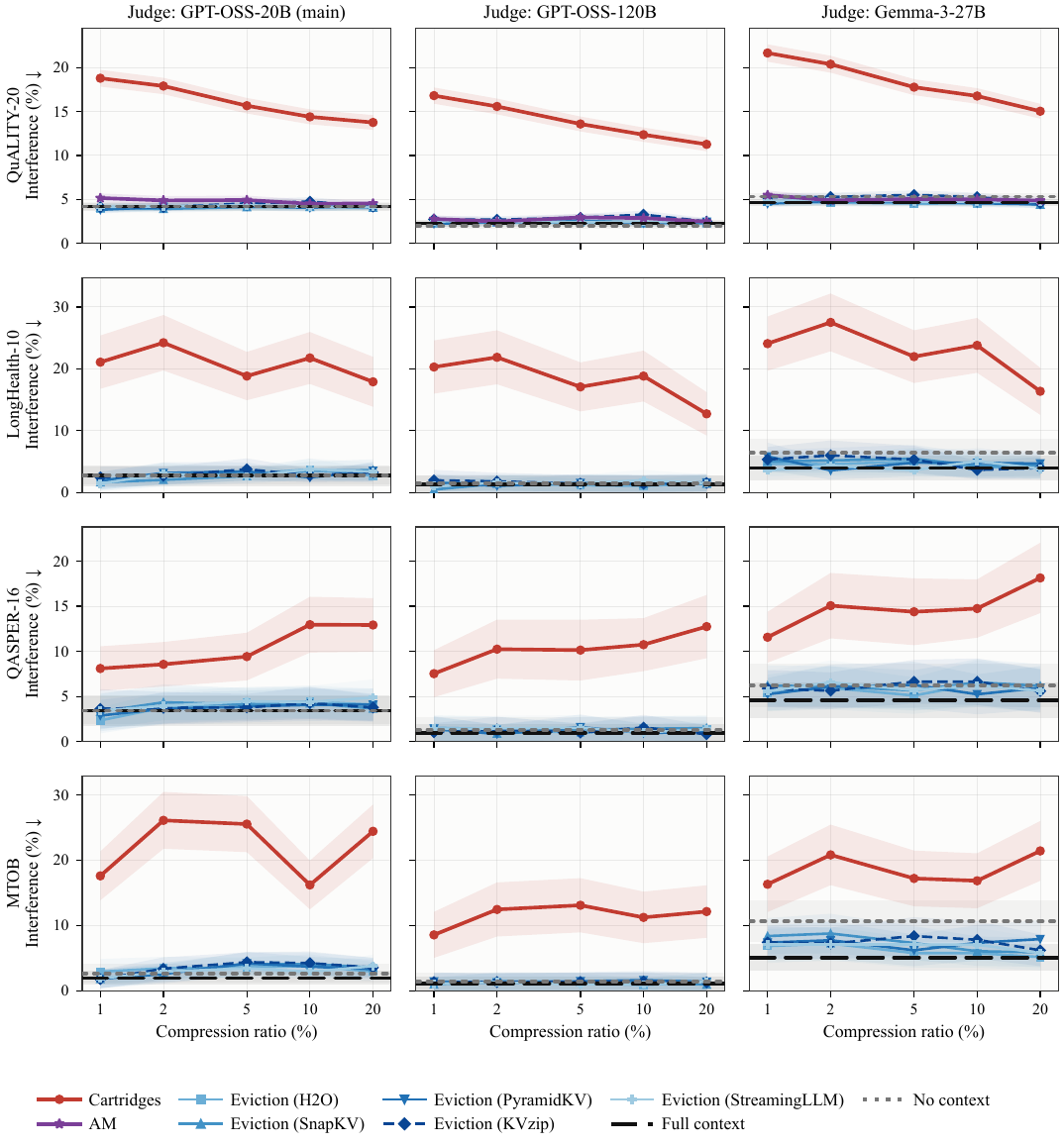}
\caption{\textit{Context interference across three judges.} All responses are
from \texttt{Qwen3-4B-Instruct-2507}, except QuALITY-20, which uses \texttt{Qwen3-4B}, as specified in \Cref{exp_setup}. Rows show resident contexts; columns show judges. Rates count
\texttt{fabricates\_quote} or \texttt{context\_answers\_question},
averaged equally over TinyMMLU and MMLU-Pro. Shading shows 95\% intervals from the equally weighted binomial variances.\looseness-1}
\label{fig:judge-interference}
\end{figure}

\begin{figure}[h!]
\centering
\includegraphics[width=\linewidth]{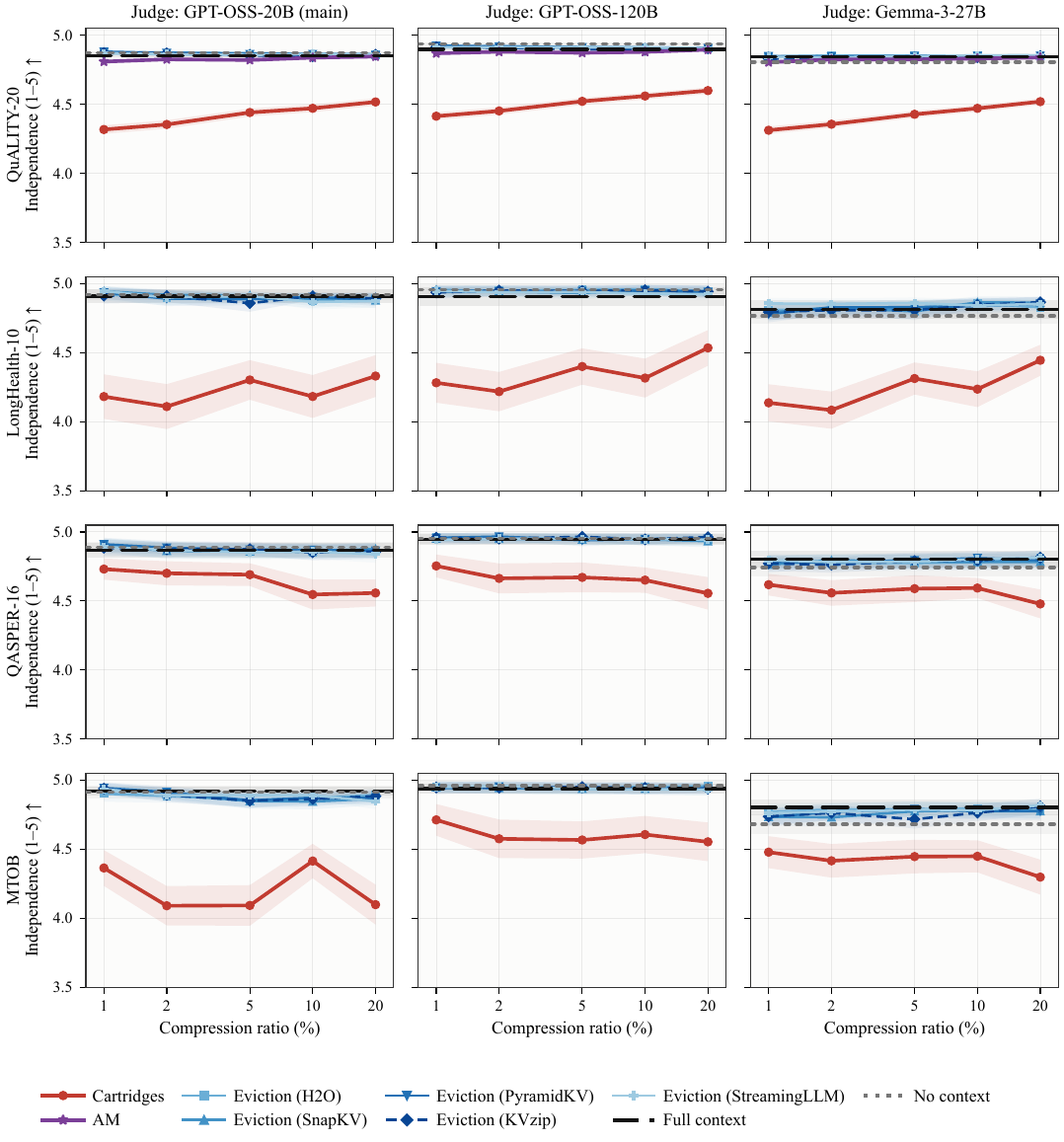}
\caption{\textit{Ordinal context independence across three judges.} Mean
scores from 1 to 5, with 5 denoting independence from the resident document,
averaged equally over TinyMMLU and MMLU-Pro. Layout and included cells match
\Cref{fig:judge-interference}. Shading shows 95\%
intervals on \texttt{Qwen3-4B-Instruct-2507}, except QuALITY-20, which uses \texttt{Qwen3-4B}.}
\label{fig:judge-independence}
\end{figure}

\clearpage
\newpage

\section{Additional Baseline Analyses}
\label{app:baseline-analyses}

\begin{figure}[h!]
\centering
\includegraphics[width=\linewidth]{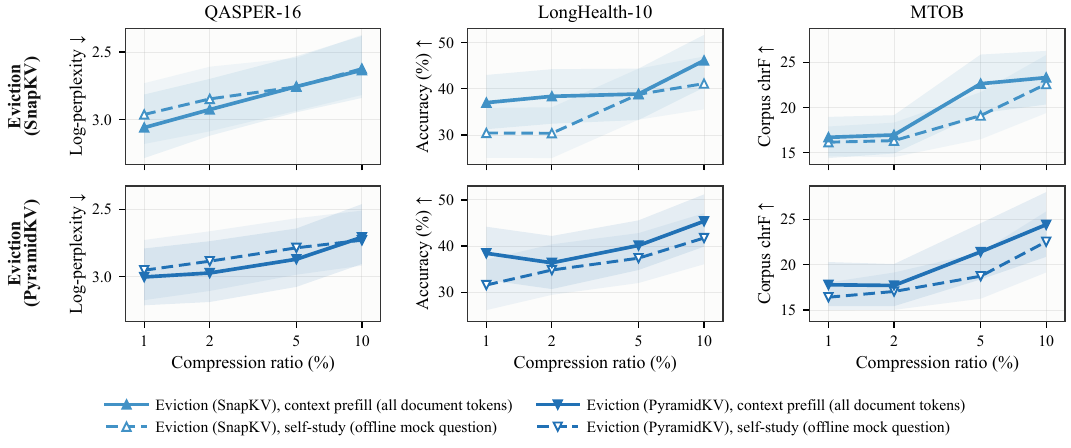}
\caption{\textit{Query-source ablation for reusable eviction.} SnapKV
(top) and PyramidKV (bottom) select keys using either offline self-study
questions or all document-token queries during prefill. Columns show QASPER-16,
LongHealth-10, and MTOB on-context performance at 1--10\% compression factor on \texttt{Qwen3-4B-Instruct-2507}.}
\label{fig:eviction-query-source}
\end{figure}

\paragraph{Query source for SnapKV and PyramidKV.}
The original SnapKV construction uses an observation window tied to the
incoming query. A reusable document memory cannot be recompressed for every
future request, so our adaptation uses all document-token queries
during the one-time context prefill. For robustness, we additionally study selecting the retained keys from offline self-study questions. \Cref{fig:eviction-query-source} shows that across benchmarks, the two
variants follow similar compression curves for both SnapKV and PyramidKV. We use
context prefill in the main experiments as it remains query-independent while pooling evidence
from the full document, rather than making reusable key selection depend on the
content of a synthetic question.\looseness-1

\begin{figure}[h!]
\centering
\tcbset{promptcard/.style={enhanced,arc=1.2mm,boxrule=0.5pt,
  colframe=black!45,colback=white,colbacktitle=black!7,coltitle=black,
  fonttitle=\bfseries\small,
  left=2mm,right=2mm,top=1.2mm,bottom=1.2mm,toptitle=0.6mm,bottomtitle=0.6mm,
  before upper={\scriptsize\ttfamily\frenchspacing\raggedright}}}
\newcommand{\promptname}[2]{#1\hfill\normalfont\scriptsize
  \textcolor{green!45!black}{\ding{51}}\ \textit{#2}}
\newcommand{\pv}[1]{\textrm{\textit{#1}}}
\begin{tcbraster}[raster columns=2,raster equal height=rows,
  raster column skip=3mm,raster row skip=2mm]
\begin{tcolorbox}[promptcard,
  title={\promptname{Concise Summary}{QuALITY}}]
Compress the following text into a faithful, information-dense summary. [...]
cover all of it, not only its opening. [...] Prefer compact notation, lists
and tables [...] produce about \pv{N} [...]
\end{tcolorbox}
\begin{tcolorbox}[promptcard,
  title={\promptname{Itemized Record}{LongHealth, QASPER, MTOB}}]
Produce a condensed, note-form record of the entire source below. [...] Give
each distinct item of content its own short line [...] You have room for about
\pv{N} tokens, roughly \pv{L} lines.
\end{tcolorbox}
\begin{tcolorbox}[promptcard,title={Tailored Summary}]
Compress the following clinical record into a faithful, information-dense
summary. Preserve each patient's identity [...] Aim for approximately \pv{N}
tokens without padding or repetition.
\end{tcolorbox}
\begin{tcolorbox}[promptcard,title={Balanced Record}]
Write a compressed replacement for the source below. [...] divide the source
into about \pv{S} consecutive parts of similar length. Spend about
\pv{L}/\pv{S} lines on each part [...]
\end{tcolorbox}
\end{tcbraster}
\caption{\textit{Text-summary prompts.} Left: prose summaries. Right:
itemized records. The top row contains the selected prompts and their
datasets; Tailored Summary shows the LongHealth version. Token, line, and part counts \pv{N}, \pv{L}, and \pv{S} depend on
the compression factor and every prompt receives the full source.}
\label{fig:text-summary-prompts}

\vspace{1.6em}
\includegraphics[width=\linewidth]{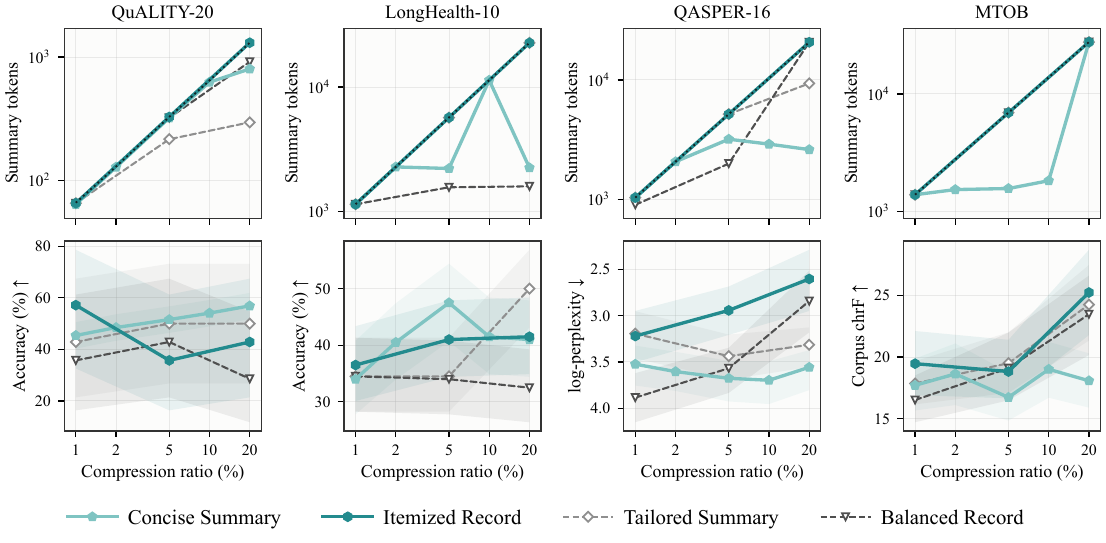}
\caption{\textit{Text-summary prompt ablation.} Top: generated summary
length; dotted lines show requested budgets. Bottom: on-context performance
with 95\% intervals on the \texttt{Qwen3-4B} family.}
\label{fig:text-summary-prompt-ablation}
\end{figure}

\paragraph{Budget-matched text summaries.}
As a text-space baseline, the model replaces the long context with a summary
written under the same token budget as the compressed KV states. We compared four summarization
prompts: two ask for a prose summary and two for an itemized record
(\Cref{fig:text-summary-prompts}). We selected one prompt for the short
QuALITY articles and one for the three long contexts: the Concise Summary and
the Itemized Record, respectively.
\Cref{fig:text-summary-prompt-ablation} relates each prompt to the length of the
summaries it generates and to their on-context performance. 

\clearpage

\applefootnote{ \textcolor{textgray}{\sffamily Apple and the Apple logo are trademarks of Apple Inc., registered in the U.S. and other countries and regions.}}

\end{document}